\documentclass[10pt,twocolumn,letterpaper]{article}

\usepackage{cvpr}              

\definecolor{cvprblue}{rgb}{0.21,0.49,0.74}
\usepackage[pagebackref,breaklinks,colorlinks,allcolors=cvprblue]{hyperref}
\usepackage{graphicx}
\usepackage{multirow} 
\usepackage{subcaption}
\usepackage{booktabs}
\usepackage{xspace}
\usepackage{pifont}
\usepackage{tikz}
\usepackage{caption}
\usepackage{xcolor}
\usepackage{booktabs}  
\usepackage[table]{xcolor}
\usepackage[most]{tcolorbox}
\definecolor{AssumeBlue}{RGB}{30,90,160}     
\definecolor{AssumeBg}{RGB}{245,248,253}     
\tcbset{
  myassumbox/.style={
    enhanced,
    colback=AssumeBg,
    colframe=AssumeBlue,
    boxrule=0.6pt,
    arc=2pt,
    left=9pt,right=9pt,top=7pt,bottom=7pt,
    before skip=6pt, after skip=6pt,
    borderline west={2pt}{0pt}{AssumeBlue}, 
    breakable
  }
}
\newtcolorbox{assumptionbox}[1][]{myassumbox,#1}
\graphicspath{{figs/}{figures/}{images/}}
\definecolor{purple}{rgb}{0.65, 0.5, 1.0}
\definecolor{light}{rgb}{0.85, 0.8, 1.0}

\def\paperID{21357}
\def\confName{CVPR}
\def\confYear{2026}

\title{Face2Scene: Using Facial Degradation as an Oracle for Diffusion-Based Scene Restoration}

\author{
Amirhossein Kazerouni$^{1,2,3,4}$ \qquad
Maitreya Suin$^{3}$ \qquad
Tristan Aumentado-Armstrong$^{3}$ \\[2pt]
Sina Honari$^{3}$ \qquad
Amanpreet Walia$^{3}$ \qquad
Iqbal Mohomed$^{3}$ \qquad
Konstantinos G. Derpanis$^{1,2,3,5}$ \\[2pt]
Babak Taati$^{1,2,4}$ \qquad
Alex Levinshtein$^{3}$ \\[6pt]
$^{1}$University of Toronto \qquad
$^{2}$Vector Institute \qquad
$^{3}$AI Center–Toronto, Samsung Electronics \\
$^{4}$University Health Network \qquad
$^{5}$York University
\\[2pt]
{\tt\small amirhossein@cs.toronto.edu, babak.taati@uhn.ca, kosta@yorku.ca} \\
{\tt\small \{m.suin, tristan.a, sina.honari, aman.walia, i.mohomed, alex.lev\}@samsung.com} \\
\small \url{https://amirhossein-kz.github.io/face2scene/}
}

\makeatletter
\newcommand{\cvprteaser}{%
  \vspace{-1em} 
  \begin{center}
    \begin{minipage}{\textwidth}
      \centering
      \includegraphics[width=\textwidth,height=7.3cm,keepaspectratio]{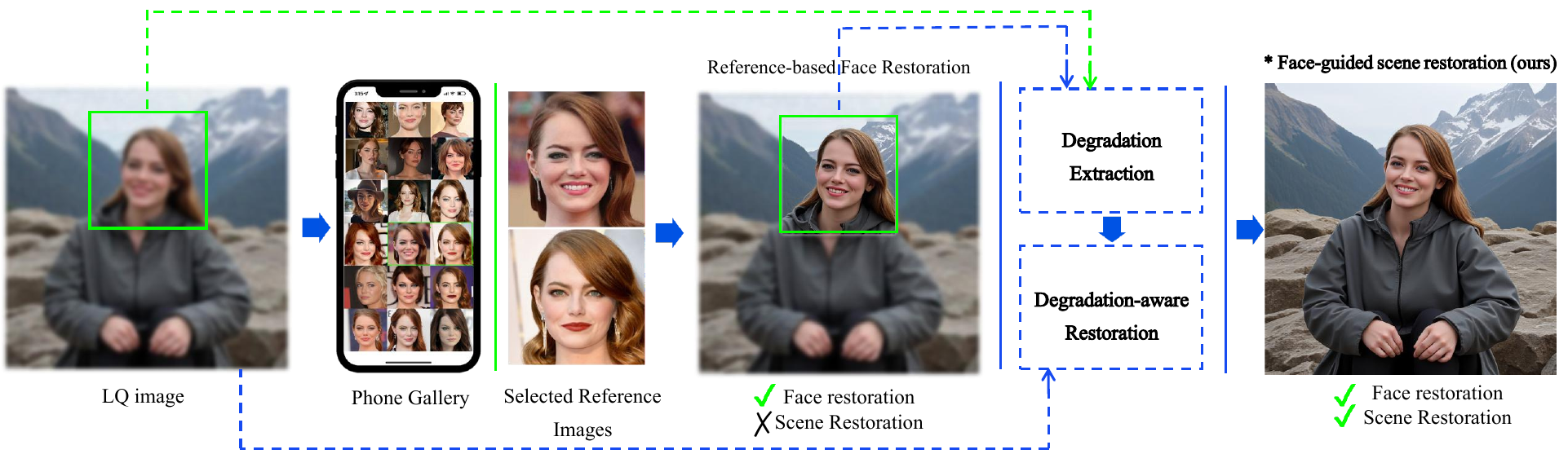}
    \end{minipage}

    \captionsetup{type=figure}
    \vspace{-0.5em}
    \captionof{figure}{
    Overview of \emph{Face2Scene}. We infer a face-derived \textit{degradation code} from identity references and the observed LQ face, then use it as oracle guidance to restore the full scene (face, body, background) with a one-step diffusion restorer. 
    }
    \label{fig:teaser}
  \end{center}
}
\g@addto@macro\@maketitle{\cvprteaser}
\makeatother

\begin{document}
\maketitle

\begin{abstract}
Recent advances in image restoration have enabled high-fidelity recovery of faces from degraded inputs using reference-based face restoration models (Ref-FR). However, such methods focus solely on facial regions, neglecting degradation across the full scene, including body and background, which limits practical usability. Meanwhile, full-scene restorers often ignore degradation cues entirely, leading to underdetermined predictions and visual artifacts. In this work, we propose \textbf{Face2Scene}, a two-stage restoration framework that leverages the face as a perceptual oracle to estimate degradation and guide the restoration of the entire image. Given a degraded image and one or more identity references, we first apply a Ref-FR model to reconstruct high-quality facial details. From the restored–degraded face pair, we extract a face-derived degradation code that captures degradation attributes (e.g., noise, blur, compression), which is then transformed into multi-scale degradation-aware tokens. These tokens condition a diffusion model to restore the full scene in a single step, including the body and background. Extensive experiments demonstrate the superior effectiveness of the proposed method compared to state-of-the-art methods. 
\end{abstract}
\vspace{-1em}
\section{Introduction}
\label{sec:intro}

Recently, image restoration~\citep{wang2018esrgan,wang2021real,yue2025arbitrary,wu2024one,yu2024scaling,yang2024pixel} has witnessed remarkable progress with the advent of diffusion models~\citep{ho2020denoising,rombach2022high}, leading to significant improvements in perceptual quality and realism. Among various sub-domains, reference-based face restoration models (Ref-FR) have emerged as an effective strategy for recovering high-quality facial details from degraded inputs by leveraging clean reference images of the same person. By transferring person-specific priors, such as facial structure, skin, or hair texture, from the reference, these methods achieve superior visual fidelity and identity-specific details~\citep{zhang2025instantrestore,liu2025faceme,tao2025overcoming,hsiao2024ref,chari2025personalized}. However, most Ref-FR methods focus exclusively on the facial region, overlooking the fact that degradations in real-world images, such as blur, noise, and compression, tend to affect both the face and its surrounding scene in a correlated manner.

As a natural step toward restoring larger images, recent works have begun to tackle human-centric restoration beyond faces to full-body recovery~\citep{gong2025haodiff,gong2025human}. In parallel, general-purpose restoration models trained on large-scale natural images show robustness to varied degradations, yet they often fail to preserve facial realism, identity consistency, and coherent human-body structure under complex full-scene degradations~\citep{yu2024scaling,kong2025dual,wang2024sinsr,liu2024adaptbir}. A key limitation is that many full-image approaches are non-degradation-aware, conditioning only on the degraded input; the problem is then underdetermined and prone to visual artifacts. To address this challenge, degradation-aware methods such as S3Diff~\citep{zhang2024degradation} and DeeDSR~\citep{bi2024deedsr} incorporate degradation cues estimated from the low-quality input to guide restoration. However, S3Diff relies on a pre-trained degradation estimator~\citep{mou2022metric} that predicts only two global scalars (noise and blur), which limits its expressiveness. DeeDSR~\citep{bi2024deedsr}, on the other hand, pretrains a degradation encoder via unsupervised contrastive learning~\citep{wang2021unsupervised}, but learning complex degradations solely from the LQ image without supervision remains challenging. In practice, these coarse cues may misinterpret perceptual phenomena (\eg, noise versus texture, lighting) as degradations, ultimately limiting fidelity in full-scene restoration.

In this work, we mitigate both the limited applicability of Ref-FR and the weak degradation-awareness of most modern restoration models, by having the former inform the latter. Figure~\ref{fig:teaser} shows an overview of our approach. 
Specifically, we assume that a high-quality restoration of a patch is informative about the degradation in the full image.
In our case, by leveraging a powerful reference-based face restoration model, \textit{we extract knowledge of the full-image degradation from the restored face patch}.
More precisely, consider a low-quality (LQ) image of a scene, which contains a face patch (crop) with known identity.
Using additional information, namely a set of same-identity reference images, a Ref-FR model reconstructs a high-quality (HQ) version of that face crop.
This reconstructed LQ-HQ pair, which utilized external information in the reference images for restoration, is then used to directly estimate the corruption affecting the image, rather than relying on blind estimation via the LQ image alone.
To apply the degradation information encoded by the LQ-HQ pair, we encode it into a face-derived degradation code, which then guides the restoration of the full scene, including body, clothing, and background.
This approach offers a much stronger degradation prior, which includes person-specificity, compared to the coarse assumptions that can be extracted from the LQ alone.
The result is a simple, practical path: use Ref-FR to infer the degradation from where the signal is strongest (the face), then use that as an oracle to restore the complete image.

Concretely, we can use any Ref-FR model, with associated reference images, to obtain an LQ-HQ pair. We then extract a degradation code from it via a contrastive model, which we call \textit{FaDeX} (Face-derived Degradation eXtractor). Next, we use the code to condition a diffusion-based restoration model, by converting it to a usable form with our trained \textit{MapNet} (Degradation Mapping Network). The resulting tokens condition the restoration, so it is not only denoising the LQ input but is also explicitly guided on \emph{what corruption to undo}. This enables consistent restoration across the scene, based on the degradation information in the LQ-HQ face residual.

In addition, a key challenge is the scarcity of large-scale full-scene image datasets with identity references, despite the ubiquity and practicality of such scenarios. 
To address this, we aggregate multiple real-world person galleries and generate synthetic full-scene images using identities from CelebRef-HQ~\cite{li2022learning}. We construct a benchmark with standardized training, validation, and test splits. This dataset provides HQ full-scene images paired with reference photos of the same identity, enabling the model to be evaluated on and trained for (i) accurate face-based degradation estimation and (ii) robust scene-level restoration conditioned on that estimate. 

We summarize our contributions as follows:
\begin{itemize}
    \item \textbf{Facial degradation as an oracle.} We recast diffusion-based scene restoration as a degradation-conditioned task by deriving a degradation code from an HQ--LQ face pair, where the HQ face is reconstructed via Ref-FR using reference images of the same identity.
    \item \textbf{FaDeX $\rightarrow$ MapNet conditioning.} \emph{FaDeX} learns a degradation embedding from the face pair, disentangled from image content; \emph{MapNet} converts it into multi-scale degradation tokens that condition a SD-Turbo~\cite{sauer2024adversarial} diffusion model in one step.
    \item \textbf{Benchmark with identity references.} To support training and evaluation, we curate a full-scene dataset by aggregating real person galleries and synthesizing scenes with CelebRef\mbox{-}HQ identities. 
\end{itemize}
Our results show the superior performance of our approach compared to existing state-of-the-art methods. 

\vspace{-0.25em}
\section{Related Work}
\vspace{-0.5em}
\label{sec:related}

\textbf{Reference-Guided Face Restoration.} 
    Reference-guided face restoration aims to recover high-quality facial details from degraded inputs using one or more clean reference images of the same identity. Early diffusion-based approaches achieved personalization through per-identity fine-tuning~\cite{chari2025personalized,varanka2024pfstorer,ding2024restoration}. Later methods, such as RestorerID~\citep{ying2024restorerid} and FaceMe~\citep{liu2025faceme}, rely on ArcFace~\citep{deng2019arcface} embeddings to transfer high-level identity and geometric information. MGFR~\citep{tao2024overcoming} expands the guidance space with text prompts, reference images, and identity features to reduce facial attribute hallucinations (e.g., correcting “big eyes” or enforcing “wearing glasses”). Ref-Guided~\citep{zhou2025reference} and HonestFace~\citep{wang2025honestface} further combine high-level identity cues (ArcFace) with low-level and semantic features, improving the restoration of texture, color tone, and skin details. 
    Methods such as RefSTAR~\citep{yin2025refstar} and Copy or Not?~\citep{chong2025copy} tackle this by incorporating explicit reference selection and region-wise masking to more precisely demarcate the regions to and from which they want to transfer information. 
    However, restoring only the face is insufficient in human-centric images, where the degradation often extends consistently across the entire body and background, motivating the need for scene-level restoration methods.

\textbf{Scene-Level Human Image Restoration.}
Scene-level human image restoration focuses on recovering high-quality body and scene content from degraded human-centric images, extending beyond traditional face-only restoration. Early efforts treated this task as a form of blind image restoration (BIR), but applying generic BIR models~\citep{wu2024one,wang2024sinsr,liu2024adaptbir,kong2025dual} to human bodies often leads to joint misalignment, facial artifacts, and loss of structural coherence due to the unique geometric constraints of human anatomy. To address these issues, specialized methods exploit human-specific priors such as pose, segmentation masks, or structural templates~\citep{liu2021accurate,wang2024prior}. DiffBody~\citep{zhang2024diffbody}  introduced diffusion models for body-region enhancement, integrating pose-attention, text cues, and body-centered sampling. OSDHuman~\citep{gong2025human} further advances the field with a one-step diffusion model guided by a high-fidelity human embedder. More recent approaches, such as HAODiff~\citep{gong2025haodiff}, model the coexistence of human motion blur and generic degradation through synthetic pipelines and propose dual-prompt classifier-free guidance for robust one-step restoration. However, most of the scene-level restoration methods do not leverage reference cues.

\textbf{Degradation-Aware Conditioning in Diffusion Models.}
Beyond simply feeding the LQ image to a fixed denoiser for scene level restoration without knowledge of the corruption process, recent methods explicitly explore where and how degradation cues enter the diffusion model. DeeDSR~\cite{bi2024deedsr} learns a degradation embedding from the LQ image and injects it into the UNet features via ControlNet~\cite{zhang2023adding} to preserve semantics under heavy corruption. S3Diff predicts two global scalars (noise, blur) from the LQ image and modulates LoRA weights accordingly—an efficient but inherently coarse, parameter-space conditioning~\cite{zhang2024degradation}. 
In contrast, text-prompt approaches describe degradations in language and map them to conditioning signals~\cite{chen2023image,qi2023tip}. Despite different injection approaches, these pipelines still infer degradation solely from the LQ image, limiting their expressiveness and reliability under complex real-world degradations. In contrast, we extract a face-derived degradation code from an HQ–LQ face pair of the same identity and use MapNet to convert it into multi-scale tokens that condition the diffusion model for scene-wide correction. By treating the restored face as a degradation oracle, our method unifies reference-guided face restoration, scene-level enhancement, and degradation-aware diffusion into a single coherent framework.
\begin{figure*}[t]
    \centering
    \includegraphics[width=\textwidth]{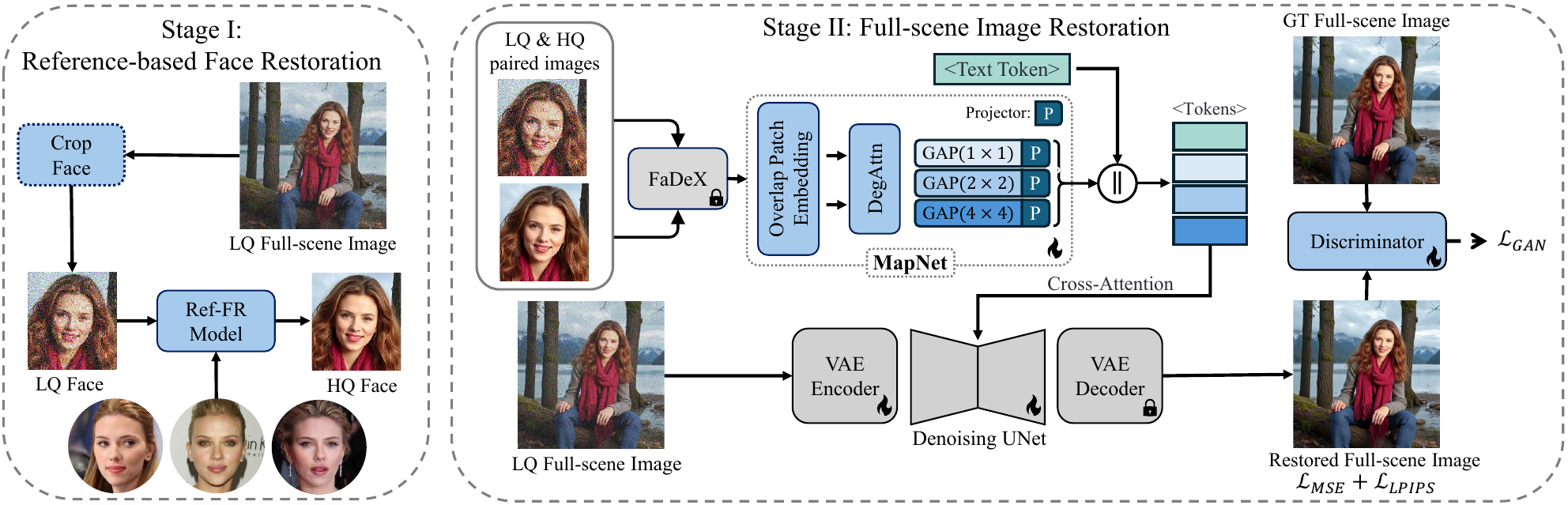}
    \vspace{-1.6em}
    \caption{Overview of the \textbf{Face2Scene} pipeline. In stage I, we leverage a set of reference faces to restore the LQ face crop. In stage II, we use the pair of LQ and HQ faces to extract a guided degradation with FaDeX and inject it into a one-step diffusion model using MapNet. The diffusion model then reconstructs the full-scene image.}
    \label{fig:main}
    \vspace{-1em}
\end{figure*}
\vspace{-0.25em}
\section{Technical Approach}
\vspace{-0.5em}
Given a large-scale pre-trained text-to-image (T2I) diffusion model, our goal is to build an efficient yet powerful full-scene restoration system that (i) uses identity references to reliably infer degradation and (ii) uses the degradation to restore the entire scene. 
We leverage readily available reference images, that each person usually has in their personal photo gallery, not only to restore the face, but to drive scene-wide restoration of body and scene background. We introduce \textbf{Face2Scene}, a two-stage framework that \textit{first} estimates degradation from faces (Section~\ref{subsec:stage1}), and \textit{then} uses them as an oracle to restore the entire scene (Section~\ref{stage2}).

\subsection{Stage 1: Reference-Based Face Restoration}
\label{subsec:stage1}
We are given a degraded full-scene image, $I^{\mathrm{LQ}}$, and one or more identity references, $\{x^{\mathrm{ref}}_k\}$. Faces that appear in full scenes have arbitrary scale and in-plane rotation and are rarely centered, while most off-the-shelf Ref-FR models assume an aligned input at a fixed resolution (typically $512{\times}512$~\cite{liu2025faceme,ying2024restorerid, zhang2025instantrestore, hsiao2024ref}) with canonical landmark placement (eyes, nose, lips). We therefore first detect the face in $I^{\mathrm{LQ}}$, extract a squared crop, $x^{\mathrm{LQ}}$, and estimate a rigid alignment (rotation and translation) that maps the detected landmarks to a canonical template. The crop is then warped into the canonical frame and resized to $512{\times}512$, yielding an aligned degraded face, $\tilde{x}^{\mathrm{LQ}}$, suitable for Ref-FR model inference. Given $\tilde{x}^{\mathrm{LQ}}$ and the aligned reference set, $\{x^{\mathrm{ref}}_k\}$, an off-the-shelf Ref-FR model, $F_\theta$, produces a high-quality reconstruction in the canonical frame,
\[
\tilde{x}^{\mathrm{HQ}} \;=\; F_\theta\big(\tilde{x}^{\mathrm{LQ}}, \{x^{\mathrm{ref}}_k\}\big).
\]
Since Ref-FR models are trained for a fixed input size, the alignment-and-resize step is crucial: faces in the original scene can be much smaller or larger than $512{\times}512$, and feeding them directly would significantly degrade restoration quality. After reconstruction, we invert the alignment by applying the inverse warp (including the inverse of the resizing) to obtain $x^{\mathrm{HQ}}$ at the native scale and location within the scene. As a result, we obtain a spatially aligned pair, $(x^{\mathrm{HQ}}, x^{\mathrm{LQ}})$, in the original scene coordinates. \autoref{fig:main} (left) gives an overview of this stage.

Faces are effective probes for estimating image degradations for three key reasons:
(i) they exhibit stable geometry with reliable landmark structure, unlike backgrounds or incidental objects with high intra-class variability;
(ii) identity references are often available, enabling a Ref-FR model to reconstruct a high-quality image, \(x^{\mathrm{HQ}}\), with minimal hallucination; and
(iii) the discrepancy between this HQ reconstruction and the observed degraded face, \(x^{\mathrm{LQ}}\), largely isolates corruption from content.
We take the pair and feed it to a degradation encoder to obtain a face-derived code, which conditions the one-step diffusion restorer, explicitly conveying the type and severity of degradation to undo.

\subsection{Stage 2: Full-Scene Image Restoration}
\label{stage2}

\textbf{FaDeX: Face-derived Degradation eXtractor.}
Accurate modeling of unknown degradations is essential for real-world restoration. Prior approaches often describe degradation using the LQ image either as text (prompts)~\cite{chen2023image,qi2023tip} or as scalar scores (e.g., noise/blur levels)~\cite{zhang2024degradation}. While such cues provide coarse priors, they struggle with mixed and complex degradations; even rich textual descriptions are difficult for current vision–language models to translate into actionable restoration signals. Leveraging Ref-FR, we form HQ--LQ face pairs that isolate corruption from content, allowing us to measure degradation rather than guess it.

\textbf{Assumption (image-wide consistency).}
\textit{An unknown, spatially homogeneous operator, $\mathcal{G}$, generates the LQ image:
$I^{\mathrm{LQ}}=\mathcal{G}(I^{\mathrm{HQ}})$.}
We construct supervision by sampling two clean images, $I_a^{\mathrm{HQ}}$ and $I_b^{\mathrm{HQ}}$, and applying the same degradation operator, $\mathcal{G}$, to obtain $I_a^{\mathrm{LQ}}=\mathcal{G}(I_a^{\mathrm{HQ}})$ and $I_b^{\mathrm{LQ}}=\mathcal{G}(I_b^{\mathrm{HQ}})$. Ref-FR then yields aligned face pairs, $(x_a^{\mathrm{HQ}},x_a^{\mathrm{LQ}})$ and $(x_b^{\mathrm{HQ}},x_b^{\mathrm{LQ}})$, for the same $\mathcal{G}$. To encourage more robust convergence and reduce dependency on a single Ref-FR, we additionally create pairs, where $x^{\mathrm{HQ}}$ is the ground-truth (GT) face crop instead of a Stage~1 reconstruction; this exposes the learner to identical $\mathcal{G}$ across diverse contents.

FaDeX employs a lightweight convolutional encoder, $E_\phi$, that takes the HQ and LQ faces channel-wise concatenated. Let $\tilde{x}=[\,x^{\mathrm{HQ}} \,\|\, x^{\mathrm{LQ}}\,]\in\mathbb{R}^{H\times W\times 6}$ denote the stacked input. The encoder outputs a face-derived
degradation embedding, $Z_{\text{face}} \;=\; E_\phi(\tilde{x}) \in \mathbb{R}^{H' \times W' \times C}.$
We pool the spatial embedding to a vector and project it with a single head:
\[
z \;=\; \operatorname{GAP}(Z_{\text{face}}) \in \mathbb{R}^{C}, \qquad
u \;=\; g_\psi(z) \in \mathbb{R}^{p},
\]
where \(\operatorname{GAP}\) denotes global average pooling and \(g_\psi\) is an MLP projection head whose output embedding is unit-normalized $q=u/\|u\|_2$ for contrastive learning.

Since degradation is assumed to be consistent within an image but varies across images, we adopt a contrastive objective~\cite{chen2020simple} that attracts embeddings sharing the same $\mathcal{G}$ and repels others. Let each sample, $i$, in a batch, $\mathcal{B}$, carry a degradation label, $g_i$, indicating the operator used to synthesize its LQ counterpart. Define positive examples, $\mathcal{P}(i)=\{\,p\in\mathcal{B}\setminus\{i\}\mid g_p=g_i\,\}$, and negative examples, $\mathcal{N}(i)=\mathcal{B}\setminus(\{i\}\cup\mathcal{P}(i))$. The loss is
\begin{align}
\mathcal{L}_{\mathrm{Deg}}
= \sum_{i\in\mathcal{B}} \sum_{p\in\mathcal{P}(i)} -
\log
\frac{\exp\!\big(\langle q_i,q_p\rangle/\tau\big)}
{\sum\limits_{a\in\mathcal{P}(i)\cup\mathcal{N}(i)} 
\exp\!\big(\langle q_i,q_a\rangle/\tau\big)} ,
\end{align}
where $\langle\cdot,\cdot\rangle$ denotes the cosine similarity and $\tau$ is a temperature.    

The resulting FaDeX model learns a representation that explains the residual between $x^{\mathrm{LQ}}$ and $x^{\mathrm{HQ}}$, thereby capturing the underlying degradation. We then freeze FaDeX and use its $Z_{\text{face}}$ embedding as the degradation code to condition the one-step SD-Turbo~\cite{sauer2024adversarial} restorer. Since this degradation code is not natively aligned with the diffusion model’s conditioning space, we introduce MapNet, a degradation mapping network that maps $Z_{\text{face}}$ from the degradation space into the diffusion space for effective conditioning.

\noindent\textbf{MapNet: Degradation Mapping Network.}
MapNet converts the spatial FaDeX features, $Z_{\text{face}}$, into a compact set of conditioning tokens for the one-step restorer. However, the pre-trained SD-Turbo restorer operates in a different representational space, aligned with its original text-image conditioning. MapNet bridges this gap, transforming the degradation features into a form compatible with the diffusion model's conditioning mechanism.

We first process $Z_{\text{face}}$ into a sequence of tokens by applying an overlap patch embedding (a $3{\times}3$ Conv with stride 2, followed by LayerNorm), yielding $F \in \mathbb{R}^{\frac{H'}{2}\times\frac{W'}{2}\times 2C}$. We then split $F$ into two branches, $F_1$ and $F_2$, each yielding a value tensor, $V_i$, and an independent attention map, $A_i$. The final interaction is computed as a residual attention, $\mathrm{DegAttn}(Z_{\text{face}}) = \big(A_1 - \lambda A_2\big)[V_1 \, ; \, V_2],$ where \(\lambda\) is a learnable scalar and \([V_1 \, ; \, V_2]\) is the channel-wise concatenation of the value tensors. This design encourages the module to refine the extracted information from the FaDeX. From the $\mathrm{DegAttn}$ output, we extract the final multi-scale degradation tokens via grid average pooling at three scales: $\mathrm{GAP}(4{\times}4)$, $\mathrm{GAP}(2{\times}2)$, and $\mathrm{GAP}(1{\times}1)$. Each pooled cell passes through a small MLP and LayerNorm, producing $16{+}4{+}1=21$ tokens. These tokens are concatenated with generic positive text tokens to condition the one-step $\mathrm{SD}$-$\mathrm{Turbo}$ restorer using the Cross-attention of the stable diffusion. \autoref{fig:main} (right) illustrates this stage. 
\vspace{-0.25em}
\subsection{Loss Functions}
\vspace{-0.25em}
To train the diffusion model, we use a reconstruction loss and an adversarial loss. The reconstruction term
\[
\mathcal{L}_{\mathrm{rec}} \;=\; \lambda_{2}\,\| \hat I - I^{\mathrm{HQ}} \|_{2}^{2} \;+\; \lambda_{\mathrm{LPIPS}}\,\mathrm{LPIPS}(\hat I, I^{\mathrm{HQ}})
\]
combines an $\ell_2$ loss with a perceptual LPIPS loss between the prediction, $\hat I = G_{\theta}(I^{\mathrm{LQ}})$, and the ground-truth, $I^{\mathrm{HQ}}$.

Inspired by adversarial distillation in ADD~\cite{sauer2024adversarial}, we further introduce a GAN term to reduce the distribution gap between restored and real HR images, but without a diffusion teacher or intermediate supervision (unlike the four-step setting in previous work~\cite{sauer2024adversarial}). The discriminator, $D_{\phi}$, follows prior work~\cite{kumari2022ensembling,zhang2024degradation} and uses a frozen DINO backbone~\cite{caron2021emerging} with multiple independent classifiers attached to different feature levels. The GAN objective is
\begin{align}
\mathcal{L}_{\text{GAN}}
&= \mathbb{E}_{I_{\text{HQ}}\sim \mathcal{P}_{\text{HQ}}}
\big[\log D_\phi(I_{\text{HQ}})\big] \notag\\[-2pt]
&\quad + \mathbb{E}_{I_{\text{LQ}}\sim \mathcal{P}_{\text{LQ}}}
\big[\log\big(1 - D_\phi(G_\theta(I_{\text{LQ}}))\big)\big].
\end{align}
Our complete loss is then
\begin{equation}
\mathcal{L}(\theta) \;=\; 
\mathcal{L}_{\text{rec}}
+ \lambda_{\text{GAN}}\,\mathcal{L}_{\text{GAN}},
\end{equation}
where $\lambda_{\text{GAN}}$ balances the adversarial signal against reconstruction, yielding a teacher-free, single-step objective that supervises only the final output.
\begin{figure*}[t]
    \centering
    \includegraphics[width=0.92\textwidth]{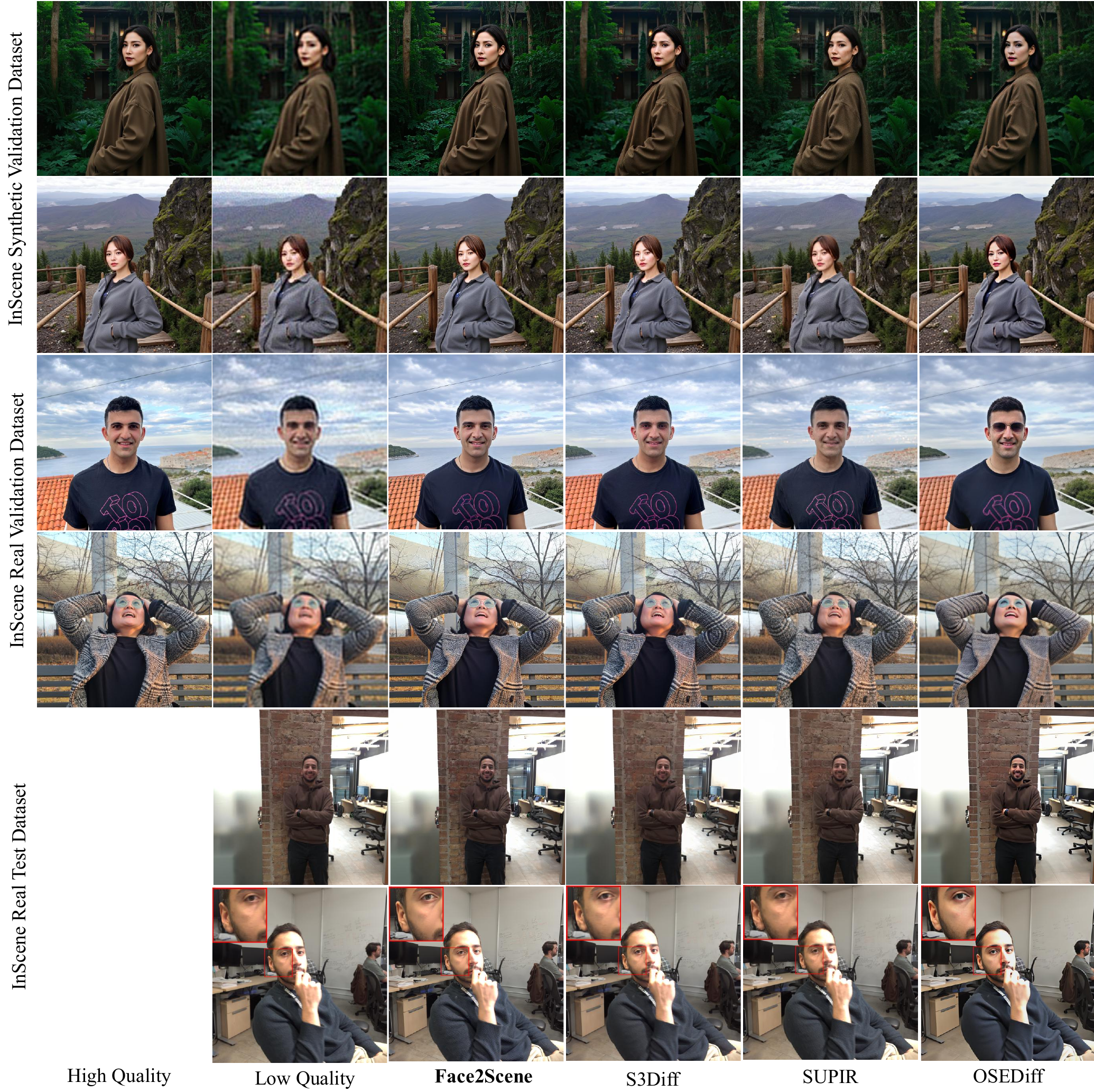}
    \caption{Visual comparison of Face2Scene with the three top-performing methods from the quantitative results (zoom in to see details).}
    \label{fig:main_result}
    \vspace{-1.25em}
\end{figure*}
\vspace{-0.5em}
\section{Experiments}
\vspace{-0.25em}

\subsection{Experimental Settings}

\begin{table}[t]
\centering
\caption{\textbf{Dataset Description.}
We organize the data into Train and Test splits. The Train split includes non-reference and reference (synthetic + real) data; the Test split consists of validation (synthetic + real) and real test subsets. CLIP-IQA (C-IQA), MANIQA (M-IQA), and MUSIQ columns show the average quality of generated (in the synthetic case) or filtered (in the real case) images.}
\label{tab:dataset}
\vspace{-0.5em}
\setlength{\tabcolsep}{4pt}
\renewcommand{\arraystretch}{1.15}
\resizebox{\linewidth}{!}{%
\begin{tabular}{lcccc|ccc}
\toprule
\multicolumn{8}{c}{\textbf{InScene dataset}} \\
\midrule
\textbf{Split} & \textbf{Subset} & \textbf{Category} & \textbf{\#Images} & \textbf{\#IDs} & \textbf{C-IQA$\uparrow$} & \textbf{M-IQA$\uparrow$} & \textbf{MUSIQ$\uparrow$} \\
\midrule
\multirow{4}{*}{Train}
& Non-reference dataset       & Real      & 29{,}697 & --    & 0.6935 & 0.4641 & 69.8439 \\
\cmidrule(lr){2-8}
& \multirow{2}{*}{Reference dataset} 
                             & Synthetic & 11{,}266 & 905   & 0.7191 &  0.4442  & 73.7415\\
&                            & Real      & 16{,}486 & 914   & 0.6775 & 0.4972 & 74.6676 \\
\cmidrule(lr){2-8}
& \textbf{Total (Train)}     & Mix       & \textbf{57{,}449} & \textbf{1{,}819} & \textbf{0.6967} & \textbf{0.4685} & \textbf{72.7510} \\
\midrule
\multirow{4}{*}{Test}
& InScene Validation Dataset
                             & Synthetic & 1{,}329  & 100   & 0.7183 & 0.4441 & 73.6618 \\
& InScene Validation Dataset   & Real  & 524 & 31    & 0.6435 & 0.4906  & 75.9653\\
& InScene Test Dataset   & Real  & 100  & 10 & -- & -- & -- \\
\bottomrule
\end{tabular}
}
\vspace{-1em}
\end{table}

\noindent \textbf{Training and Testing Datasets.}
One of the key challenges in our setting is obtaining HQ images of identities embedded in diverse scenes. To the best of our knowledge, no public dataset provides such identity-conditioned, scene-level data. We therefore construct both synthetic and real-world datasets.

\noindent$\,\bullet\,$\emph{Synthetic training data.}
We first build a synthetic dataset of 12{,}595 samples by sampling faces from 1{,}005 identities in the CelebRef-HQ~\cite{li2022learning} dataset and using InfiniteYou~\cite{jiang2025infiniteyou} to generate full-body images in random scenes. Generation is guided by an anti-blur LoRA 
to avoid overly blurry backgrounds, enabling diverse scenes that highlight the effectiveness of our restoration pipeline. To enforce identity consistency, we compute ArcFace~\cite{deng2019arcface} embedding similarity between each generated image and its reference images, and discard samples with identity scores below 0.55. Out of the generated samples, we use 11,266 images with 905 identities for training.

\noindent$\,\bullet\,$\emph{Real-world training data.}
For real-world training, we collect 16{,}486 gallery images from 914 identities. In addition, we incorporate human-centric, artistic photographs from the Conceptual 12M (CC12M) dataset~\cite{changpinyo2021conceptual}. We apply a multi-stage filtering process and treat the resulting CC12M subset as a non-reference training set. First, we discard images whose shorter side is smaller than 1024 pixels. Next, we crop along the longer dimension (rather than using a center crop) to ensure that the human subject remains visible in the frame. Finally, we filter images using CLIPIQA, MANIQA, and MUSIQ scores, retaining only images with CLIPIQA~$\geq 0.65$, MUSIQ~$\geq 65$, and MANIQA~$\geq 0.4$. This yields ${\sim}$57.5K training images in total.

\noindent$\,\bullet\,$\emph{Validation data.}
For the synthetic InScene validation set, we randomly sample 100 identities out of the 1{,}005 synthetic identities and apply the Real-ESRGAN~\cite{wang2021real} degradation pipeline to generate low-quality counterparts, resulting in 1{,}329 degraded samples. For the real InScene validation set, we collect gallery images from 31 previously unseen identities (524 images in total) and apply the same Real-ESRGAN degradation pipeline to synthesize LQ images.

\noindent$\,\bullet\,$\emph{Test data.}
For the real-world test set, we capture photos of 10 identities under naturally degraded conditions with a Samsung S25 Edge mobile (e.g., blur, noise, compression). For each person, we also gather clean reference images from their galleries, forming a paired real test dataset with authentic degradations and corresponding identity references (see~\autoref{tab:dataset} for details).

\noindent \textbf{Evaluation Metrics.}
For quantitative evaluation, we report results using standard reference-based pixel-wise comparison metrics, PSNR and SSIM~\cite{wang2004image}, and perceptual metrics, LPIPS~\cite{zhang2018unreasonable} and DISTS~\cite{ding2020image}. 
We also report a suite of common no-reference image quality assessment (NR-IQA) metrics, including several
single-image models
(MUSIQ~\cite{ke2021musiq}, CLIP-IQA~\cite{wang2023exploring}, MAN-IQA~\cite{yang2022maniqa},  LIQE~\cite{zhang2023blind}, 
and TOPIQ~\cite{chen2024topiq}), as well as
the distribution-based FID~\cite{heusel2017gans}.

\noindent\textbf{Implementation Details.} 

\noindent$\,\bullet\,$\textit{Stage 1: Training FaDeX.}
FaDeX is trained with contrastive learning on synthetically generated LQ--HQ face pairs.
We apply the Real-ESRGAN~\cite{wang2021real} degradation pipeline to HQ images, detect faces in the LQ images, and use FaceMe~\citep{liu2025faceme} to obtain reference-based restored HQ face of the same identity.
We additionally use a non-reference subset by pairing degraded faces with their ground-truth counterparts as HQ crops.
FaDeX is trained for 200 epochs with a learning rate of $3\times10^{-2}$ and batch size of 20 on two RTX~4090 GPUs.
The final checkpoint is then frozen and used as the degradation extractor in Stage~2.

\noindent$\,\bullet\,$\textit{Stage 2: Training the restoration model.}
For full-scene restoration, we build on top of the S3Diff~\cite{zhang2024degradation} codebase.
We use the Real-ESRGAN pipeline for the full-image degradation as well and for each clean training image pre-compute five independently degraded versions.
For every degraded version, we obtain an LQ--HQ face pair via FaceMe~\cite{liu2025faceme}; at each epoch we randomly sample one of the five precomputed pairs.
We also include the non-reference dataset by treating the ground-truth face as the HQ restored face.
Training runs for 20K iterations with an effective batch size of 64 on eight A100 GPUs and a learning rate of $2\times10^{-5}$.
SD-Turbo~\cite{sauer2024adversarial} is used as the base T2I diffusion model and is fine-tuned with our degradation-aware conditioning.
LoRA modules are inserted into the VAE encoder and U-Net with ranks 16 and 32, respectively.
The loss combines pixel, perceptual, and adversarial terms with
$\lambda_{\text{L2}} = 2.0$, $\lambda_{\text{LPIPS}} = 5.0$, and $\lambda_{\text{GAN}} = 0.5$. Please see the supplement for additional implementation details.





\vspace{-0.25em}
\subsection{Experimental Results}
\vspace{-0.25em}
\autoref{tab:synthetic_val} presents quantitative results on synthetic and real InScene validation sets. Across both validations, Face2Scene consistently achieves the best overall performance. On the synthetic benchmark, our method obtains the lowest distortion scores (DISTS and LPIPS) and the highest perceptual and no-reference quality metrics, while also delivering the strongest FID among all competing approaches. Notably, S3Diff, the second-best method, lags behind Face2Scene across nearly all perceptual criteria despite operating under a similar one-step inference budget. This demonstrates the substantial benefit of using face-derived degradation estimation instead of relying on LQ-only degradation. 

On the real validation set, which is more challenging due to the diverse degradations of natural images, Face2Scene again surpasses all baselines by a large margin. While S3Diff remains the closest competitor, its FID, distortion scores, and perceptual quality metrics are consistently inferior to ours. The improvements are particularly prominent in DISTS, LPIPS, MUSIQ, CLIP-IQA, and MAN-IQA, highlighting that our degradation code generalizes well beyond faces and provides stable and scene-consistent guidance for restoring the entire image. Collectively, these results confirm that leveraging facial references yields more accurate degradation modeling and significantly enhances full-image restoration quality. In \autoref{fig:main_result}, we present qualitative results on the synthetic and real validation sets, as well as the real test set, comparing our method with state-of-the-art models, mirroring the trends in the quantitative results where our approach achieves superior perceptual quality and fewer artifacts. Please refer to the supplementary material for additional visual comparisons.

\noindent \textbf{Model Complexity.}
Our implementation inherits the efficiency of the underlying one-step diffusion model, but incurs the additional cost of Ref-FR and degradation extraction.
Specifically, our method takes 3.3s
(stage I: 2.0s; stage II: 1.3s) to process a $1024 \times 1024$ image, which is slower than pure one-step models (OSEDiff: 1.3s; S3Diff: 1.4s), but on par with light multi-step models (PASD: 5.1s). 
However, it is much faster than heavy multi-step models
(DiffBIR: 34.6s; SUPIR: 19.9s), 
while also performing better.
See supplemental for details.

\begin{table*}[t]
\centering
\caption{\textbf{Quantitative comparison on the InScene synthetic and real validation sets.} 
Arrows indicate whether lower ($\downarrow$) or higher ($\uparrow$) values are better. 
Here, $1{+}\mathcal{N}$ denotes a single diffusion step plus $\mathcal{N}$ additional inference steps of a Ref-FR model. C-IQA and M-IQA denote CLIP-IQA and MANIQA, respectively. Each cell is color-coded to represent the \colorbox{purple}{\kern-\fboxsep best\kern-\fboxsep} and \colorbox{light}{\kern-\fboxsep second-best\kern-\fboxsep} performance.}
\vspace{-0.75em}
\label{tab:synthetic_val}
\resizebox{1\textwidth}{!}{%
\setlength{\tabcolsep}{8pt}
\begin{tabular}{l|c|ccccc|ccccc}
\toprule
\multicolumn{12}{c}{\textbf{InScene Synthetic Validation Dataset}} \\ 
\midrule
\textbf{Methods} & \textbf{Step} & \textbf{DISTS$\downarrow$} & \textbf{LPIPS$\downarrow$} & \textbf{PSNR$\uparrow$} & \textbf{SSIM$\uparrow$} & \textbf{FID$\downarrow$} & \textbf{MUSIQ$\uparrow$} & \textbf{C-IQA$\uparrow$} & \textbf{M-IQA$\uparrow$} &  \textbf{LIQE$\uparrow$} & \textbf{TOPIQ$\uparrow$} \\
\midrule
SUPIR~\cite{yu2024scaling}      & 50 & 0.1361 & 0.3123 & 24.0834 & \cellcolor{purple}{0.6293} & 24.85 & 70.2011 & 0.6015 & 0.3629 &  4.0243 & 0.6011 \\
DiffBIR~\cite{lin2024diffbir}   & 50 & 0.1831 & 0.3958 & 23.5234 & 0.5301 & 36.15 & 68.5089  & 0.6975  & \cellcolor{purple}{0.4476} & 3.3834 & \cellcolor{light}{0.6419}  \\
ResShift~\cite{yue2024efficient}  & 4 & 0.1859     & 0.3648     & \cellcolor{light}{24.6416}      & 0.5947      & 25.64    & 64.5717      & 0.6158      & 0.3493           & 2.9746      & 0.5090      \\
PASD~\cite{yang2024pixel}       & 20 & 0.1623 & 0.4014 & \cellcolor{purple}{25.4177} & 0.5644  & 21.29 & 65.3052 & 0.4745 & 0.2943 & 3.0844 & 0.4861 \\
\midrule
OSEDiff~\cite{wu2024one}        & 1  & 0.1287 & 0.2881 & 24.2400 & \cellcolor{light}{0.6057} & 22.67 & 71.0863 & 0.6634 & 0.3803 & 4.1167 & 0.6297 \\
SinSR~\cite{wu2024one}        & 1  & 0.1776 & 0.3877 & 24.1338 & 0.5575 & 26.73 & 67.4416 & \cellcolor{light}{0.7149} & 0.4189 & 3.3682 & 0.5759 \\
InvSR~\cite{yue2025arbitrary}   & 1  & 0.1318     & 0.3207     & {23.8608}      & 0.5901      & 22.92    & 71.1180      & 0.7074      & 0.4051           & 4.0872      & 0.6201      \\
S3Diff~\cite{zhang2024degradation} & 1 & \cellcolor{light}{0.1131} & \cellcolor{light}{0.2557} & {23.5955} & 0.5916 & \cellcolor{light}{18.06} & \cellcolor{light}{72.1764} & 0.6980 & 0.3858 &  \cellcolor{light}{4.4248} & 0.6233 \\
\midrule
\textbf{Face2Scene (ours)}      & $1{+}\mathcal{N}$ & \cellcolor{purple}{0.1007} & \cellcolor{purple}{0.2421} & 22.9040 & 0.5574 & \cellcolor{purple}{15.26} & \cellcolor{purple}{74.7630} & \cellcolor{purple}{0.7640} & \cellcolor{light}{0.4347} & \cellcolor{purple}{4.7157} & \cellcolor{purple}{0.6515} \\
\midrule
\multicolumn{12}{c}{\textbf{InScene Real Validation Dataset}} \\ 
\midrule
\textbf{Methods} & \textbf{Step} & \textbf{DISTS$\downarrow$} & \textbf{LPIPS$\downarrow$} & \textbf{PSNR$\uparrow$} & \textbf{SSIM$\uparrow$} & \textbf{FID$\downarrow$} & \textbf{MUSIQ$\uparrow$} & \textbf{C-IQA$\uparrow$} & \textbf{M-IQA$\uparrow$} &  \textbf{LIQE$\uparrow$} & \textbf{TOPIQ$\uparrow$} \\
\midrule
SUPIR~\cite{yu2024scaling}      & 50 & 0.2361 & 0.5563 & 17.6401 & 0.4783 & 45.22 & 70.4331 & 0.5474 & 0.4015 & 4.1305 & 0.6071 \\
DiffBIR~\cite{lin2024diffbir}   & 50 & 0.2855 & 0.6441 & 16.8264 & 0.4075 & 73.59 & 64.5391 & 0.6243 & 0.4525 & 3.0365 & 0.6063 \\
ResShift~\cite{yue2024efficient}       & 4 & 0.2802 & 0.6058 & 17.5564 & 0.4802 & 48.45 & 62.9389 & 0.6074 & 0.3833 & 3.0714 & 0.4918 \\
PASD~\cite{yang2024pixel}       & 20 & 0.2483 & 0.5786 & \cellcolor{light}{18.0514} & \cellcolor{light}{0.5221} & 49.90 & 70.5563 & 0.5390 & 0.3903 & 3.7782 & 0.5881 \\
\midrule
OSEDiff~\cite{wu2024one}        & 1  & 0.2355 & 0.5393 & 17.6212 & 0.5078 & 51.10 & 73.1528 & 0.6269 & 0.4346 & 4.4328 & 0.6569 \\
SinSR~\cite{wu2024one}        & 1  & 0.2624 & 0.6122 & 17.6535 & 0.4677 & 49.44 & 67.0641 & \cellcolor{light}{0.6990} & 0.4422 & 3.4859 & 0.5671 \\
InvSR~\cite{yue2025arbitrary}   & 1  & 0.2311 & 0.5467 & 17.4140 & 0.4916 & 49.35 & 72.9144 & 0.6911 & \cellcolor{light}{0.4569} & 4.4271 & 0.6441 \\
S3Diff~\cite{zhang2024degradation} & 1 & \cellcolor{light}{0.2231} & \cellcolor{light}{0.5149} & 17.1439 & 0.4894 & \cellcolor{purple}{38.64} & \cellcolor{light}{73.8209} & 0.6734 & 0.4480  & \cellcolor{light}{4.7060} & \cellcolor{light}{0.6627} \\
\midrule
\textbf{Face2Scene (ours)}      & $1{+}\mathcal{N}$ & \cellcolor{purple}{0.1178} & \cellcolor{purple}{0.2502} & \cellcolor{purple}{22.8975} & \cellcolor{purple}{0.6197} & \cellcolor{light}{42.21} & \cellcolor{purple}{75.3739} & \cellcolor{purple}{0.7015} & \cellcolor{purple}{0.4714} &  \cellcolor{purple}{4.8044} & \cellcolor{purple}{0.6777} \\
\bottomrule
\end{tabular}}
\end{table*}


\subsection{Ablation Studies}

\noindent \textbf{Effectiveness of extracting degradation as a cue for the restoration.}
To show the impact of our degradation extraction module (i.e., FaDeX and MapNet, shown in the ``top branch'' of~\autoref{fig:main}, right inset for stage II), which operates on the LQ-HQ face patch pair, we ablate it from our network. The resulting model takes in only the full LQ image.~\autoref{tab:gt_face_inserted} shows that our model manages to pull useful degradation information from the LQ-HQ pair.

\begin{table*}[t]
\centering
\caption{\textbf{Scene and face restoration analysis.} “\textit{GT Face Inserted}” indicates we composite the ground-truth face region into the restored image to isolate each method’s impact on the rest of the scene. 
The two rows (“\textit{Face only}”) report metrics computed only within the facial region (without GT insertion). LIQE for face only is not reported because the model is sensitive to input size and cannot handle small face sizes. The last two rows show our method with and without the proposed degradation estimation.
The best value per metric is \textbf{bolded}.}
\label{tab:gt_face_inserted}
\vspace{-0.5em}
\resizebox{\textwidth}{!}{
\setlength{\tabcolsep}{8pt}
\begin{tabular}{l|ccccc|ccccc|c}
\toprule
\multicolumn{12}{c}{\textbf{InScene Synthetic Validation Dataset}} \\ 
\midrule
\textbf{Methods} 
& \textbf{DISTS$\downarrow$} 
& \textbf{LPIPS$\downarrow$} 
& \textbf{PSNR$\uparrow$} 
& \textbf{SSIM$\uparrow$} 
& \textbf{FID$\downarrow$} 
& \textbf{MUSIQ$\uparrow$} 
& \textbf{C-IQA$\uparrow$} 
& \textbf{M-IQA$\uparrow$} 
& \textbf{LIQE$\uparrow$} 
& \textbf{TOPIQ$\uparrow$} 
& \textbf{Wins$\uparrow$} \\
\midrule
S3Diff (\textit{GT Face Inserted})~\cite{zhang2024degradation}  
& 0.1000 & 0.2392 & \textbf{24.0985} & \textbf{0.6231} & 16.05 
& 72.5181 & 0.6991 & 0.3933 & 4.4730 
& 0.6441 & \textbf{2/10} \\

\textbf{Face2Scene (\textit{GT Face Inserted})}
& \textbf{0.0908} & \textbf{0.2268} & 23.3759 & 0.5904 & \textbf{13.86} 
& \textbf{74.4256} & \textbf{0.7591} & \textbf{0.4368}
& \textbf{4.6983} & \textbf{0.6593} & \textbf{8/10} \\
\midrule
S3Diff (\textit{Face only})
& 0.1313 & 0.1510 & \textbf{23.3827} & \textbf{0.6268} & 26.96 
& 64.4375 & 0.7362 & 0.4942 
& -- & 0.6984 & \textbf{2/9} \\
\textbf{Face2Scene (\textit{Face only})}
& \textbf{0.1201} & \textbf{0.1446} & 23.0089 & 0.6118 & \textbf{18.50} 
& \textbf{67.0781} & \textbf{0.7606} & \textbf{0.5398} 
& -- & \textbf{0.7237} & \textbf{7/9} \\
\midrule
\textbf{Face2Scene \textit{w} Degradation Estimation} & \textbf{0.1007} & \textbf{0.2421} & \textbf{22.9040} & \textbf{0.5574} & \textbf{15.26} & \textbf{74.7630} & \textbf{0.7640} & \textbf{0.4347} & \textbf{4.7157} & \textbf{0.6515} & \textbf{10/10}\\
\textbf{Face2Scene \textit{w/o} Degradation Estimation}
& 0.1293 & 0.2711 & 22.7160 & 0.5476 & 17.81
& 71.7807 & 0.7230 & 0.4234
& 4.3540 & 0.6265 & \textbf{0/10} \\
\bottomrule
\end{tabular}
}
\vspace{-0.75em}
\end{table*}

\begin{figure}[t] 
    \centering
    \includegraphics[width=0.49\linewidth]{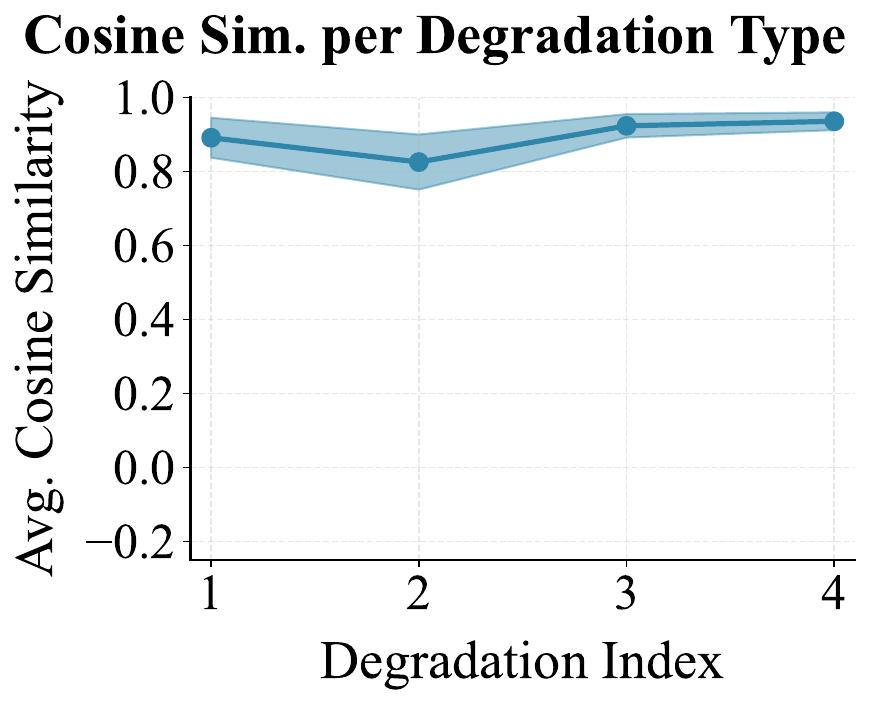}\hfill\includegraphics[width=0.50\linewidth]{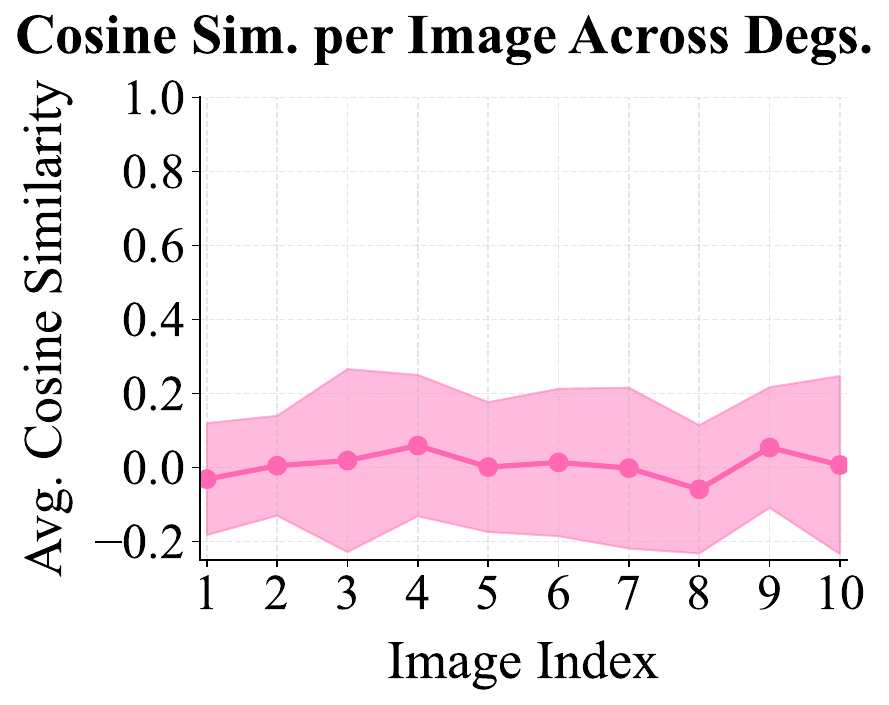}
    \vspace{-0.6em}
    \caption{\textbf{Cosine similarity analysis.}
    We show the cosine similarity across embeddings of image pairs with different degradations.
    (Left) similarities per degradation type (averaged over images).
    (Right) similarities per image, averaged over degradation types.
    Shaded area shows standard deviation.
    This confirms FaDeX isolates degradation from image content.
    }
    \vspace{-1em}
    \label{fig:avg_cossim}
\end{figure}

\noindent \textbf{What do we learn from FaDeX?}
Ideally, FaDeX would encode degradations \textit{without} including information about image content.
To evaluate this, we apply four degradations (different manually chosen RealESRGAN settings \cite{wang2021real}; see supplement for details) on 10 random images and observe the cosine similarity between them on the output of the FaDeX. \autoref{fig:avg_cossim} shows the average cosine similarity between different images using the same degradation (left) and the average cosine similarity across the same image using different degradations (right). The graphs clearly show FaDeX concentrates dominantly on degradations, while minimizing extraction of image content.

\noindent \textbf{Impact of identity references on scene-level restoration.}
Since we have access to restored faces via Ref-FR, we verify that our improvements are \textit{not} only due to those faces alone, but rather come from improvements across the scene (including non-face areas).
To evaluate this, we replaced the faces in the output restored images with their GT counterparts.
This removes or reduces the impact of the restored faces themselves, meaning the resulting metrics should reflect ``non-face'' quality and fidelity.
To prevent blending artifacts due to this insertion, we make the transition on the edges gradual: using a face mask $M$, with the center being one and transitioning to zero at the edges, we alter the model prediction $P$ to be 
$M \odot I_\mathrm{GT} + (1-M) \odot P$, 
with $\odot$ being an element-wise product. 
When applied to our model and S3Diff (the closest model to ours in both performance and architecture),
there is a clear improvement of our model on the entire image, excluding the face region, as reported in ~\autoref{tab:gt_face_inserted}.


\vspace{-0.5em}
\section{Discussion}
\vspace{-0.5em}
\noindent\textbf{Limitations.} One limitation of our approach is the assumption of a unified degradation across the entire image. While this assumption is reasonable when considering many artifacts, including noise and JPEG compression, in other cases, it may not hold. 
For instance, blur can be incurred in a spatially varying manner, due to depth-of-field or localized object motion.
One way to address this might be to add a pixel or depth-dependent blur estimation module. As this is an extension of our approach, we leave it for future work.

\noindent\textbf{Conclusions.}
In this paper, we demonstrated that reference-based face restoration can be leveraged as an oracle to first extract the degradations of a scene and then guide a one-step diffusion model to recover the full-scene image, including the body and background. 
In particular, by leveraging both the LQ and the reconstructed HQ face, we designed a degradation estimation module that isolates degradation information from image content, and extracts multi-scale restoration codes, which guide the diffusion model on the reconstruction of the entire scene, including outside of the face regions.
Our results show that such an approach outperforms existing blind scene restoration models, even those that perform degradation estimation (via the LQ input alone), validating the utility of our reference-based guidance.

\clearpage
\twocolumn[\section*{}]
\noindent
\begin{minipage}{\textwidth}
    \centering
    \tableofcontents  
    \thispagestyle{empty}
\end{minipage}

\setcounter{page}{1}
\maketitlesupplementary

\section{Datasets}

In this section, we provide details regarding the datasets considered in our paper:
(i) the synthetic dataset generated by InfiniteYou,
(ii) real gallery images (synthetically degraded for evaluation), and
(iii) the real-world mobile-captured data, with real-world degradations.

\subsection{InScene Synthetic Validation Dataset}
\begin{figure}[t]
    \centering
    \includegraphics[width=1\linewidth]{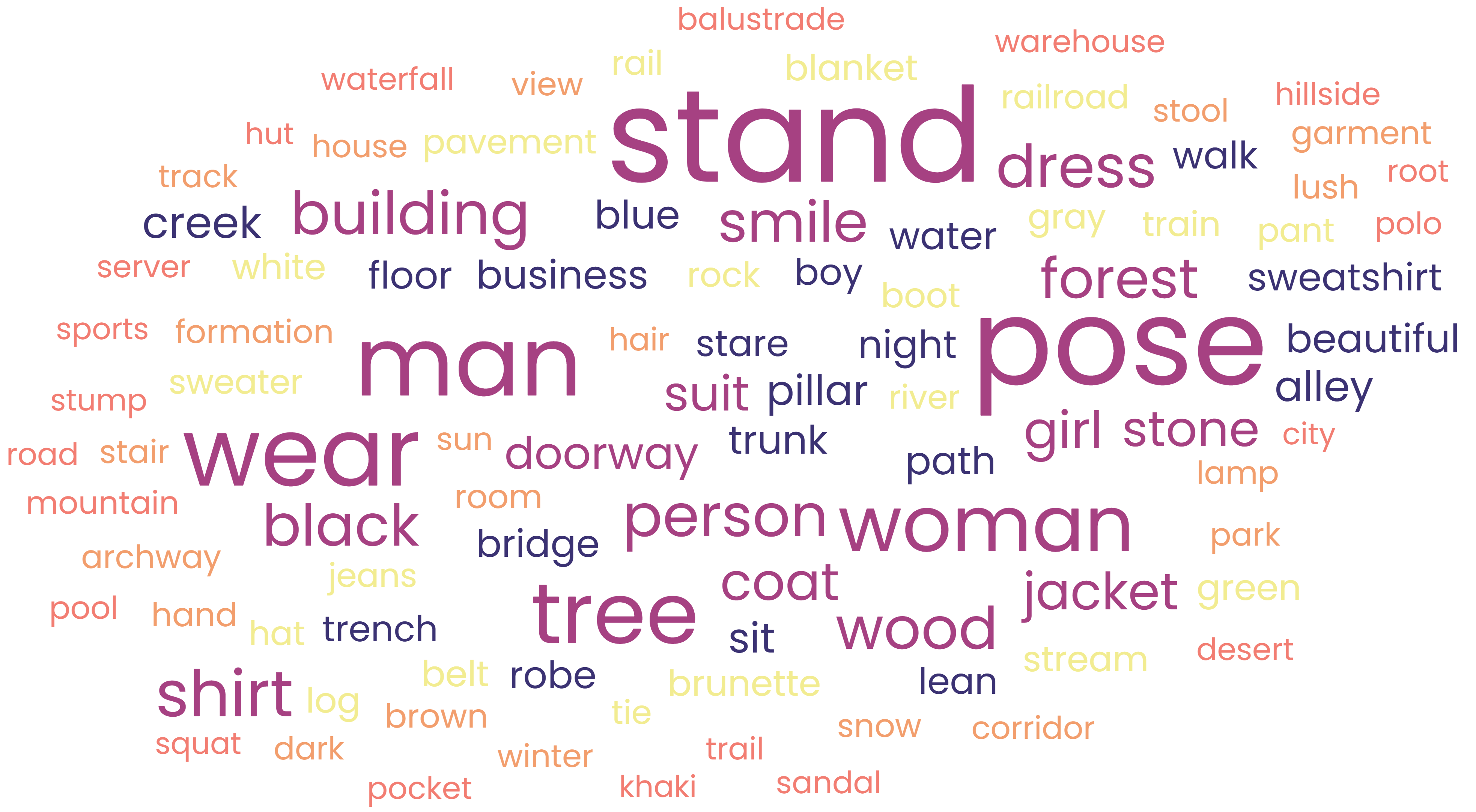}
    \caption{\textbf{Scene diversity visualization.} Word-cloud representation of the semantic tags extracted from our synthetic dataset using the RAM++~\cite{zhang2024recognize} image-tagging model. Larger words indicate tags that occur more frequently across generated images, highlighting the broad diversity of scenes captured in our dataset.}
    \label{fig:word_cloud}
\end{figure}

We first construct a synthetic dataset of 12{,}595 samples by sampling faces from 1{,}005 identities in the CelebRef-HQ~\cite{li2022learning} dataset and using InfiniteYou~\cite{jiang2025infiniteyou} to generate full-body images in diverse scenes. We use the \texttt{Stage~1} weights from InfiniteYou for generation, as they provide stronger identity preservation compared to \texttt{Stage~2}. Generation is guided by an anti-blur LoRA\footnote{\url{https://huggingface.co/Shakker-Labs/FLUX.1-dev-LoRA-AntiBlur}} to reduce excessive background blur, resulting in varied and visually rich scenes that better highlight the effectiveness of our restoration pipeline. 
The diversity of these scenes is illustrated via a word-cloud visualization in~\autoref{fig:word_cloud}, where semantic tags are extracted from each image using the RAM++~\cite{zhang2024recognize} image tagging model.

\begin{figure*}[t]
    \centering
    \includegraphics[width=1\textwidth]{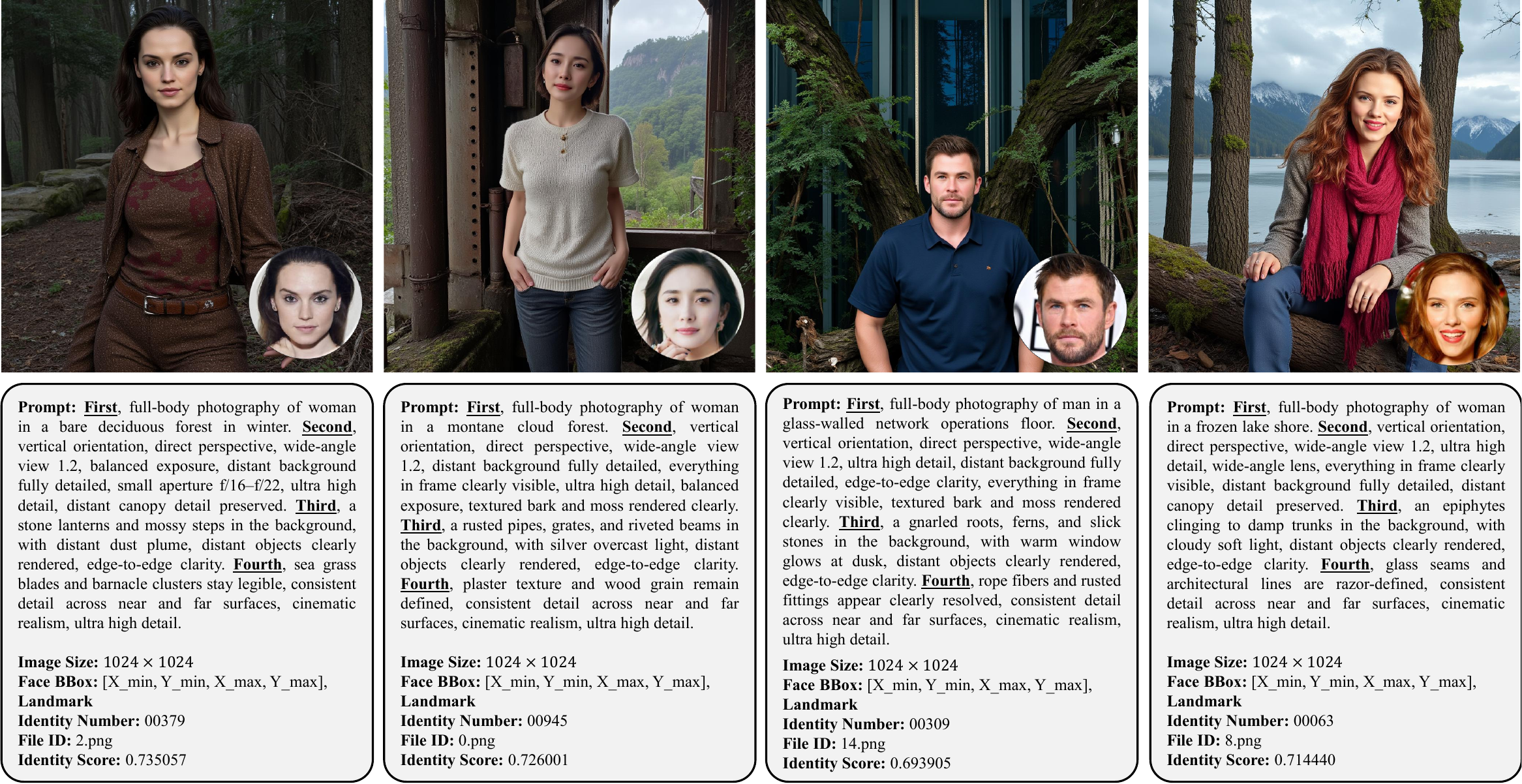}
    \caption{\textbf{Four representative images of our synthetic InScene dataset.} Each sample is shown alongside its structured prompt. We also show the stored metadata, including facial landmarks, bounding boxes, and associated identity labels.}
    \label{fig:sample_inf_dataset}
\end{figure*}

As illustrated in~\autoref{fig:sample_inf_dataset}, we also provide example prompts: the \textit{first} and \textit{second} sentences describe the subject, while the \textit{third} and \textit{fourth} sentences describe the background in detail to increase the chance of generating highly detailed scene images.

To ensure identity consistency, we compute the ArcFace~\cite{deng2019arcface} embedding similarity between each generated image and its corresponding reference images, and discard samples with identity scores below 0.55. After filtering, we retain 11{,}266 images from 905 identities for training and 100 identities for the evaluation. For each generated sample, as shown in~\autoref{fig:sample_inf_dataset}, we store metadata including the \textit{prompt}, \textit{image resolution}, \textit{detected face bounding box}, \textit{facial landmarks}, the \textit{File ID} (identity index in CelebRef-HQ~\cite{li2022learning}), and the ArcFace \textit{identity similarity score} between the generated face and the reference identity image used during generation.

\subsection{InScene Real Validation Dataset}
\begin{figure*}[t]
    \centering
    \includegraphics[width=1\textwidth]{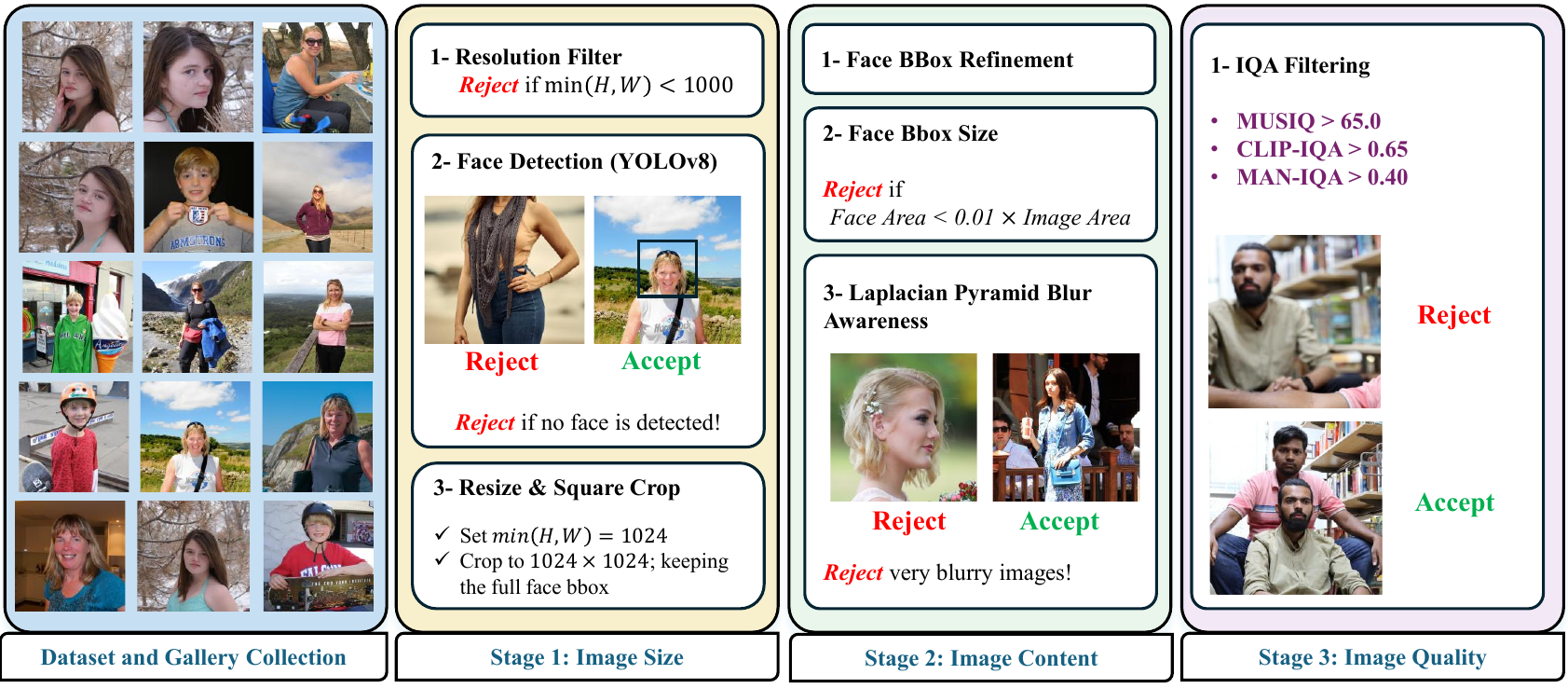}
    \vspace{-0.25em}
    \caption{\textbf{Real dataset preprocessing pipeline.} 
Our real-world images are standardized through a three-stage filtering process. 
\textbf{Stage 1} removes images with insufficient resolution and ensures a detectable face is present. 
\textbf{Stage 2} enforces content quality by verifying that the face region is sufficiently large and the image is not overly blurred 
(e.g., due to defocus blur in the background, as in the example image shown above). 
\textbf{Stage 3} applies no-reference image quality metrics (MUSIQ, CLIP-IQA, and MANIQA) to retain only high-quality samples.} 
\vspace{-1em}
    \label{fig:preprocess}
\end{figure*}

As illustrated in \autoref{fig:preprocess}, we preprocess the real-world gallery and evaluation images using a three-stage pipeline that standardizes resolution, content, and perceptual quality before training and evaluation.

\noindent\textbf{Stage 1: Image size.}
We first apply a resolution filter and discard any image whose shorter side is smaller than 1000 pixels, ensuring sufficient spatial detail for both face and scene restoration. We then run a YOLOv8-based \cite{redmon2016you} face detector\footnote{\url{https://github.com/YapaLab/yolo-face}} and reject images in which no face is detected. For the remaining images, we normalize the spatial support by resizing such that $\min(H, W) = 1024$ and cropping to a $1024 \times 1024$ square while preserving the full detected face bounding box.

\noindent\textbf{Stage 2: Image content.}
Next, we refine the face bounding box and measure its area relative to the full image. Images are discarded if the face occupies less than $1\%$ of the image area, which removes cases where the subject is too small or dominated by background. Following prior work \cite{gong2025human}, we also apply a blur measure based on the variance of the Laplacian, to identify overly blurred samples and reject images that fall below a blur-awareness threshold, keeping only images with sufficiently sharp facial and scene details.

\noindent\textbf{Stage 3: Image quality.}
In the final stage, we perform no-reference image quality assessment using three metrics: 
MUSIQ \cite{ke2021musiq}, CLIP-IQA \cite{wang2023exploring}, and MANIQA \cite{yang2022maniqa}. We retain only images that satisfy MUSIQ $\geq 65.0$, CLIP-IQA $\geq 0.65$, and MANIQA $\geq 0.40$, filtering out low-quality or heavily degraded samples. This pipeline yields a curated real dataset with consistent resolution, reliable face content, and high perceptual quality, suitable for robust training and evaluation of our restoration framework.

\subsection{InScene Real Test Dataset.}
We captured 100 real-world test images from 10 different identities using a Samsung S25 Edge smartphone. All images were recorded in RAW format with automatic exposure and ISO values set to 3200, 1600, and 800 to emulate diverse real-world degradations. To further introduce motion-related degradations, we additionally captured a subset of images with a \emph{motion blur}. We then processed the RAW files using the \texttt{rawpy} library\footnote{\url{https://github.com/letmaik/rawpy}} to obtain RGB images. These processed RGB images serve as the real degraded inputs for evaluating our model under realistic capture conditions.

\section{Implementation Details}
\begin{figure*}[t]
    \centering
    \includegraphics[width=0.9\textwidth]{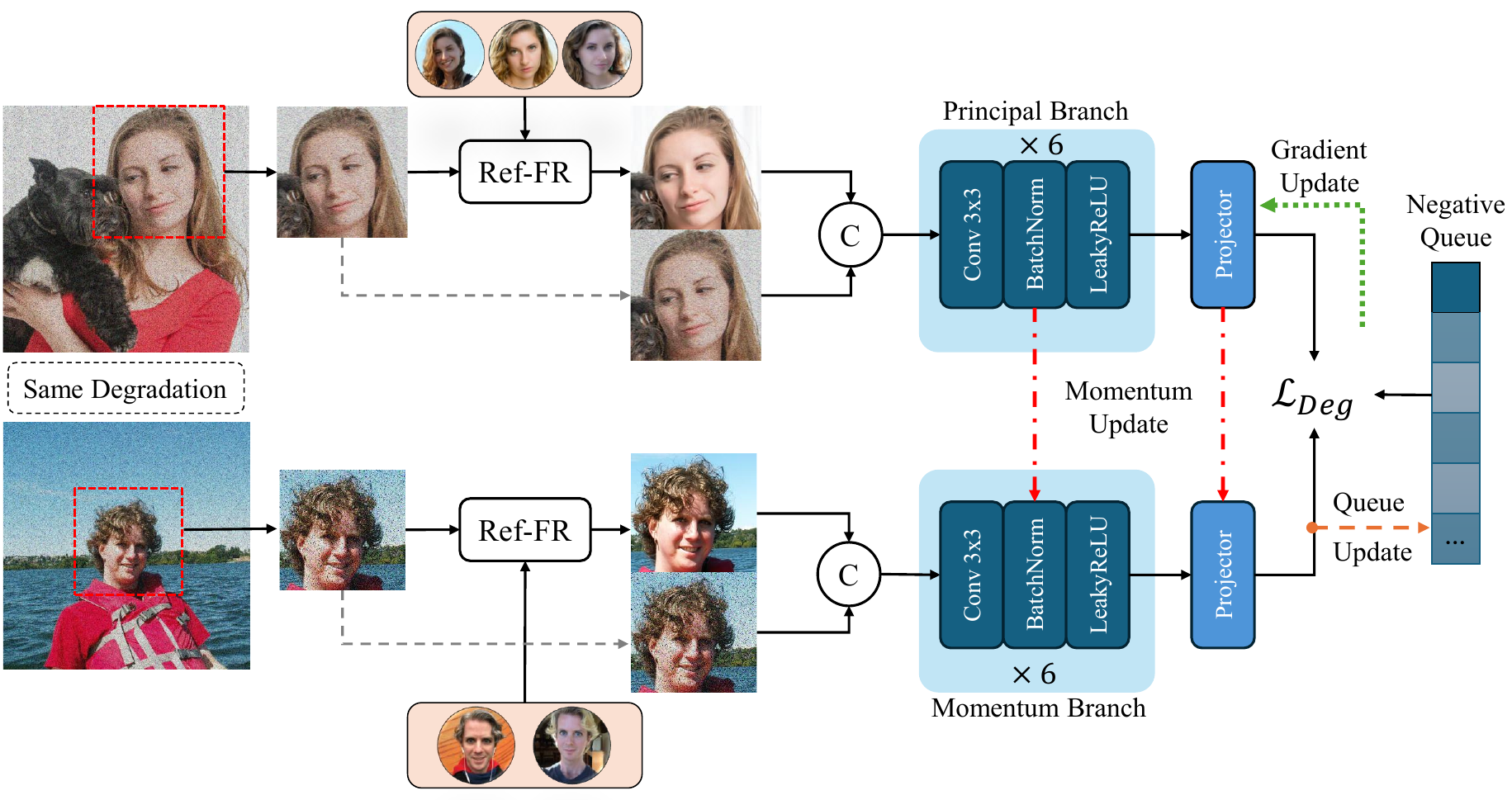}
    \caption{\textbf{Overview of FaDeX contrastive learning.} Momentum-encoder contrastive training setup with dual encoders and a queue of negatives to learn degradation-discriminative face embeddings.}
    \label{fig:fadex_impl}
\end{figure*} 
\subsection{FaDeX Details}
Figure~\ref{fig:fadex_impl} illustrates the contrastive training setup used for FaDeX, which follows a momentum-encoder design similar to MoCo \cite{he2020momentum}.
Given HQ--LQ face pairs produced by the Ref-FR module, we feed two independent inputs of the concatenated faces, where they share a similar degradation, into a \emph{principal} encoder branch and a \emph{momentum} encoder branch.
Both branches share the same architecture: a stack of $3{\times}3$ convolution, BatchNorm \cite{ioffe2015batch}, and LeakyReLU blocks followed by a projection head (Linear + LeakyReLU + Linear). 
The principal branch is updated by standard backpropagation, while the momentum branch is an exponential moving average of the principal encoder parameters and is never directly optimized. The output of each branch is passed through the projector to obtain $d$-dimensional face-degradation embeddings.
Embeddings from the momentum branch are enqueued into a first-in–first-out \emph{negative queue} that maintains a large, constantly refreshed set of past samples.
At each iteration, for a given sample $i$ in the current mini-batch, the embedding from the principal branch acts as the query, $q_i$, embeddings from other samples with the same degradation operator, $\mathcal{G}$, form the positive set, and embeddings drawn from the queue (and non-matching samples in the batch) serve as negatives.
The contrastive degradation loss, $\mathcal{L}_{\mathrm{Deg}}$, is then computed over this large set of negatives.
After the gradient update on the principal branch, we update the momentum branch via EMA and refresh the negative queue by enqueuing the current momentum embeddings and dequeuing the oldest ones.
This implementation stabilizes training, enlarges the pool of negative degradations without increasing batch size, and yields robust, degradation-discriminative FaDeX features that we subsequently freeze and use as conditioning for the one-step restorer.

\subsection{Additional One-Step Diffusion Model Implementation Details}
During training, we adopt an online negative prompt learning scheme inspired by previous work~\cite{zhang2024degradation}. For each mini-batch, with probability, \(p_n\), we replace the ground-truth HR target with its synthesized LR counterpart to form a negative target, while the remaining samples keep their original HR targets as positives. We pair negative targets with a fixed negative prompt (e.g., ``oil painting, cartoon, blur, dirty, messy, low quality, deformation, low resolution, oversmooth'') and positive targets with a generic high-quality prompt (``a high-resolution, 8K, ultra-realistic sharp image, vibrant colors, and natural lighting''). During inference, we apply classifier-free guidance \cite{ho2022classifier} with both prompts: the VAE encoder, \(\mathbf{E}_\theta\), and UNet denoiser, \(\epsilon_\theta\), produce features for the positive and negative prompts, \(z_{\text{pos}} = \epsilon_\theta(\mathbf{E}_\theta(I_{\text{LR}}), t_{\text{pos}})\) and \(z_{\text{neg}} = \epsilon_\theta(\mathbf{E}_\theta(I_{\text{LR}}), t_{\text{neg}})\), and the final representation is obtained as
\[
z_{\text{out}} = z_{\text{neg}} + \lambda_{\text{cfg}}(z_{\text{pos}} - z_{\text{neg}}),
\]
where \(\lambda_{\text{cfg}}\) is the guidance scale. This teaches the model to associate negative prompts with low-quality artifacts and positive prompts with desired restoration quality, while reusing synthesized LR images and thus adding no extra overhead to the training pipeline. We use \(\lambda_{\text{cfg}} = 1.10\) in all our experiments.

\noindent\textbf{Degradation Attention Implementation}
We implement the degradation attention module, $\mathrm{DegAttn}$ with eight heads, $512$ input channels, and $1024$-dimensional output tokens that match the cross-attention embedding size of the diffusion model.
Starting from the FaDeX feature map $Z_{\text{face}} \in \mathbb{R}^{B\times 256\times H\times W}$, we first apply an overlapping patch embedding, which halves the spatial resolution and expands the channels to $512$.
We then split this tensor along the channel dimension into two branches, $F_1$ and $F_2$, and flatten them into token sequences that serve as the two inputs to $\mathrm{DegAttn}$. Inside $\mathrm{DegAttn}$, we build separate queries and keys from $F_1$ and $F_2$, each using half of the eight heads (four heads per branch), yielding two self-attention maps $A_1$ and $A_2$.
Values are computed from the concatenation $[F_1; F_2]$, so that each head operates on a richer $2d_h$ value vector that jointly encodes both streams.
A learnable scalar, $\lambda$, controls how much the second branch’s attention should be subtracted from the first. The result $Y_{\text{ch}}$ is normalized with RMSNorm and projected back to $256$ channels. From $Y_{\text{ch}}$, we derive multi-scale degradation tokens using three pooling stages described in the method section.

\subsection{Reference-Based Face Restoration Model Selection} Among the public and open-source reference-based restoration models that outperform simple restoration models, we found InstantRestore~\cite{zhang2025instantrestore}, FaceMe~\cite{liu2025faceme}, Ref-LDM~\cite{hsiao2024ref}, and RestorerID~\cite{ying2024restorerid}. Unfortunately, most methods are not publicly available. Among these, FaceMe gave the best results while being efficient at inference. Therefore, we adopt FaceMe as our Ref-FR model to generate a high-quality face given reference images.

\subsection{FaDeX Cosine-Similarity Experiment}
In the paper, we evaluate whether FaDeX encodes degradations while being invariant to image content.
To this end, we apply four fixed RealESRGAN-style degradation presets to 10 randomly selected images and compute cosine similarities in the FaDeX embedding space.
We denote these degradation settings by $d_1,\dots,d_4$ and keep all hyperparameters deterministic, so each $d_i$ corresponds to a reproducible degradation operator.
Each preset uses the standard two-stage RealESRGAN pipeline.

\begin{itemize}
    \item \noindent\textbf{$d_1$: strong downscale + strong blur + high JPEG.}
    Stage~one applies a strong downscale, isotropic Gaussian blur, gray Gaussian noise, Poisson noise, and high-quality JPEG compression. Stage~two keeps the resolution, applies a milder isotropic blur and very high JPEG quality, followed by a final sinc filter.
    This yields heavily smoothed, downsampled images with strong blur and mild JPEG artifacts.
    \item \noindent\textbf{$d_2$: strong upscale + low JPEG, then downscale + anisotropic blur.}
    Stage~one strongly upscales the image, adds colored Poisson noise, anisotropic Gaussian blur, and low JPEG quality. Stage~two downsamples the image, adds additional Poisson noise, applies anisotropic blur, and uses low–mid JPEG quality, without a final sinc filter. Overall, $d_2$ produces moderately blurred images with strong Poisson noise and noticeable JPEG artifacts.
    \item \noindent\textbf{$d_3$: almost-clean identity with sinc filtering.}
    Both stages keep the original resolution and use maximum JPEG quality. Stage~one has no Gaussian noise and minimal Poisson noise with a sinc filter; Stage~two adds only a very small Gaussian noise and another sinc filter.
    This preset produces nearly artifact-free images with very mild sharpening/sinc effects and negligible noise.
    \item \noindent\textbf{$d_4$: moderate downscale + light noise + very low JPEG, then upscale + plateau blur.}
    Stage~one moderately downsamples the image, adds light colored Gaussian noise, mild Poisson noise, plateau-anisotropic blur, and very low JPEG quality. Stage~two then upscales the image, applies plateau-anisotropic blur again, very low JPEG quality, and a final sinc filter. This yields images with structured blur, strong JPEG artifacts, and a slight scale inconsistency.
\end{itemize}

\section{Experimental Results}

\subsection{Quantitative Results: InScene Test Dataset}

{
In~\autoref{tab:realworld}, we compare models on the InScene real test dataset captured by the Samsung S25 Edge phone. As there is no ground-truth image in this case, we report on commonly used non-reference image quality assessment metrics (MUSIQ, Clip-IQA, MANIQA, LIQE, and TOPIQ). The results show that our approach compares favorably to other baselines, being either best or second-best according to four of the five models. In particular, compared to the second best model in the table, DiffBIR, our model outputs sharper images with more details, as observed in Figures~\ref{fig:real_test_samples},
\ref{fig:real_test_samples_triple}, and
\ref {fig:real_test_samples_174_0}.
}

\begin{table}[t]
    \centering
    \vspace{-0.5em}
    \caption{\textbf{Quantitative comparison on the InScene Test Dataset.} 
    Since real-world photographs do not have a ground-truth image against which to compare,
        we quantitatively evaluate on our InScene Test data via no-reference IQA models.
Arrows indicate whether lower ($\downarrow$) or higher ($\uparrow$) values are better. C-IQA and M-IQA denote CLIP-IQA and MANIQA, respectively. Each cell is color-coded to represent the \colorbox{purple}{\kern-\fboxsep best\kern-\fboxsep} and \colorbox{light}{\kern-\fboxsep second-best\kern-\fboxsep} performance.}
    \vspace{-0.5em}
    \resizebox{\linewidth}{!}{%
    \begin{tabular}{l|ccccc}
    \toprule
    \multicolumn{6}{c}{\textbf{InScene Test Dataset}} \\ 
    \midrule
    \textbf{Methods} 
    & \textbf{MUSIQ$\uparrow$} 
    & \textbf{C-IQA$\uparrow$} 
    & \textbf{M-IQA$\uparrow$} 
    & \textbf{LIQE$\uparrow$} 
    & \textbf{TOPIQ$\uparrow$} \\ 
    \midrule
    SUPIR~\cite{yu2024scaling}      & 60.4119 & 0.3506 & 0.3108 & 2.8347 & 0.4210  \\
    DiffBIR~\cite{lin2024diffbir}   & 71.4031 & \cellcolor{purple}{0.6701} & \cellcolor{purple}{0.5238} & 4.3273 & \cellcolor{light}{0.6751} \\
    ResShift~\cite{yue2024efficient}  & 65.8328 & 0.5414 & 0.3827 & 3.7667 & 0.4916 \\
    PASD~\cite{yang2024pixel}       & 71.6666 & 0.5920 & 0.4849 & 4.3644 & 0.6483 \\
    \midrule
    OSEDiff~\cite{wu2024one}        & 72.6963 & 0.6327 & 0.4819 & \cellcolor{light}{4.4864} & 0.6583 \\
    SinSR~\cite{wang2024sinsr}        & 71.6175 & 0.5965 & \cellcolor{light}{0.4947} & 4.3157 & 0.6746 \\
    InvSR~\cite{yue2025arbitrary}   & \cellcolor{light}{72.7069} & 0.4559 & 0.4145 & 4.3888 & 0.6665 \\
    S3Diff~\cite{zhang2024degradation} & 69.4941 & 0.5423 & 0.4014 & 4.1888 & 0.5805 \\
    \midrule
    \textbf{Face2Scene} & \cellcolor{purple}{73.3047} & \cellcolor{light}{0.6407}  & 0.4859 & \cellcolor{purple}{4.6973} & \cellcolor{purple}{0.6846} \\
    \bottomrule
    \end{tabular}
    }
\label{tab:realworld}
\vspace{-0.25em}
\end{table}

\subsection{Qualitative Results}

\subsubsection{InScene Synthetic and Real Validation Datasets}
\begin{figure*}[htb]
    \centering
    \newcommand{\cropheight}{0.20\textwidth}
    \resizebox{0.92\textwidth}{!}{
\begin{tabular}{@{} c *{6}{c} @{}}

    \rotatebox{90}{LQ} &
    \includegraphics[width=0.20\textwidth]{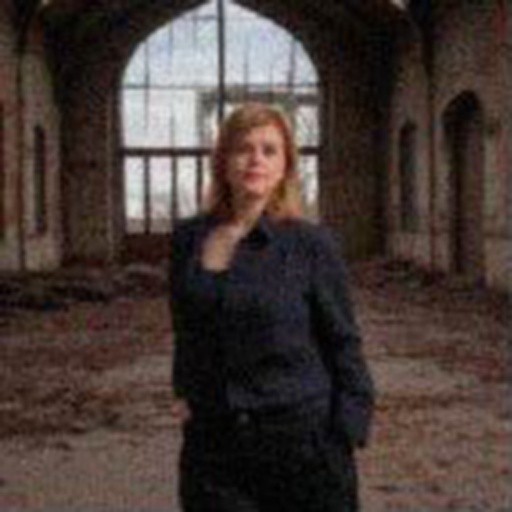} &
    \includegraphics[height=\cropheight]{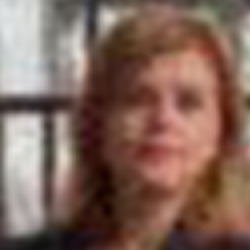} &
    \includegraphics[height=\cropheight]{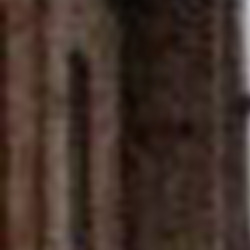} &
    \includegraphics[width=0.20\textwidth]{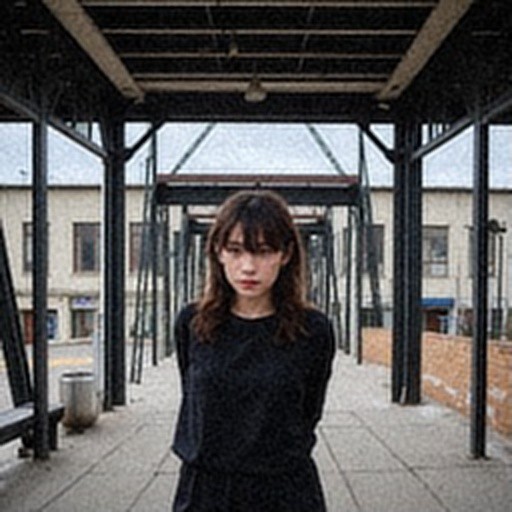} &
    \includegraphics[height=\cropheight]{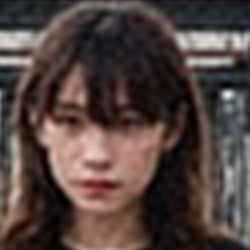} &
    \includegraphics[height=\cropheight]{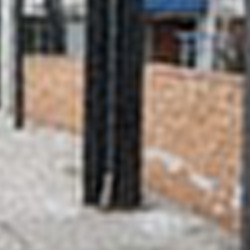}  \\

    \rotatebox{90}{GT} &
    \includegraphics[width=0.20\textwidth]{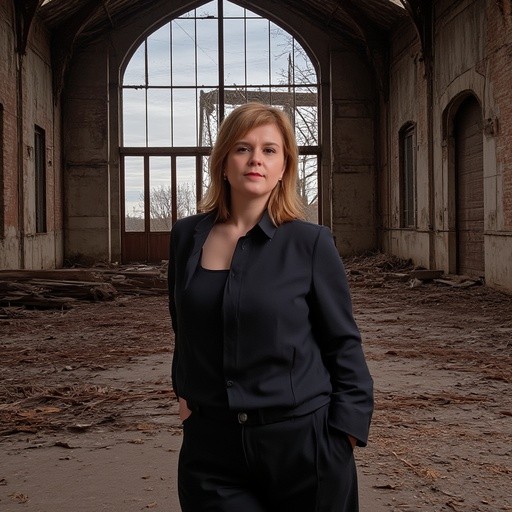} &
    \includegraphics[height=\cropheight]{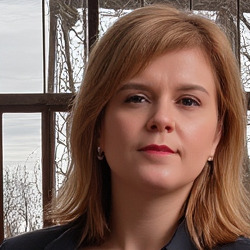} &
    \includegraphics[height=\cropheight]{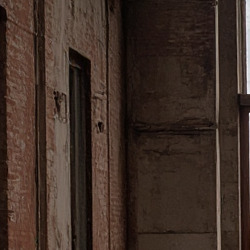} &
    \includegraphics[width=0.20\textwidth]{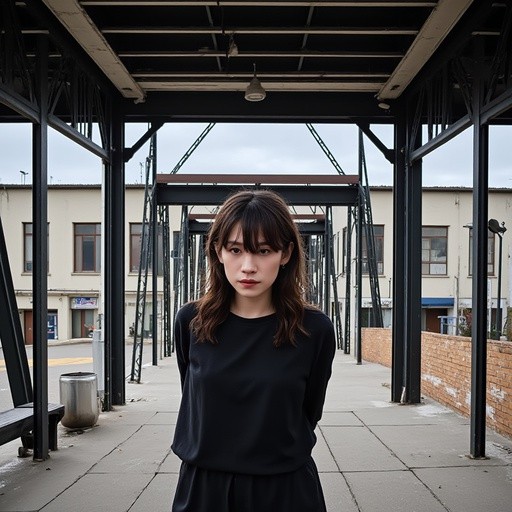} &
    \includegraphics[height=\cropheight]{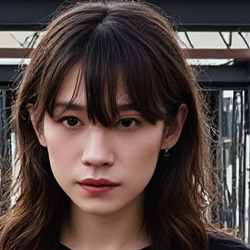} &
    \includegraphics[height=\cropheight]{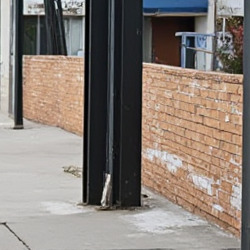}  \\
\end{tabular}
    }
    \caption{\textbf{Visual comparison on InScene Synthetic dataset across seven methods.}}
    \label{fig:syn_visual_results_part1}
\end{figure*}

\begin{figure*}[htb]
    \centering
    \newcommand{\cropheight}{0.20\textwidth}
    \resizebox{0.88\textwidth}{!}{
    \begin{tabular}{@{} c *{6}{c} @{}}

    \rotatebox{90}{LQ} &
    \includegraphics[width=0.20\textwidth]{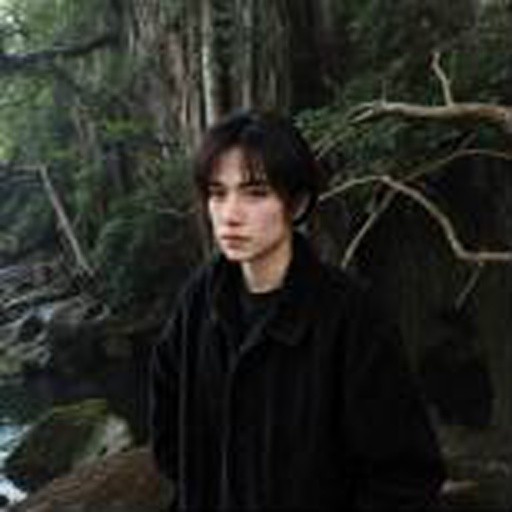} &
    \includegraphics[height=\cropheight]{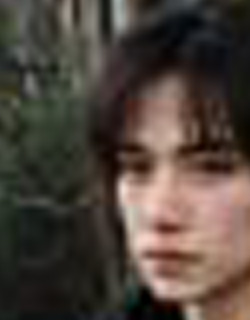} &
    \includegraphics[height=\cropheight]{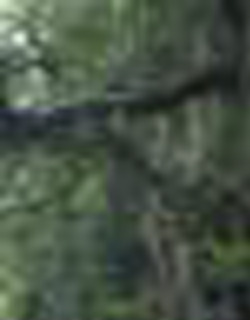} &
    \includegraphics[width=0.20\textwidth]{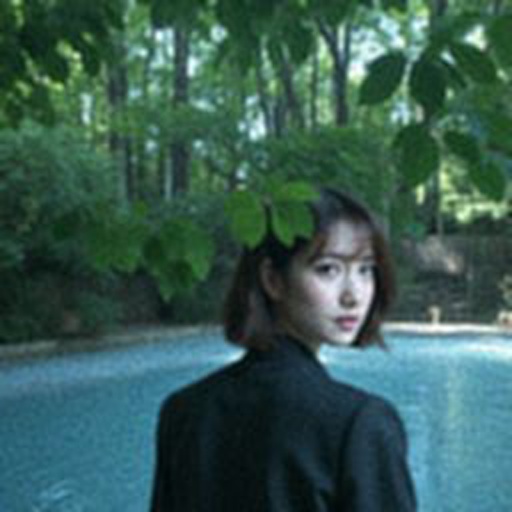} &
    \includegraphics[height=\cropheight]{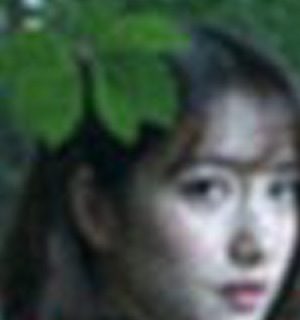} &
    \includegraphics[height=\cropheight]{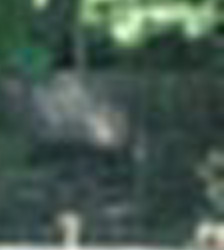}  \\

    \rotatebox{90}{GT} &
    \includegraphics[width=0.20\textwidth]{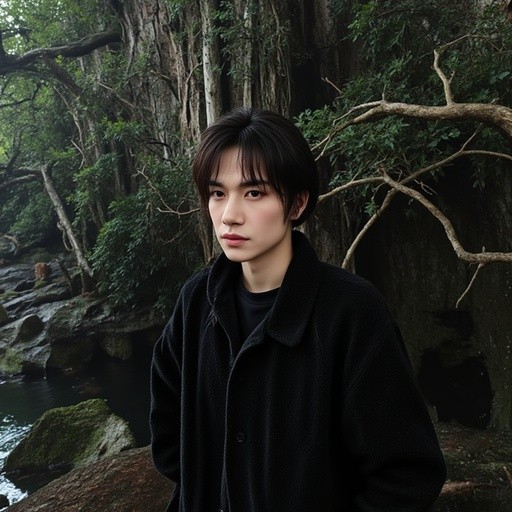} &
    \includegraphics[height=\cropheight]{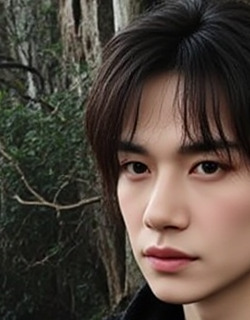} &
    \includegraphics[height=\cropheight]{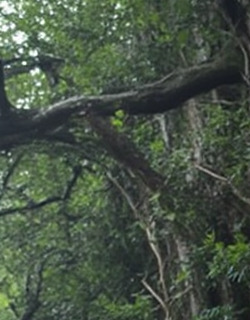} &
    \includegraphics[width=0.20\textwidth]{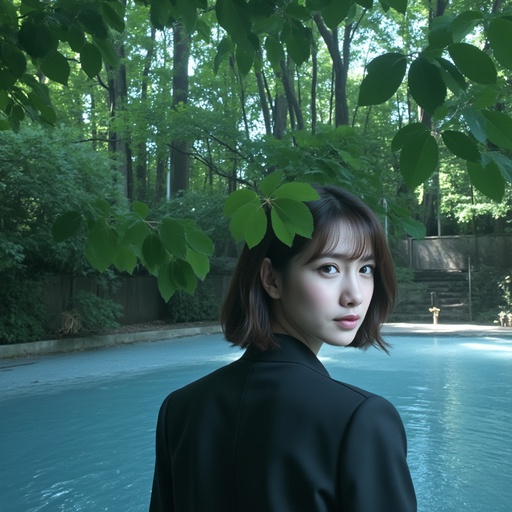} &
    \includegraphics[height=\cropheight]{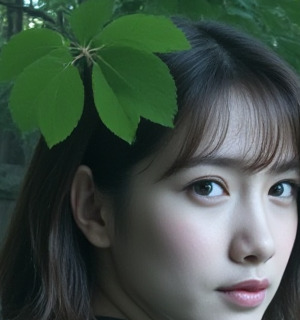} &
    \includegraphics[height=\cropheight]{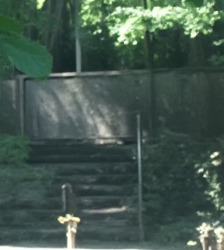}  \\
\end{tabular}
    }
    \vspace{-1em}
    \caption{\textbf{Visual comparison on InScene Synthetic dataset across seven methods.}}
    \label{fig:syn_visual_results_part2}
\end{figure*}

\begin{figure*}[htb]
    \centering
    \newcommand{\cropheight}{0.20\textwidth}
    \resizebox{0.92\textwidth}{!}{
    \begin{tabular}{@{} c *{6}{c} @{}}

    \rotatebox{90}{LQ} &
    \includegraphics[width=0.20\textwidth]{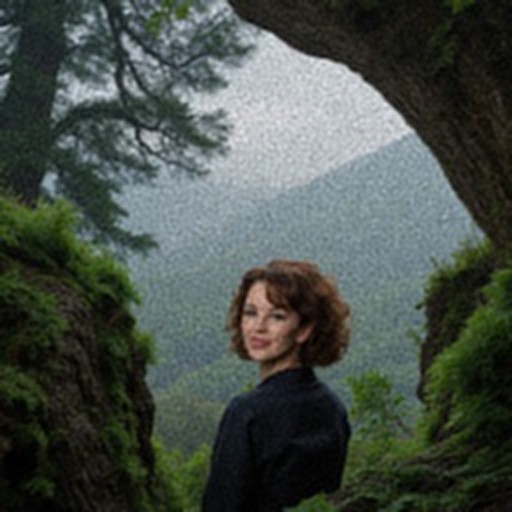} &
    \includegraphics[height=\cropheight]{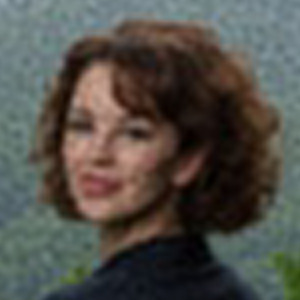} &
    \includegraphics[height=\cropheight]{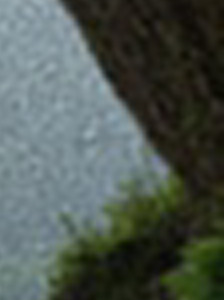} &
    \includegraphics[width=0.20\textwidth]{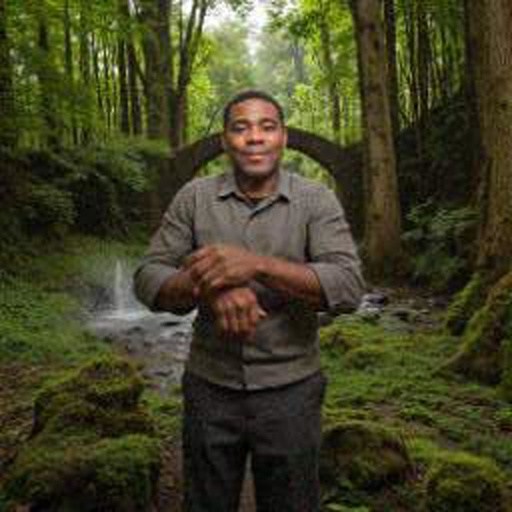} &
    \includegraphics[height=\cropheight]{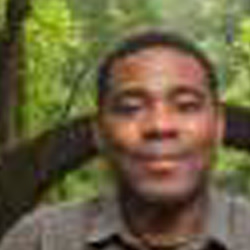} &
    \includegraphics[height=\cropheight]{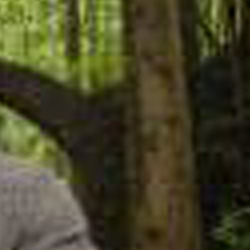}  \\

    \rotatebox{90}{GT} &
    \includegraphics[width=0.20\textwidth]{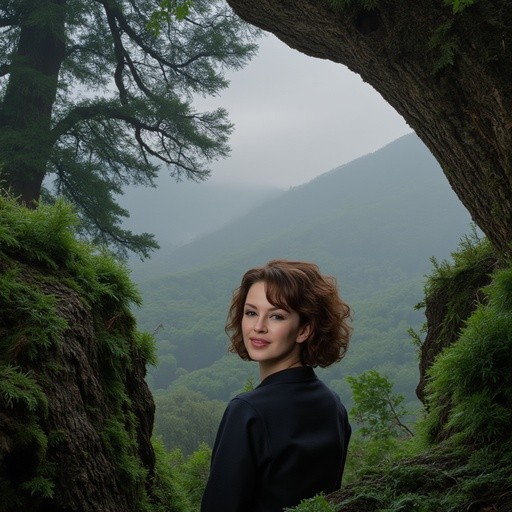} &
    \includegraphics[height=\cropheight]{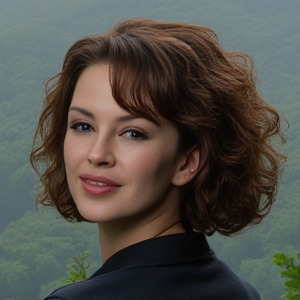} &
    \includegraphics[height=\cropheight]{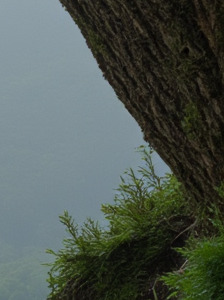} &
    \includegraphics[width=0.20\textwidth]{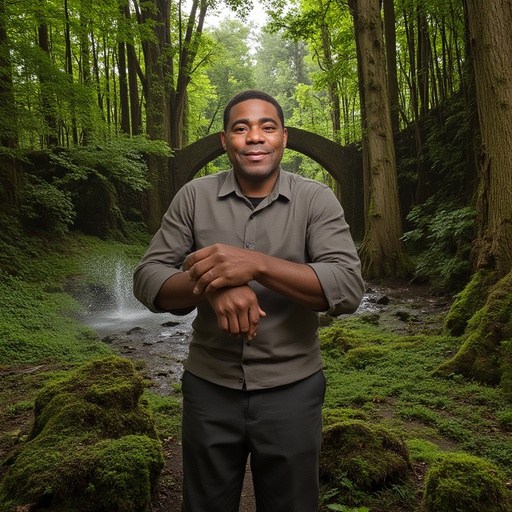} &
    \includegraphics[height=\cropheight]{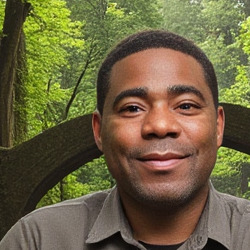} &
    \includegraphics[height=\cropheight]{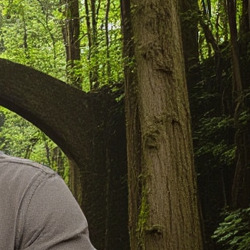}  \\
\end{tabular}
    }
    \caption{\textbf{Visual comparison on InScene Synthetic dataset across seven methods.}}
    \label{fig:syn_visual_results_part3}
\end{figure*}

\begin{figure*}[htb]
    \centering
    \newcommand{\cropheight}{0.20\textwidth}
    \resizebox{0.92\textwidth}{!}{
    \begin{tabular}{@{} c *{6}{c} @{}}

    \rotatebox{90}{LQ} &
    \includegraphics[width=0.20\textwidth]{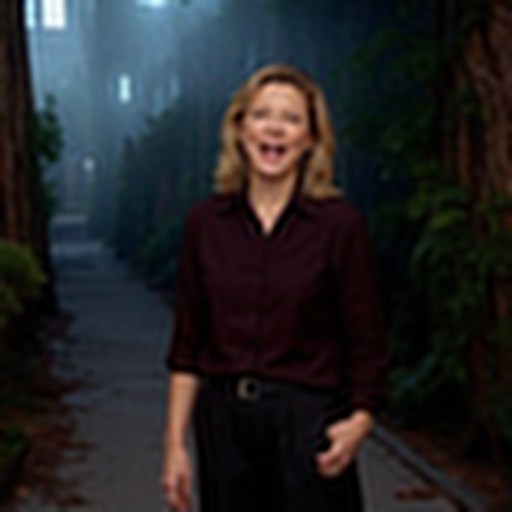} &
    \includegraphics[height=\cropheight]{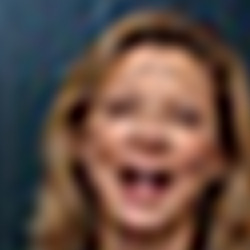} &
    \includegraphics[height=\cropheight]{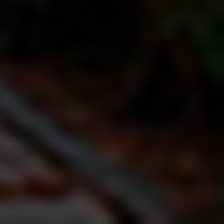} &
    \includegraphics[width=0.20\textwidth]{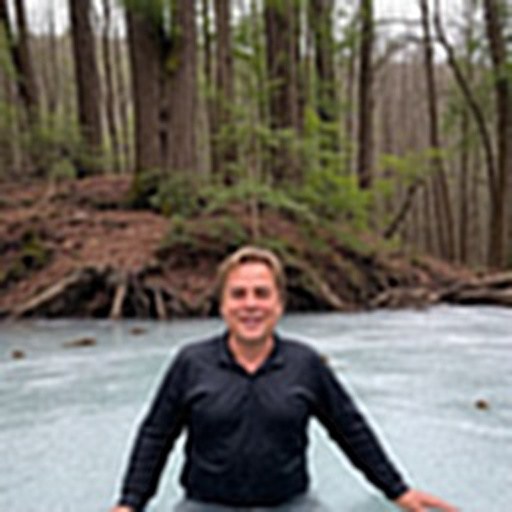} &
    \includegraphics[height=\cropheight]{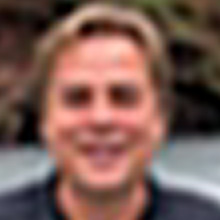} &
    \includegraphics[height=\cropheight]{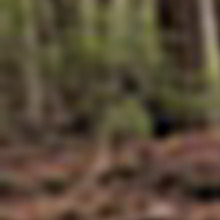}  \\

    \rotatebox{90}{GT} &
    \includegraphics[width=0.20\textwidth]{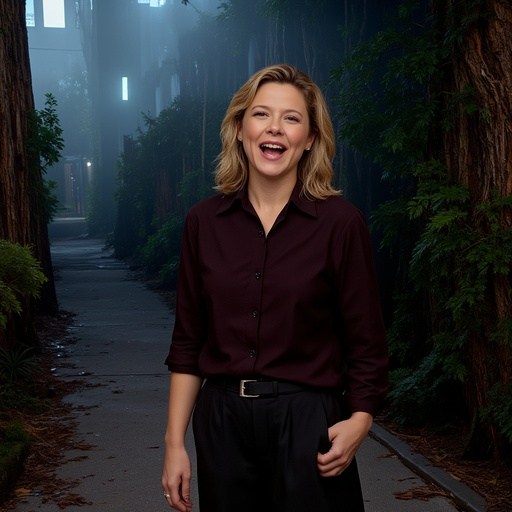} &
    \includegraphics[height=\cropheight]{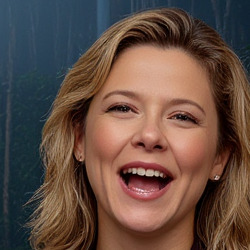} &
    \includegraphics[height=\cropheight]{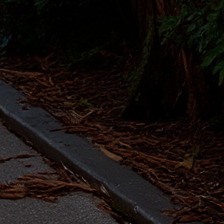} &
    \includegraphics[width=0.20\textwidth]{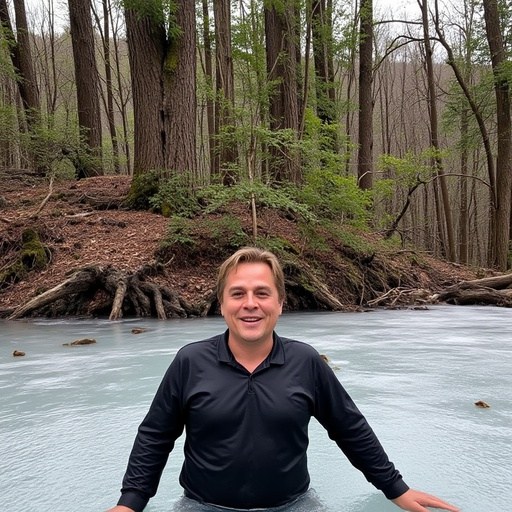} &
    \includegraphics[height=\cropheight]{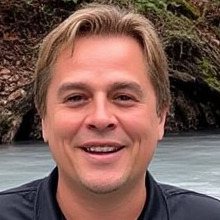} &
    \includegraphics[height=\cropheight]{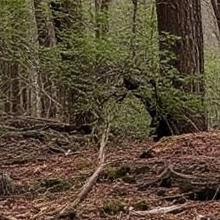}  \\
\end{tabular}
    }
    \caption{\textbf{Visual comparison on InScene Synthetic dataset across seven methods.}}
    \label{fig:syn_visual_results_part4}
\end{figure*}

\begin{figure*}[htb]
    \centering
    \newcommand{\cropheight}{0.20\textwidth}
    \resizebox{0.92\textwidth}{!}{
    \begin{tabular}{@{} c *{6}{c} @{}}

    \rotatebox{90}{LQ} &
    \includegraphics[width=0.20\textwidth]{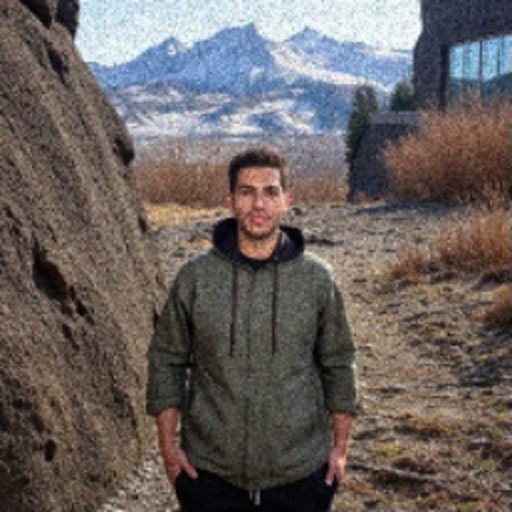} &
    \includegraphics[height=\cropheight]{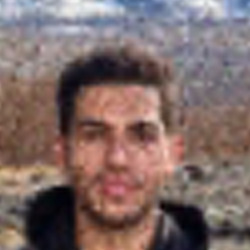} &
    \includegraphics[height=\cropheight]{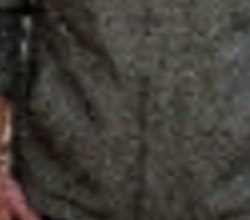} &
    \includegraphics[width=0.20\textwidth]{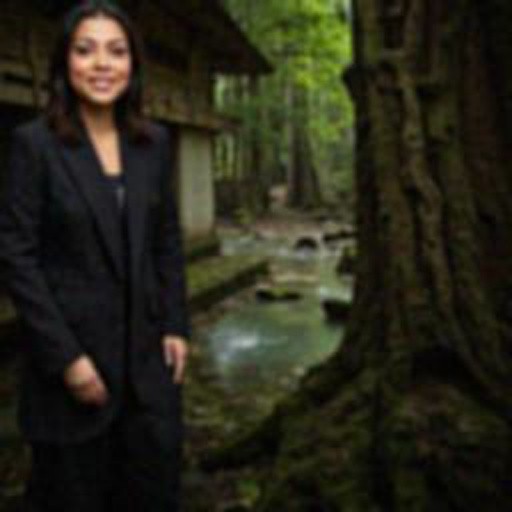} &
    \includegraphics[height=\cropheight]{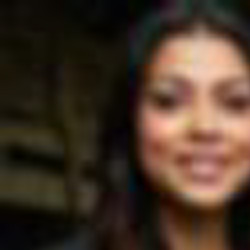} &
    \includegraphics[height=\cropheight]{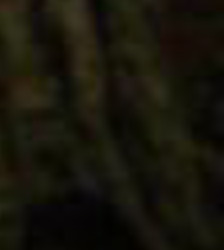}  \\

    \rotatebox{90}{GT} &
    \includegraphics[width=0.20\textwidth]{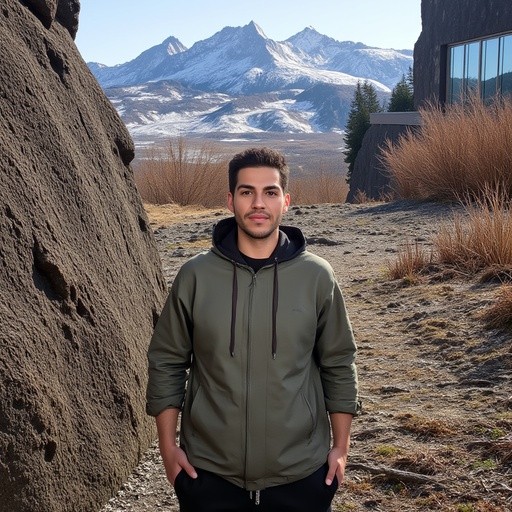} &
    \includegraphics[height=\cropheight]{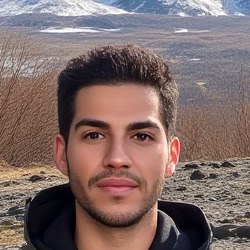} &
    \includegraphics[height=\cropheight]{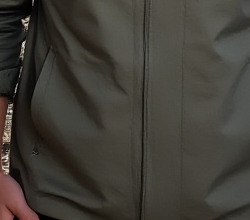} &
    \includegraphics[width=0.20\textwidth]{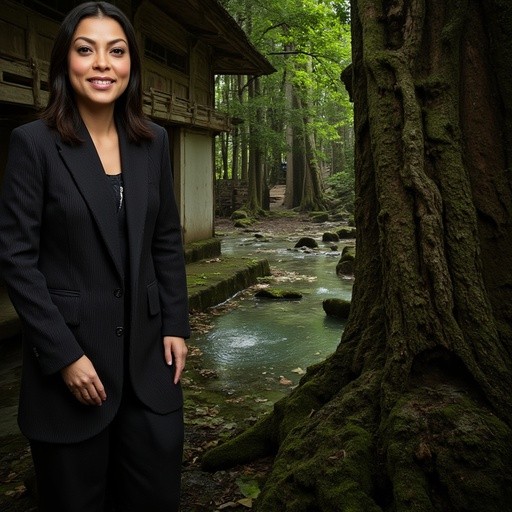} &
    \includegraphics[height=\cropheight]{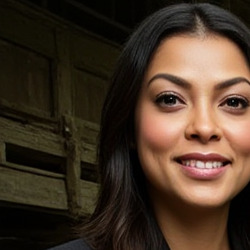} &
    \includegraphics[height=\cropheight]{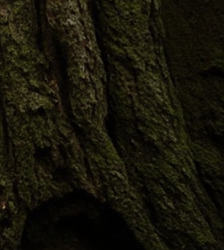}  \\
\end{tabular}
    }
    \caption{\textbf{Visual comparison on InScene Synthetic dataset across seven methods.}}
    \label{fig:syn_visual_results_part5}
\end{figure*}

\begin{figure*}[htb]
    \centering
    \newcommand{\cropheight}{0.20\textwidth}
    \resizebox{0.92\textwidth}{!}{
    \begin{tabular}{@{} c *{6}{c} @{}}

    \rotatebox{90}{LQ} &
    \includegraphics[width=0.20\textwidth]{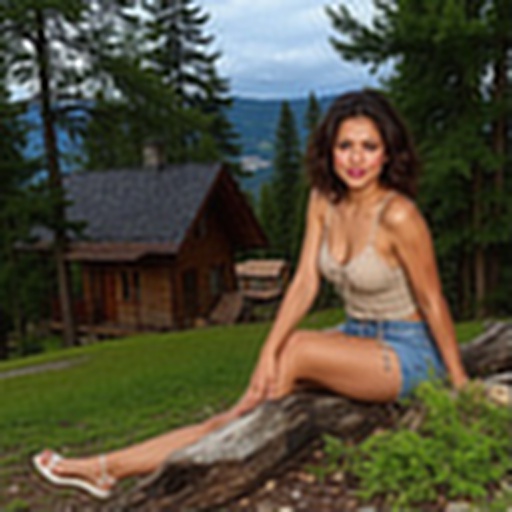} &
    \includegraphics[height=\cropheight]{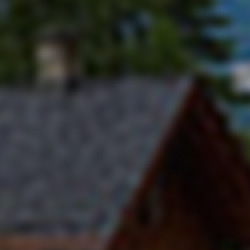} &
    \includegraphics[height=\cropheight]{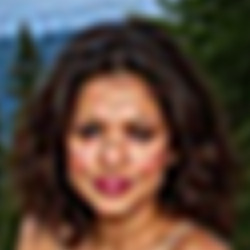} &
    \includegraphics[width=0.20\textwidth]{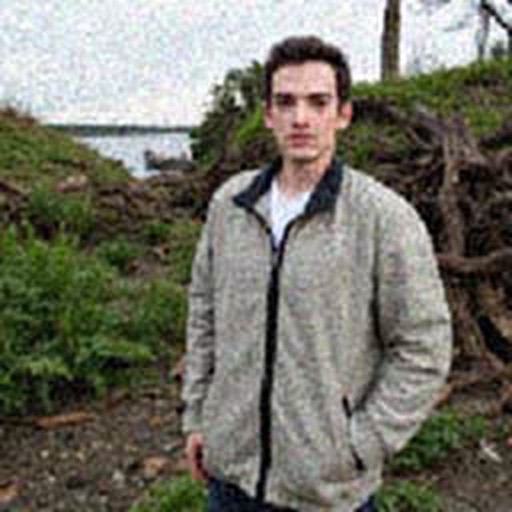} &
    \includegraphics[height=\cropheight]{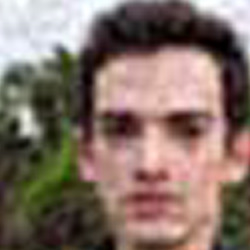} &
    \includegraphics[height=\cropheight]{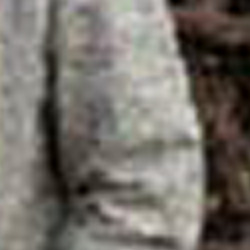}  \\

    \rotatebox{90}{GT} &
    \includegraphics[width=0.20\textwidth]{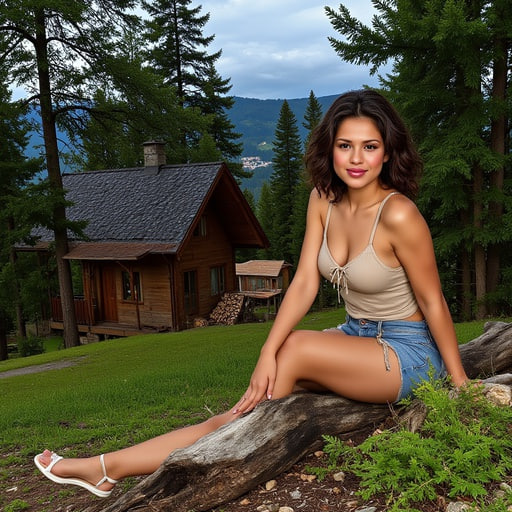} &
    \includegraphics[height=\cropheight]{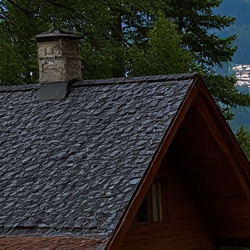} &
    \includegraphics[height=\cropheight]{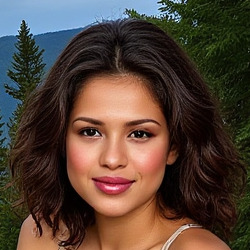} &
    \includegraphics[width=0.20\textwidth]{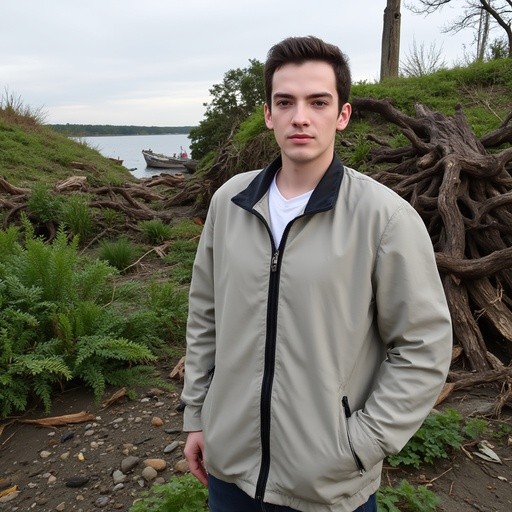} &
    \includegraphics[height=\cropheight]{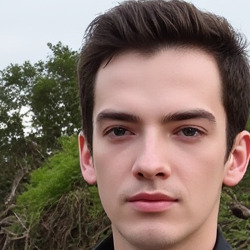} &
    \includegraphics[height=\cropheight]{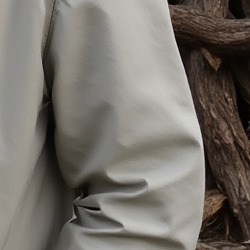}  \\
\end{tabular}
    }
    \caption{\textbf{Visual comparison on InScene Synthetic dataset across seven methods.}}
    \label{fig:syn_visual_results_part6}
\end{figure*}

\begin{figure*}[htb]
    \centering
    \newcommand{\cropheight}{0.20\textwidth}
    \resizebox{0.88\textwidth}{!}{
    \begin{tabular}{@{} c *{6}{c} @{}}

    \rotatebox{90}{LQ} &
    \includegraphics[width=0.20\textwidth]{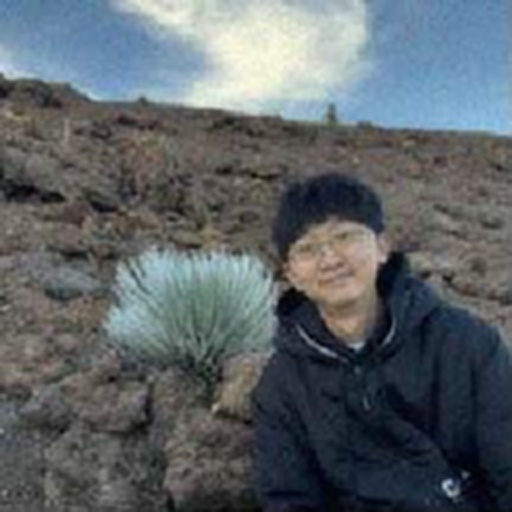} &
    \includegraphics[height=\cropheight]{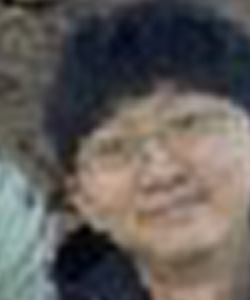} &
    \includegraphics[height=\cropheight]{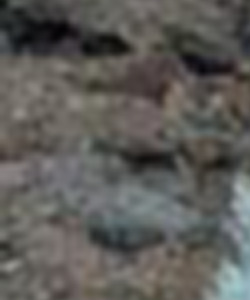} &
    \includegraphics[width=0.20\textwidth]{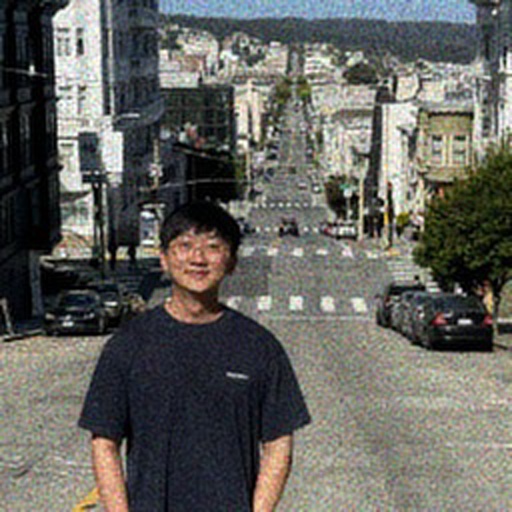} &
    \includegraphics[height=\cropheight]{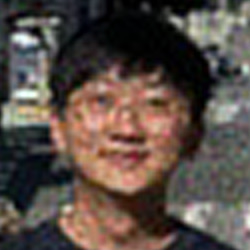} &
    \includegraphics[height=\cropheight]{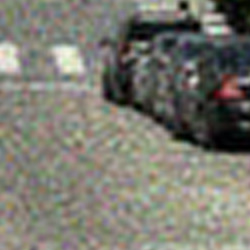}  \\

    \rotatebox{90}{GT} &
    \includegraphics[width=0.20\textwidth]{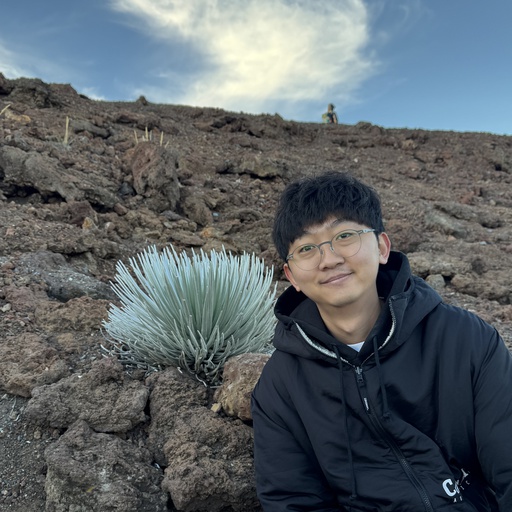} &
    \includegraphics[height=\cropheight]{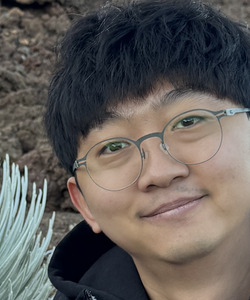} &
    \includegraphics[height=\cropheight]{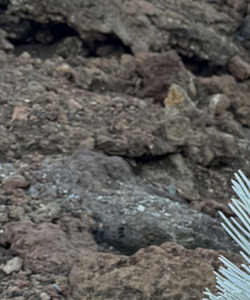} &
    \includegraphics[width=0.20\textwidth]{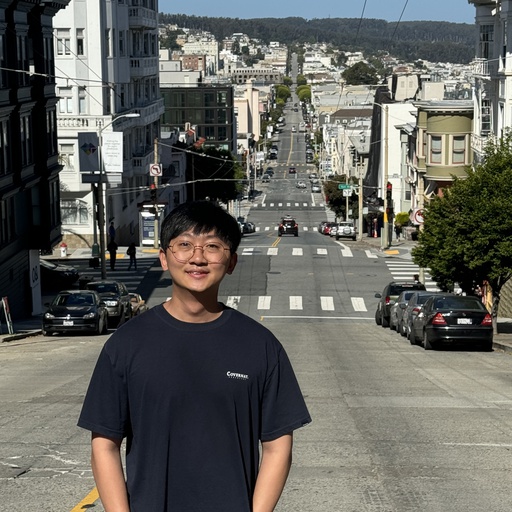} &
    \includegraphics[height=\cropheight]{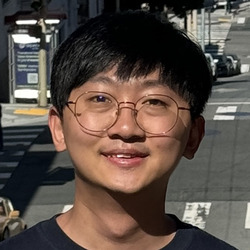} &
    \includegraphics[height=\cropheight]{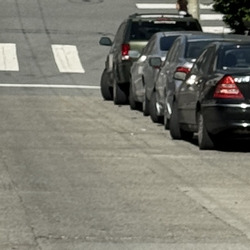}  \\
\end{tabular}
    }
    \caption{\textbf{Visual comparison on real validation across seven methods.}}
    \label{fig:real_visual_results_part_00}
\end{figure*}

\begin{figure*}[htb]
    \centering
    \newcommand{\cropheight}{0.20\textwidth}
    \resizebox{0.92\textwidth}{!}{
    \begin{tabular}{@{} c *{6}{c} @{}}

    \rotatebox{90}{LQ} &
    \includegraphics[width=0.20\textwidth]{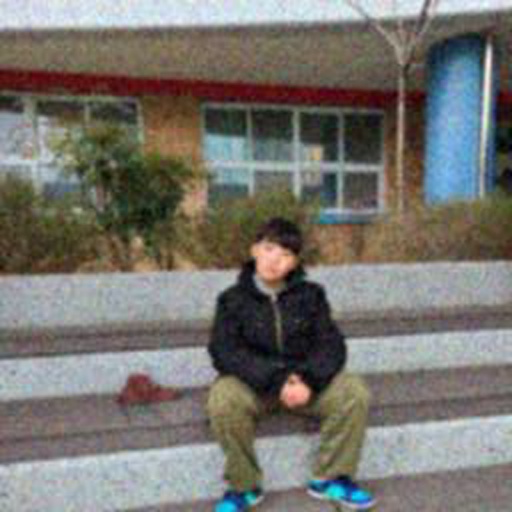} &
    \includegraphics[height=\cropheight]{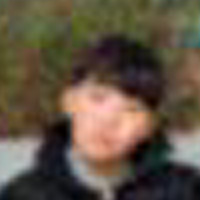} &
    \includegraphics[height=\cropheight]{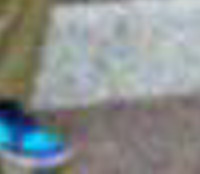} &
    \includegraphics[width=0.20\textwidth]{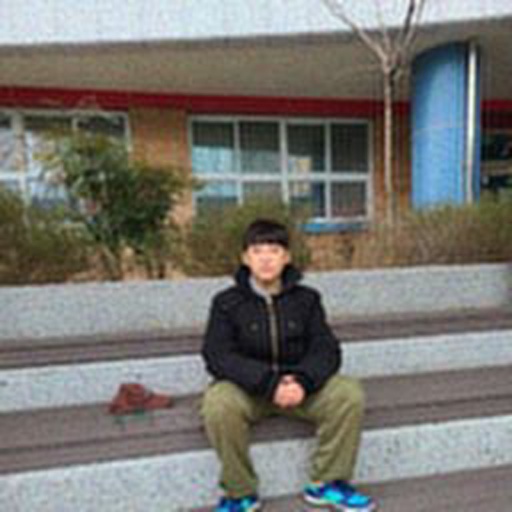} &
    \includegraphics[height=\cropheight]{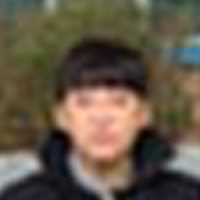} &
    \includegraphics[height=\cropheight]{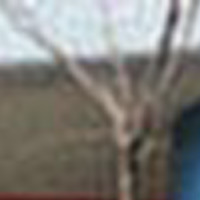}  \\

    \rotatebox{90}{GT} &
    \includegraphics[width=0.20\textwidth]{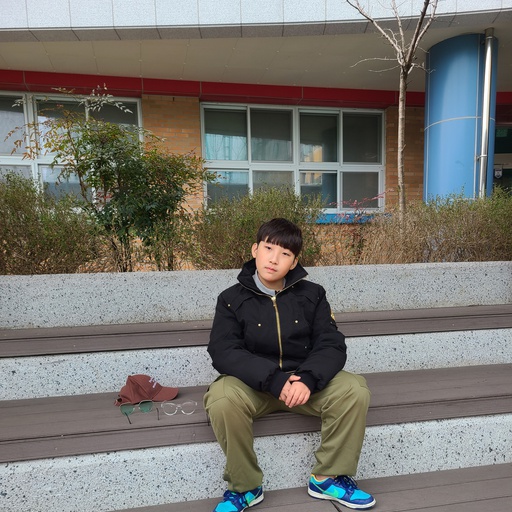} &
    \includegraphics[height=\cropheight]{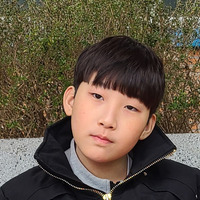} &
    \includegraphics[height=\cropheight]{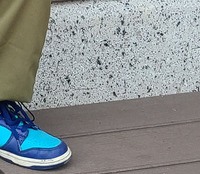} &
    \includegraphics[width=0.20\textwidth]{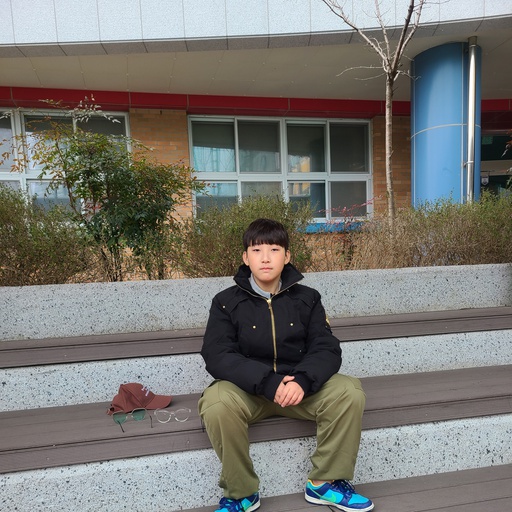} &
    \includegraphics[height=\cropheight]{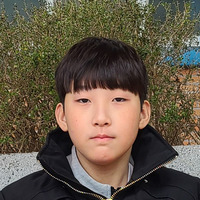} &
    \includegraphics[height=\cropheight]{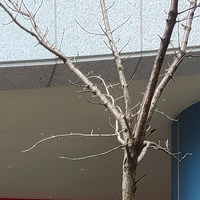}  \\
\end{tabular}
    }
    \caption{\textbf{Visual comparison on real validation across seven methods.}}
    \label{fig:real_visual_results_part0}
\end{figure*}

As shown in the qualitative comparisons on the 
InScene 
synthetic 
(Figures~\ref{fig:syn_visual_results_part1},
\ref{fig:syn_visual_results_part2},
 \ref{fig:syn_visual_results_part3},
 \ref{fig:syn_visual_results_part4},
 \ref{fig:syn_visual_results_part5}, and
 \ref{fig:syn_visual_results_part6}), and 
real validation datasets 
(Figures~\ref{fig:real_visual_results_part_00},
and \ref{fig:real_visual_results_part0})
conditioning only on the face-derived degradation prior enables our method to restore substantially more coherent backgrounds with richer textures, directly aligning with our core motivation. In the facial regions, our approach also produces more human-like appearances with fewer distortions than existing restoration or diffusion-based generative models. These qualitative observations hold consistently for both the synthetic and real validation sets and align with the quantitative trends, where our method achieves higher perceptual scores and stronger no-reference image quality metrics. Together, these results confirm that estimating degradation solely from faces provides an effective prior that improves both global scene quality and facial fidelity across synthetic and real data.

\subsubsection{InScene Test Dataset}
\begin{figure*}[htb]
    \centering
    \newcommand{\cropheight}{0.20\textwidth}
    \resizebox{0.95\textwidth}{!}{
    \begin{tabular}{@{} c *{6}{c} @{}}

    \rotatebox{90}{LQ} &
    \includegraphics[width=0.20\textwidth]{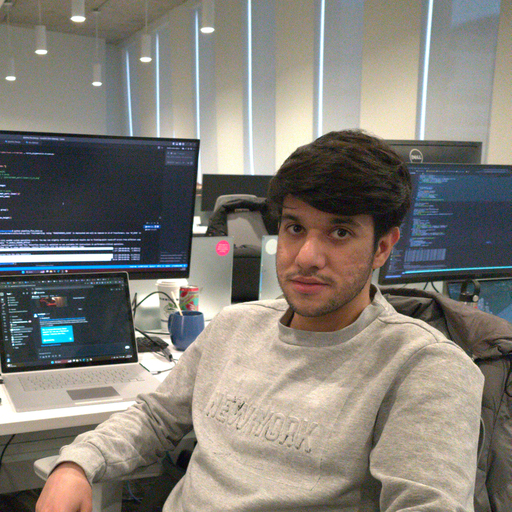} &
    \includegraphics[height=\cropheight]{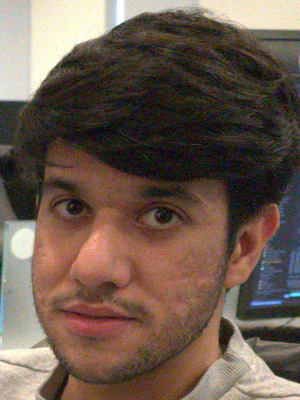} &
    \includegraphics[width=0.20\textwidth]{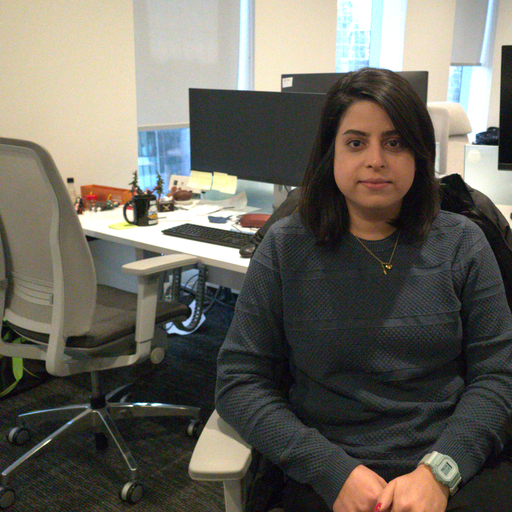} &
    \includegraphics[height=\cropheight]{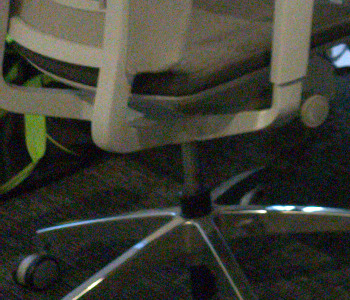} &
    \includegraphics[width=0.20\textwidth]{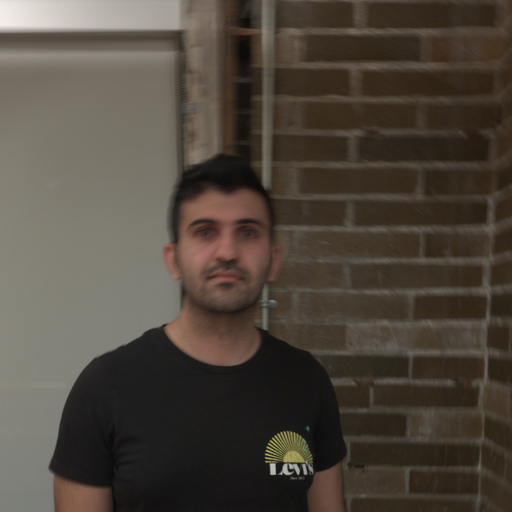} &
    \includegraphics[height=\cropheight]{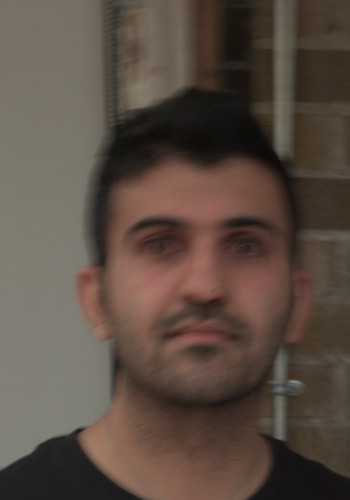} \\

    \rotatebox{90}{Face2Scene} &
    \includegraphics[width=0.20\textwidth]{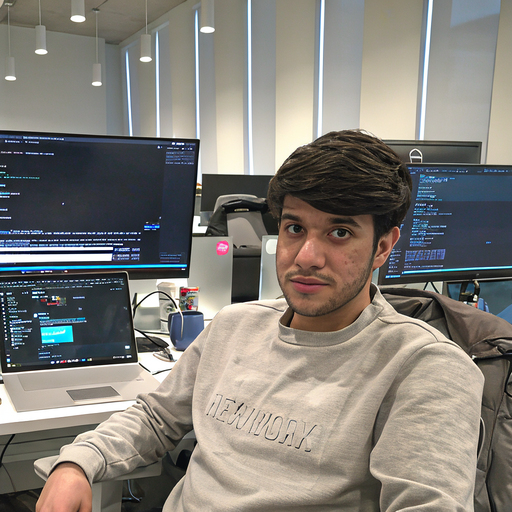} &
    \includegraphics[height=\cropheight]{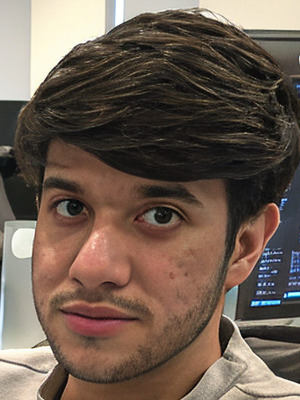} &
    \includegraphics[width=0.20\textwidth]{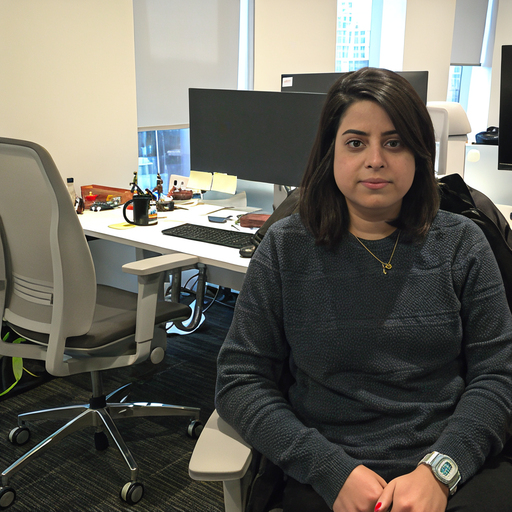} &
    \includegraphics[height=\cropheight]{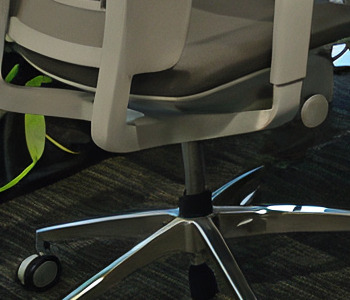} &
    \includegraphics[width=0.20\textwidth]{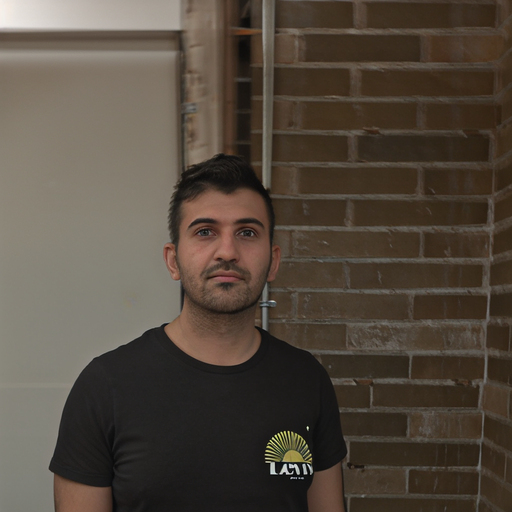} &
    \includegraphics[height=\cropheight]{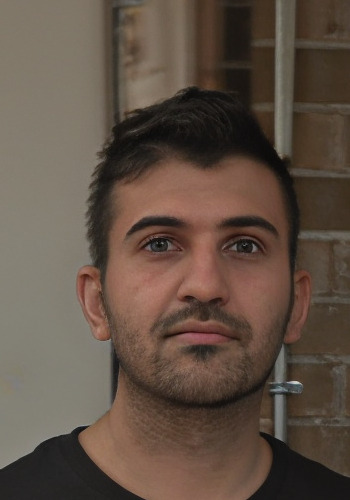} \\
\end{tabular}
    }
    \caption{\textbf{Visual comparison across restoration methods on real test samples.}}
    \label{fig:real_test_samples_triple}
\end{figure*}

\begin{figure*}[htb]
    \centering
    \newcommand{\cropheight}{0.20\textwidth}
    \resizebox{0.92\textwidth}{!}{
    \begin{tabular}{@{} c *{6}{c} @{}}

    \rotatebox{90}{LQ} &
    \includegraphics[width=0.20\textwidth]{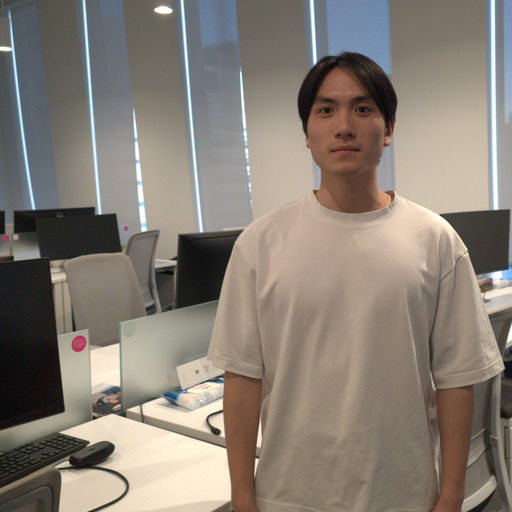} &
    \includegraphics[height=\cropheight]{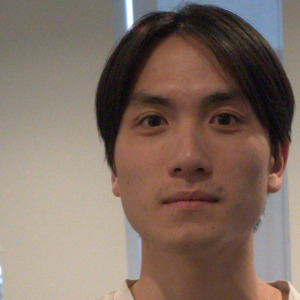} &
    \includegraphics[height=\cropheight]{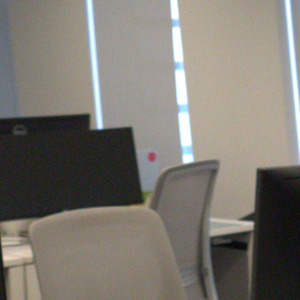} &
    \includegraphics[width=0.20\textwidth]{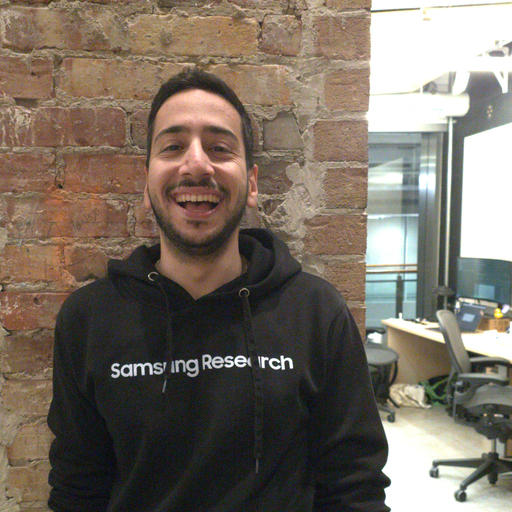} &
    \includegraphics[height=\cropheight]{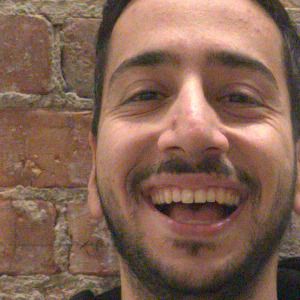} &
    \includegraphics[height=\cropheight]{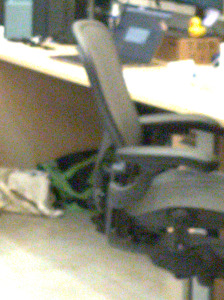} \\

    \rotatebox{90}{Face2Scene} &
    \includegraphics[width=0.20\textwidth]{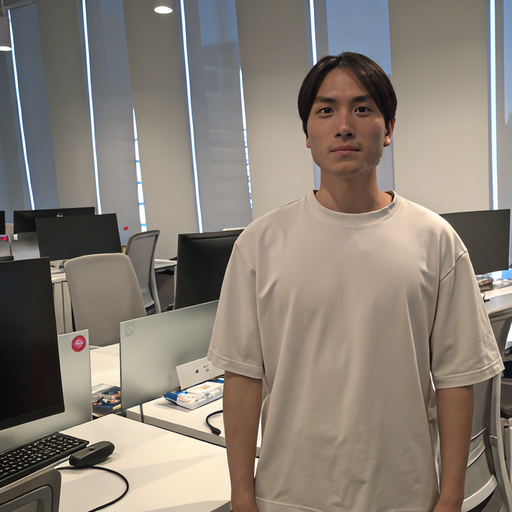} &
    \includegraphics[height=\cropheight]{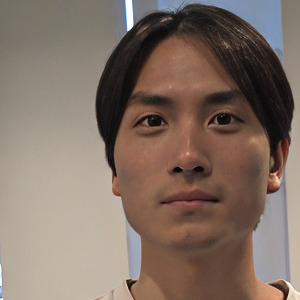} &
    \includegraphics[height=\cropheight]{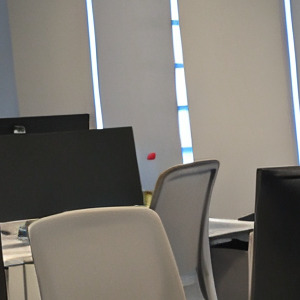} &
    \includegraphics[width=0.20\textwidth]{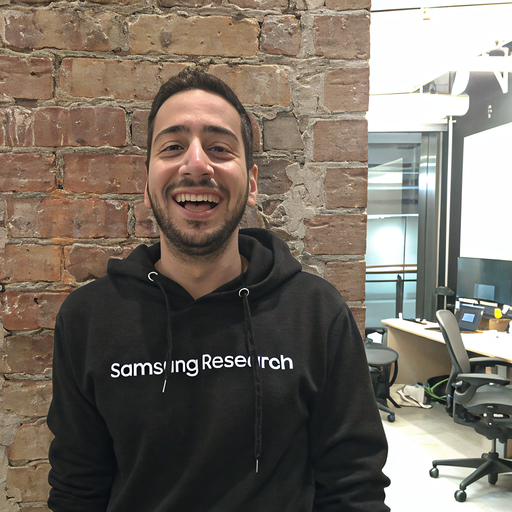} &
    \includegraphics[height=\cropheight]{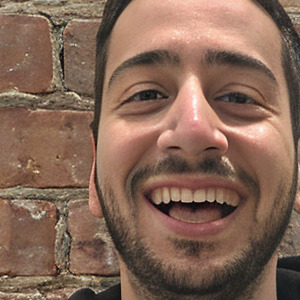} &
    \includegraphics[height=\cropheight]{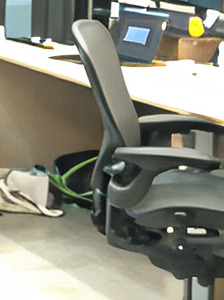} \\

    \rotatebox{90}{S3Diff} &
    \includegraphics[width=0.20\textwidth]{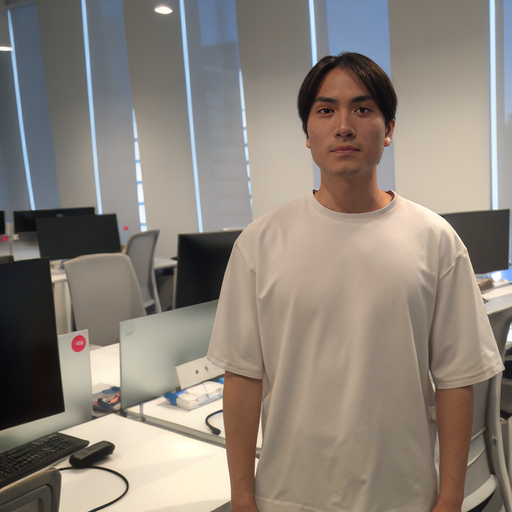} &
    \includegraphics[height=\cropheight]{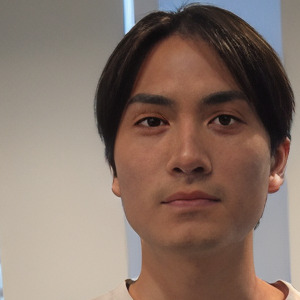} &
    \includegraphics[height=\cropheight]{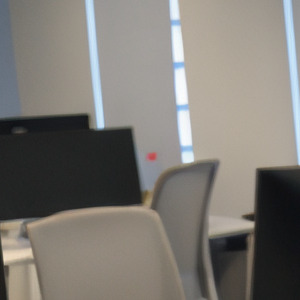} &
    \includegraphics[width=0.20\textwidth]{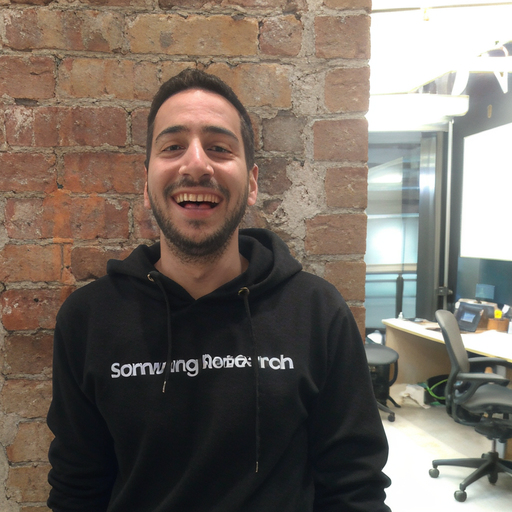} &
    \includegraphics[height=\cropheight]{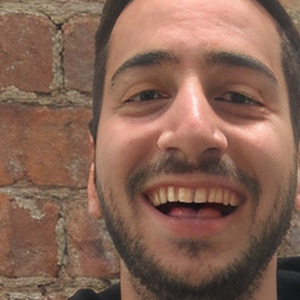} &
    \includegraphics[height=\cropheight]{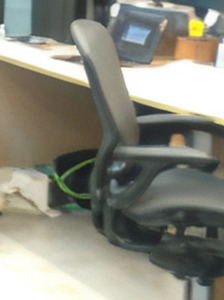} \\
\end{tabular}
    }
    \caption{\textbf{Visual comparison across restoration methods on real test samples.}}
    \label{fig:real_test_samples_174_0}
\end{figure*}

{
Figures~\ref{fig:real_test_samples},
\ref{fig:real_test_samples_triple}, and
\ref {fig:real_test_samples_174_0} show real test examples captured by the Samsung S25 Edge mobile device and restored by different methods. As can be observed in the examples, our method reconstructs a higher quality face with less artifacts. Further, on the background regions, our model yields sharper outputs with more details and less noise. 
In particular, 
the reconstructed face generated by our method tends to maintain better fidelity to the LQ input, meaning it can recover from blurriness and noise, yet also avoid hallucinated artifacts.
For example, in \autoref{fig:real_test_samples_triple} (rightmost column), our model is able to deblur the face with plausible details, whereas either output a blurry face (most other methods) or an artifacted one (OSEDiff and S3Diff).
Aside from the face, we also see high frequency detail recovery is improved; for instance, in \autoref{fig:real_test_samples} (left columns), the horizontal stripes are preserved sharply by our method, while others blur them, restore them in a slightly cartoonish manner, or remove them entirely.
}
\subsection{Model Complexity}
\begin{figure}[t]
    \centering
    \includegraphics[width=1\linewidth]{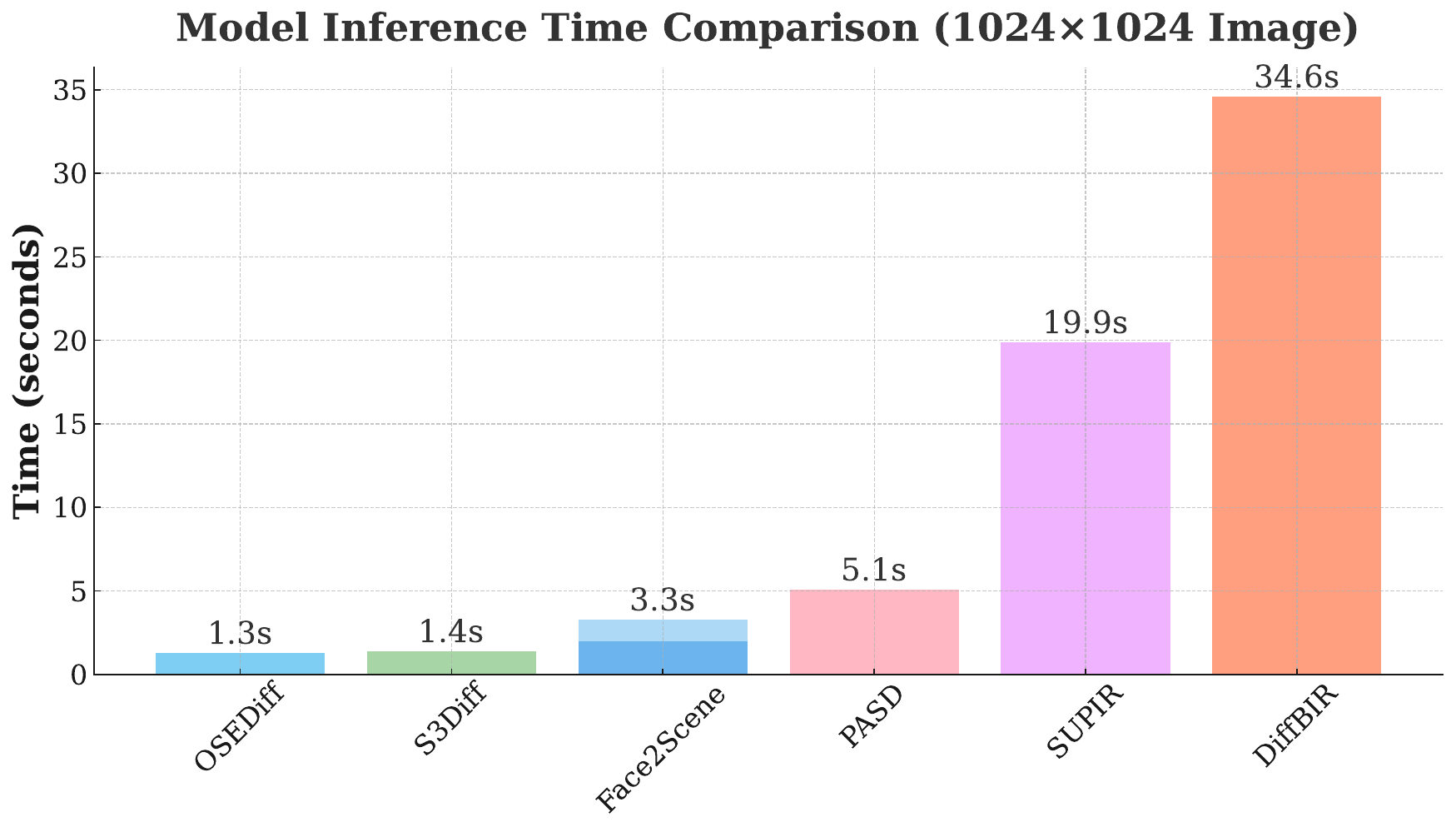}
    \caption{\textbf{Model complexity comparison.} 
    We compare inference time across different diffusion-based restoration models.
    For Face2Scene, we show the separate time costs of the reference-based preprocessing and our final restoration as two different shades.
    While our method is not as efficient as standard one-step models, which do not require an initial face restoration step, it is still considerably faster than current multistep models, despite performing on-par or better than them, in terms of image quality and restoration fidelity.
    }
    \label{fig:model_inference_time}
\end{figure}

To contextualize the efficiency of our approach, we compare the inference time of Face2Scene against several state-of-the-art restoration baselines (OSEDiff \cite{wu2024one}, S3Diff \cite{zhang2024degradation}, PASD \cite{yang2024pixel}, SUPIR \cite{yu2024scaling}, and DiffBIR \cite{lin2024diffbir}), 
as shown in~\autoref{fig:model_inference_time}.  
All timings are measured on an NVIDIA A6000~48GB GPU using $1024{\times}1024$ input images, averaged over 100 samples. For each model, we follow the official inference pipeline. Notably, in our method, the degradation extractor operates on the cropped face region, whose size varies across images. Since the module is fully resolution-agnostic, its runtime naturally scales with the face-crop resolution:  
smaller faces lead to faster processing than the reported average, whereas larger faces incur a slightly higher cost.  

\section{Ablation Studies}

\subsection{Impact of CFG scaling factor}
\begin{figure*}[t]
    \centering
    \includegraphics[width=1\textwidth]{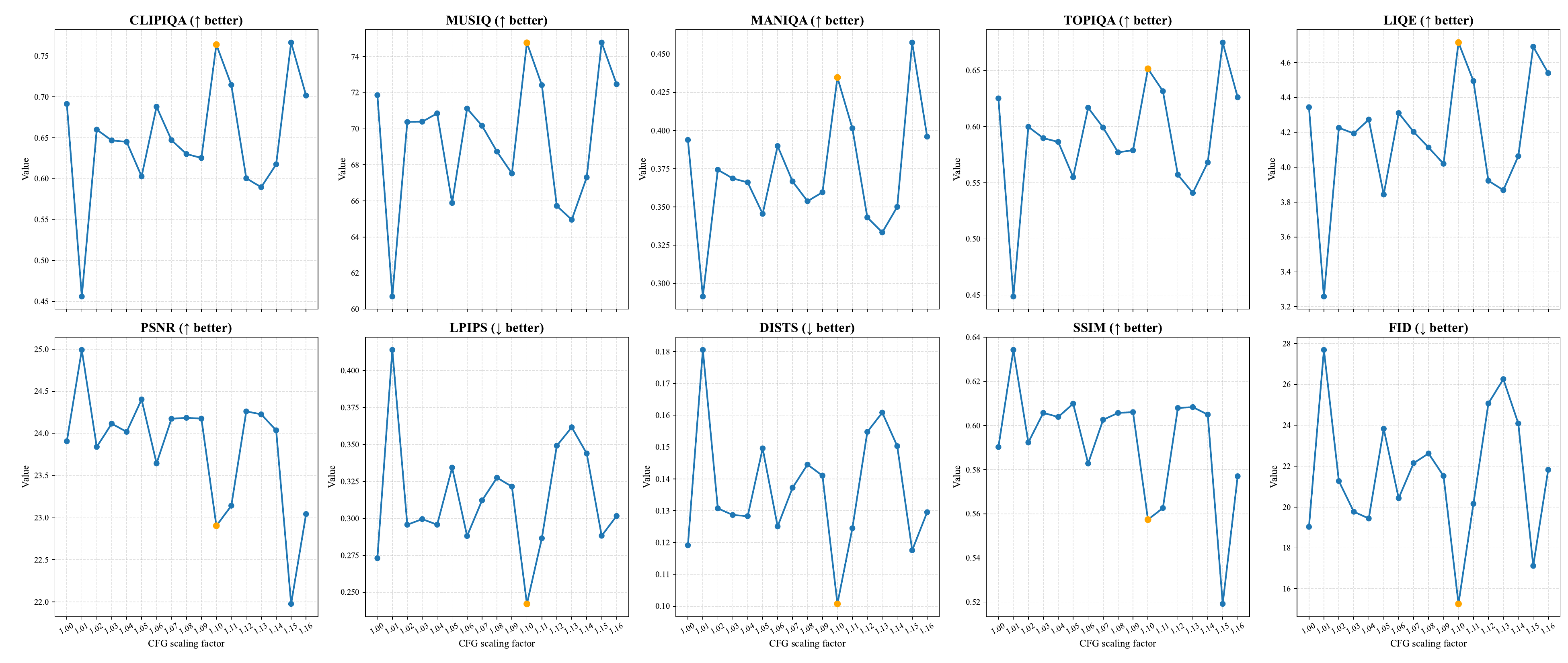}
    \caption{\textbf{Effect of CFG scaling.} 
We vary the CFG scaling factor from 1.00 to 1.16 on the InScene synthetic validation set. 
The top row shows no-reference metrics and the bottom row shows reference-based metrics, illustrating the trade-off between perceptual quality and fidelity.
In particular, although there is some noise in the trends, one can see that the no-reference quality metrics tend to improve as the CFG weight increases, while the low-level fidelity (PSNR and SSIM) tend to decline. 
The orange dot signifies the CFG weight that we chose, 
    which obtains high perceptual quality while maintaining fidelity (particularly according to the full-reference perceptual metrics).
}
    \label{fig:cfg_metric}
\end{figure*}

We study the effect of the classifier-free guidance (CFG) scaling factor on our restoration pipeline.  
To isolate this behavior, we sweep the scaling value from 1.00 (using only the positive prompt) up to 1.16 and evaluate performance on the InScene Synthetic Validation set. \autoref{fig:cfg_metric} presents the results, where the first row shows no-reference metrics 
(CLIPIQA, MUSIQ, MANIQA, TOPIQ, LIQE) and the second row shows reference-based metrics 
(PSNR, LPIPS, DISTS, SSIM, FID).

Overall, we observe a consistent trend: increasing the CFG scaling factor improves no-reference metrics, indicating that the generated images become visually sharper and more perceptually appealing. However, this comes at the cost of reduced reference-based performance, where fidelity to the ground-truth image gradually decreases as the scaling factor increases. This trade-off reflects the classical balance between perceptual sharpness and reconstruction accuracy \cite{blau2018perception}; higher guidance pushes the model toward more confident, high-frequency outputs, while lower guidance yields more faithful but slightly softer reconstructions. Based on this analysis, we set the default value to $\lambda_{\text{cfg}} = 1.10$ for all experiments, which provides a strong perceptual gain while retaining competitive fidelity.

\begin{table*}[t]
\centering
\caption{\textbf{Impact of LoRA rank on the VAE encoder and UNet.} 
We evaluate different LoRA rank configurations for the VAE and UNet to analyze their effect on restoration quality across perceptual, fidelity, and no-reference metrics on the InScene synthetic validation set.
We observe that low ranks heavily damage performance, on both perceptual distances (LPIPS and DISTS) and no-reference quality.
Our 16/32 configuration performs quite well on both types, though it is at the expense of pixel-level metrics (PSNR and SSIM). 
Increasing the rank beyond this causes a slight decline according to most metrics.
}
\label{tab:lora_exp}
\resizebox{\textwidth}{!}{
\setlength{\tabcolsep}{8pt}
\begin{tabular}{cc|ccccc|ccccc}
\toprule
\multicolumn{12}{c}{\textbf{InScene Synthetic Validation Dataset}} \\ 
\midrule
\textbf{VAE LoRA Rank}
& \textbf{UNet LoRA Rank}
& \textbf{DISTS$\downarrow$} 
& \textbf{LPIPS$\downarrow$} 
& \textbf{PSNR$\uparrow$} 
& \textbf{SSIM$\uparrow$} 
& \textbf{FID$\downarrow$} 
& \textbf{MUSIQ$\uparrow$} 
& \textbf{C-IQA$\uparrow$} 
& \textbf{M-IQA$\uparrow$} 
& \textbf{LIQE$\uparrow$} 
& \textbf{TOPIQ$\uparrow$} \\
\midrule
4 & 4 
& 0.1771 & 0.4034 & \cellcolor{purple}{24.1619} & \cellcolor{purple}{0.6153} & 29.15 
& 62.0885 & 0.5289 & 0.3093 & 3.6797 & 0.4913 \\

8 & 8 
& 0.1523 & 0.3505 & 23.9881 & 0.6035 & 23.28 
& 65.8073 & 0.6009 & 0.3433 &3.9864 & 0.5469 \\

16 & 16 
& 0.1244 & 0.2881 & 23.4638 & 0.5834 & 19.58 
& 71.1710 & 0.6928 & 0.3906 & 4.3963 & 0.5984 \\

16 & 32 
& \cellcolor{purple}{0.1007} & \cellcolor{purple}{0.2421} & 22.9040 & 0.5574 & \cellcolor{purple}{15.26} & \cellcolor{purple}{74.7630} & \cellcolor{purple}{0.7640} & {0.4347} & \cellcolor{purple}{4.7157} & {0.6515} \\

32 & 32 
& 0.1076 & 0.2536 & 23.3672 & 0.5736 & 16.61 
& 73.7835 & 0.7545 & \cellcolor{purple}{0.4434} & 4.6329 & \cellcolor{purple}{0.6607} \\
\bottomrule
\end{tabular}
}
\end{table*}

\subsection{Impact of LoRA Rank}
\autoref{tab:lora_exp} shows how the LoRA rank for the VAE encoder and UNet affects restoration quality. 
Very low ranks (4/4, 8/8) substantially degrade both fidelity (PSNR/SSIM) and perceptual/no-reference metrics. 
Increasing the rank to 16/16 already yields strong gains, and the asymmetric setting (VAE: 16, UNet: 32) further improves DISTS, LPIPS, FID, and all no-reference scores, while keeping PSNR and SSIM competitive. 
We therefore adopt the 16/32 configuration as our default trade-off between performance and parameter efficiency.

\begin{table*}[t]
\centering
\caption{\textbf{Ablation on hierarchical components.}
We incrementally enable global, intermediate, and local components to assess their impact on restoration quality across perceptual, fidelity, and no-reference metrics.
Our complete configuration, which combines tokens across scales, performs the best across all perceptual evaluation measures (i.e., excluding the low-level distortion measures, PSNR and SSIM).
}
\label{tab:hierarchical_ablation}
\resizebox{\textwidth}{!}{
\begin{tabular}{l|ccccc|ccccc}
\toprule
\multicolumn{11}{c}{\textbf{InScene Synthetic Validation Dataset}} \\ 
\midrule
\textbf{Configuration}
& \textbf{DISTS$\downarrow$} 
& \textbf{LPIPS$\downarrow$} 
& \textbf{PSNR$\uparrow$} 
& \textbf{SSIM$\uparrow$} 
& \textbf{FID$\downarrow$} 
& \textbf{MUSIQ$\uparrow$} 
& \textbf{C-IQA$\uparrow$} 
& \textbf{M-IQA$\uparrow$} 
& \textbf{LIQE$\uparrow$} 
& \textbf{TOPIQ$\uparrow$} \\
\midrule
Global only (1 Token)
& 0.1180 & 0.2671 & 23.6223 & 0.5862 & 19.50 
& 72.0117 & 0.6497 & 0.3769 & 4.2772 & 0.5942 \\

Global + Intermediate (5 Tokens)
& 0.1294 & 0.2997 & \cellcolor{purple}{23.7422} & \cellcolor{purple}{0.5921} & 21.92 
& 69.9288 & 0.6582 & 0.3665 & 4.2190 & 0.5820 \\

Global + Intermediate + Local (21 Tokens)
& \cellcolor{purple}{0.1007} & \cellcolor{purple}{0.2421} & 22.9040 & 0.5574 & \cellcolor{purple}{15.26} & \cellcolor{purple}{74.7630} & \cellcolor{purple}{0.7640} & \cellcolor{purple}{0.4347} & \cellcolor{purple}{4.7157} & \cellcolor{purple}{0.6515}\\
\bottomrule
\end{tabular}
}
\end{table*}


\begin{table*}[t]
    \centering
    \caption{\textbf{Identity similarity on the InScene validation datasets.} 
    Our ID score measures similarity in terms of identity, meaning higher values indicate better identity preservation.
    Since our reference-aware degradation estimator explicitly discards non-degradation information, including identity, we investigated identity preservation as a metric specifically.
    We find that our method still performs best, in terms of retaining or inferring identity.
    This suggests our degradation estimation is accurate enough to help the model recover the true underlying face details, thereby preserving the correct identity as well.
    }
    \vspace{-0.5em}
    \resizebox{\textwidth}{!}{%
    \begin{tabular}{l|l|ccccccccc}
    \toprule
    \textbf{Dataset} 
    & \textbf{Metric $\uparrow$} 
    & SUPIR~\cite{yu2024scaling} 
    & DiffBIR~\cite{lin2024diffbir} 
    & ResShift~\cite{yue2024efficient}
    & PASD~\cite{yang2024pixel} 
    & OSEDiff~\cite{wu2024one} 
    & SinSR~\cite{wang2024sinsr}
    & InvSR~\cite{yue2025arbitrary} 
    & S3Diff~\cite{zhang2024degradation} 
    & \textbf{Face2Scene} \\
    \midrule

    \textbf{Synthetic} 
    & \multirow{2}{*}{ID Score}
    & 0.4434 & 0.4433 & 0.4477 & 0.4030 & 0.4477 & 0.4239 & 0.4055 & 0.4684 & \textbf{0.4867} \\

    \textbf{Real} 
    & 
    & 0.4333 & 0.4286 & 0.4692 & 0.4333 & 0.3766 & 0.4334 & 0.3892 & 0.4598 & \textbf{0.4881} \\

    \bottomrule
    \end{tabular}
    }
    \label{tab:identity_similarity}
\end{table*}


\begin{table*}[!]
\centering
\caption{
\textbf{Quantitative comparison on the InScene synthetic and real validation sets.} 
To investigate the importance of training with our dataset and provide a fairer comparison, 
    we retrain the S3Diff model on our InScene dataset under the same experimental settings,
    resulting in the model denoted S3Diff\textsuperscript{**}.
Although doing so does result in improved performance over the standard S3Diff, 
    our approach still outperforms S3Diff\textsuperscript{**} by a significant margin, 
    particularly according to the full-reference metrics on the Real set. 
Since our method is built upon S3Diff, this suggests our novel reference-derived degradation estimation module is necessary to obtain strong performance.
Arrows indicate whether lower ($\downarrow$) or higher ($\uparrow$) values are better. 
C-IQA and M-IQA denote CLIP-IQA and MANIQA, respectively.
Each cell is color-coded to represent the \colorbox{purple}{\kern-\fboxsep best\kern-\fboxsep} and \colorbox{light}{\kern-\fboxsep second-best\kern-\fboxsep} performance. 
}

\label{tab:synthetic_val_our_train}
\resizebox{\textwidth}{!}{%
\setlength{\tabcolsep}{8pt}
\begin{tabular}{l|ccccc|ccccc}
\toprule
\multicolumn{11}{c}{\textbf{InScene Synthetic Validation Dataset}} \\ 
\midrule
\textbf{Methods} 
& \textbf{DISTS$\downarrow$} 
& \textbf{LPIPS$\downarrow$} 
& \textbf{PSNR$\uparrow$} 
& \textbf{SSIM$\uparrow$} 
& \textbf{FID$\downarrow$} 
& \textbf{MUSIQ$\uparrow$} 
& \textbf{C-IQA$\uparrow$} 
& \textbf{M-IQA$\uparrow$} 
& \textbf{LIQE$\uparrow$} 
& \textbf{TOPIQ$\uparrow$} \\
\midrule
S3Diff~\cite{zhang2024degradation} 
& {0.1131} 
& {0.2557} 
& \cellcolor{purple}{23.5955} 
& \cellcolor{purple}{0.5916} 
& {18.06} 
& {72.1764} 
& 0.6980 
& 0.3858 
& {4.4248} 
& 0.6233 \\

S3Diff\textsuperscript{**} 
& \cellcolor{light}{0.1039} & \cellcolor{light}{0.2486} & 22.4990 & 0.5382 & \cellcolor{light}{15.81} & \cellcolor{light}{74.5945} & \cellcolor{light}{0.7605} & \cellcolor{purple}{0.4384} & \cellcolor{light}{4.7127} & \cellcolor{light}{0.6450} \\

\textbf{Face2Scene (ours)}      
& \cellcolor{purple}{0.1007} 
& \cellcolor{purple}{0.2421} 
& \cellcolor{light}{22.9040} 
& \cellcolor{light}{0.5574} 
& \cellcolor{purple}{15.26} 
& \cellcolor{purple}{74.7630} 
& \cellcolor{purple}{0.7640} 
& \cellcolor{light}{0.4347} 
& \cellcolor{purple}{4.7157} 
& \cellcolor{purple}{0.6515} \\
\midrule

\multicolumn{11}{c}{\textbf{InScene Real Validation Dataset}} \\ 
\midrule
\textbf{Methods} 
& \textbf{DISTS$\downarrow$} 
& \textbf{LPIPS$\downarrow$} 
& \textbf{PSNR$\uparrow$} 
& \textbf{SSIM$\uparrow$} 
& \textbf{FID$\downarrow$} 
& \textbf{MUSIQ$\uparrow$} 
& \textbf{C-IQA$\uparrow$} 
& \textbf{M-IQA$\uparrow$} 
& \textbf{LIQE$\uparrow$} 
& \textbf{TOPIQ$\uparrow$} \\
\midrule
S3Diff~\cite{zhang2024degradation} 
& {0.2231} 
& {0.5149} 
& \cellcolor{light}{17.1439} 
& 0.4894 
& \cellcolor{light}{38.64} 
& {73.8209} 
& 0.6734 
& 0.4480  
& {4.7060} 
& {0.6627} \\

S3Diff\textsuperscript{**}
& \cellcolor{light}{0.2176} & \cellcolor{light}{0.5069} & 16.9445 & 0.4702 & \cellcolor{purple}{37.63} & \cellcolor{light}{74.6644} & \cellcolor{light}{0.6872} & \cellcolor{purple}{0.4745} & \cellcolor{light}{4.7249} & \cellcolor{light}{0.6613} \\
\midrule
\textbf{Face2Scene (ours)}      
& \cellcolor{purple}{0.1178} 
& \cellcolor{purple}{0.2502} 
& \cellcolor{purple}{22.8975} 
& \cellcolor{purple}{0.6197} 
& {42.21} 
& \cellcolor{purple}{75.3739} 
& \cellcolor{purple}{0.7015} 
& \cellcolor{light}{0.4714} 
& \cellcolor{purple}{4.8044} 
& \cellcolor{purple}{0.6777} \\
\bottomrule
\end{tabular}}
\end{table*}

\subsection{Impact of Multi-scale Degradation Tokens}
To understand how the granularity of degradation encoding influences the restoration process, we ablate the number of degradation tokens used for conditioning the diffusion model. As shown in~\autoref{tab:hierarchical_ablation}, we compare three configurations: using only a single global token, adding a set of intermediate-scale tokens, and finally incorporating the full set of global, intermediate, and local tokens.

We observe a clear performance trend: increasing the number of degradation tokens consistently improves most perceptual and no-reference metrics. While global-only conditioning captures coarse degradation properties, adding intermediate tokens leads to modest gains and more stable fidelity metrics (e.g., PSNR, SSIM). The full multi-scale configuration, which uses 21 tokens, yields the strongest performance across nearly all perceptual measures, indicating that richer degradation representations provide the diffusion model with more precise conditioning.

In practice, the number of tokens is constrained by the spatial embedding size of the degradation features. Our design uses a $4{\times}4$ grid at the latest scale, which implies that the smallest usable face crop is $16{\times}16$ pixels; already a very small face. Taking this constraint into account, we adopt the 21-token configuration as a good balance between expressive multi-scale conditioning and robustness to small face crops.

\subsection{Identity Preservation}
{
In~\autoref{tab:identity_similarity}, we show the face identity preservation accuracy of the different methods. 
In particular, for each model output, we cropped the face region of the super-resolved image and computed the ArcFace~\cite{deng2019arcface} cosine similarity between the GT and restored face. The results show that our approach most reliably preserves identity information.
}

\subsection{Performance Comparison: Our Model vs. S3Diff on Our Dataset}
{
To provide a fairer comparison to our approach, in~\autoref{tab:synthetic_val_our_train}, we compare against the closest model to ours (S3Diff~\cite{zhang2024degradation}) trained on the same InScene dataset and using the exact same experimental setup. 
We call this variant S3Diff\textsuperscript{**}. 
While the results of the two models are close on the synthetic validation dataset, 
there is a noticeable gap in the real validation dataset between our approach and S3Diff variants, especially for reference-based metrics. This is likely due to having more diverse test examples in the real validation set, compared to the synthetic validation set. 
Indeed, compared to the synthetic dataset (derived from celebrity photos), the real data not only has a more complex distribution of people, but the content of the images themselves (e.g., the pose of the inserted people, or the level of detail in the background) is more complicated.
Overall, the results suggest that the improvement gained by our model is more generalizable to out-of-domain data.
We hypothesize this is due to
more accurate degradation estimation and injection, which avoids overfitting to the training data, in part by explicitly disentangling content from degradation.
}

\subsection{Robustness to Reference Quality}

We analyze the sensitivity of Face2Scene to the quality of the reference-based face restoration module (Ref-FR) on InScene Synthetic Validation Dataset.  We generate Ref-FR outputs with three different quality levels: \textit{good}, \textit{medium} (default), and \textit{bad}, using a controlled synthetic perturbation, and compare the resulting scene restoration in~\autoref{tab:degrade}. 
The corresponding restored-face quality is shown in the last row of the table, where the effect of the synthetic perturbation is reflected in the quantitative results. As shown, Face2Scene degrades \emph{gracefully} as Ref-FR quality worsens and remains competitive with S3Diff, indicating that Stage~II does not require near-perfect Ref-FR to be effective.

\begin{table*}[!]
\centering
\resizebox{1\linewidth}{!}{%
\begin{tabular}{l|ccccc|ccccc}
\toprule
\textbf{Method}
& DISTS$\downarrow$
& LPIPS$\downarrow$
& PSNR$\uparrow$
& SSIM$\uparrow$
& FID$\downarrow$
& MUSIQ$\uparrow$
& C-IQA$\uparrow$
& M-IQA$\uparrow$
& LIQE$\uparrow$
& TOPIQ$\uparrow$ \\
\midrule
\textbf{Face2Scene}
& 0.099 / 0.101 / 0.106
& 0.239 / 0.242 / 0.251
& 23.01 / 22.90 / 22.86
& 0.562 / 0.557 / 0.562
& 15.03 / 15.26 / 17.83
& 74.40 / 74.76 / 74.81
& 0.762 / 0.764 / 0.770
& 0.431 / 0.435 / 0.442
& 4.70 / 4.72 / 4.76
& 0.643 / 0.652 / 0.652 \\
\textbf{S3Diff}
& 0.113
& 0.256
& 23.60
& 0.592
& 18.06
& 72.18
& 0.698
& 0.386
& 4.42
& 0.623 \\
\midrule
\textbf{FaceMe}
& 0.219 / 0.227 / 0.238
& 0.375 / 0.403 / 0.452
& 25.65 / 24.73 / 22.98
& 0.729 / 0.711 / 0.515
& 58.74 / 62.73 / 68.37
& 74.21 / 74.57 / 74.29
& 0.676 / 0.684 / 0.688
& 0.558 / 0.564 / 0.562
& 4.90 / 4.93 / 4.89
& 0.679 / 0.681 / 0.680 \\
\bottomrule
\end{tabular}}
\caption{
\textbf{Robustness to reference restoration quality on the InScene Synthetic validation set.}
For Face2Scene, the Stage~I Ref-FR model produces reference faces at three quality levels (\textit{good / medium / bad}). 
The table reports the resulting scene restoration performance when these references are used in Stage~II. 
As the reference quality decreases, Face2Scene shows only moderate degradation across perceptual and no-reference quality metrics, indicating that the scene restoration stage is robust to imperfect face restoration. 
In particular, even with degraded references, Face2Scene remains competitive with S3Diff, demonstrating that Stage~II does not require near-perfect reference faces to operate effectively.
}
\label{tab:degrade}
\vspace{-0.8em}
\end{table*}

\subsection{Comparison with Face Restoration Models}

Face-only restoration methods are not directly comparable to our primary setting, as they assume cropped and aligned face inputs and do not reconstruct the full scene. Nevertheless, we evaluate Face2Scene on the dedicated face restoration benchmark CelebA-Ref, which contains 100 identities (1064 images) that we used for the InScene synthetic validation split and are unseen during training (see~\autoref{tab:dataset}). Despite not being designed as a specialized face restoration model, Face2Scene achieves the best overall performance on most metrics compared with strong non-reference face restoration baselines (\autoref{tab:celebA_ref}). This improvement is expected, as our method leverages identity reference images that provide consistent appearance and identity cues, whereas competing methods must hallucinate missing facial details from a single degraded face crop.

\begin{table}[!]
\centering
\footnotesize
\resizebox{1.0\linewidth}{!}{%
\begin{tabular}{l|ccccc|cc}
\toprule
\textbf{Method}
& \textbf{DISTS$\downarrow$}
& \textbf{LPIPS$\downarrow$}
& \textbf{PSNR$\uparrow$}
& \textbf{SSIM$\uparrow$}
& \textbf{FID$\downarrow$}
& \textbf{MUSIQ$\uparrow$}
& \textbf{LIQE$\uparrow$} \\
\midrule
\textbf{CodeFormer~\cite{zhou2022towards}}
& 0.1665 & 0.2504 & \cellcolor{light}{26.0058} & \cellcolor{light}{0.7089} & 26.27 & \cellcolor{light}{75.3111} & 4.8531 \\
\textbf{OSDFace~\cite{wang2025osdface}}
& \cellcolor{light}{0.1606} & \cellcolor{light}{0.2485} & 24.7144 & 0.7009 & \cellcolor{light}{22.31} & \cellcolor{purple}{75.4782} & \cellcolor{light}{4.8672} \\
\textbf{RestoreFormer++~\cite{wang2023restoreformer++}}
& 0.2012 & 0.3925 & 20.4579 & 0.5954 & 26.67 & 74.5379 & 4.7613 \\
\midrule
\textbf{Face2Scene}
& \cellcolor{purple}{0.1432} & \cellcolor{purple}{0.2180} & \cellcolor{purple}{26.4253} & \cellcolor{purple}{0.7367} & \cellcolor{purple}{19.21} & 74.6088 & \cellcolor{purple}{4.8789} \\
\bottomrule
\end{tabular}}
\caption{\textbf{Face restoration comparison on CelebA-Ref.} Face2Scene achieves the best performance on most metrics against dedicated face restoration baselines. This is consistent with the use of reference images, which provide additional identity and appearance cues.
}
\label{tab:celebA_ref}
\end{table}

\subsection{Spatially Varying Degradation}
\label{sec:spatial_blur}

\begin{figure*}[htb]
    \centering
    \newcommand{\cropheight}{0.20\textwidth}
    \resizebox{0.95\textwidth}{!}{
    \begin{tabular}{@{} c *{6}{c} @{}}

    \rotatebox{90}{LQ} &
    \includegraphics[width=0.20\textwidth]{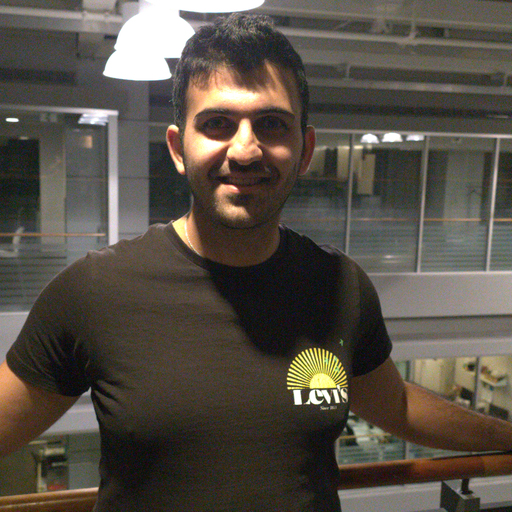} &
    \includegraphics[height=\cropheight]{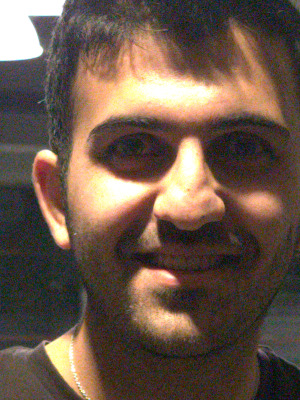} &
    \includegraphics[height=\cropheight]{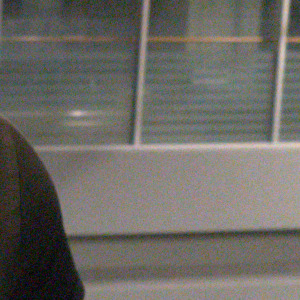} &
    \includegraphics[width=0.20\textwidth]{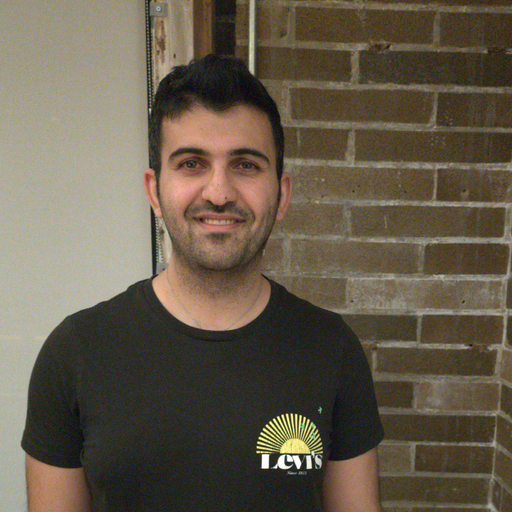} &
    \includegraphics[height=\cropheight]{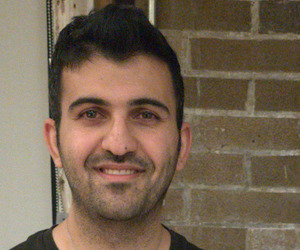} &
    \includegraphics[height=\cropheight]{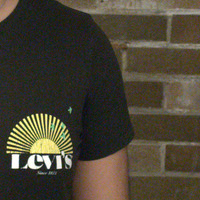} \\

    \rotatebox{90}{Face2Scene} &
    \includegraphics[width=0.20\textwidth]{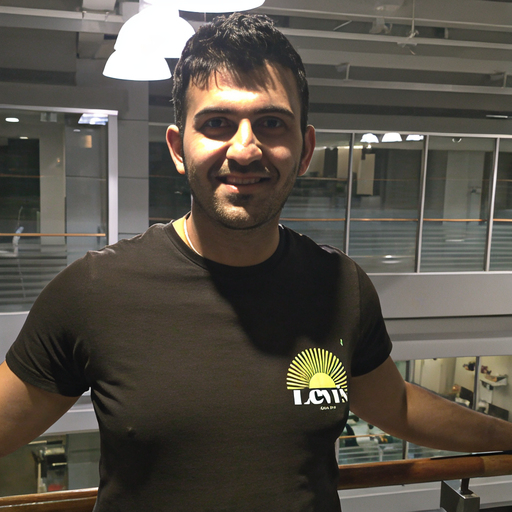} &
    \includegraphics[height=\cropheight]{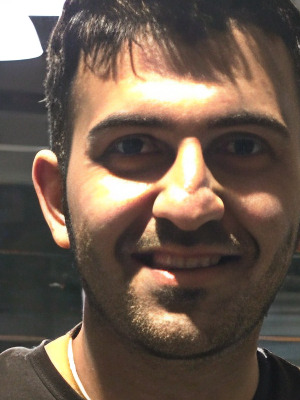} &
    \includegraphics[height=\cropheight]{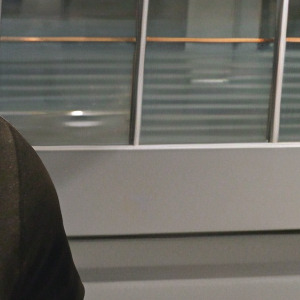} &
    \includegraphics[width=0.20\textwidth]{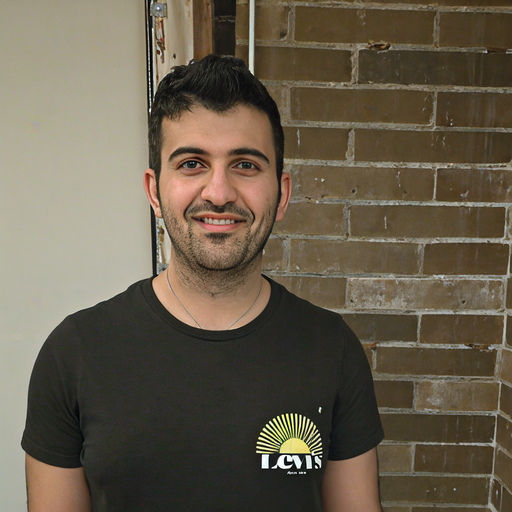} &
    \includegraphics[height=\cropheight]{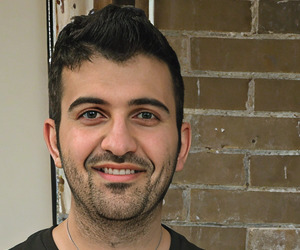} &
    \includegraphics[height=\cropheight]{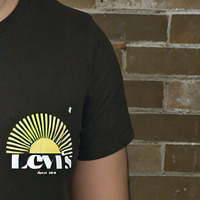} \\

    \rotatebox{90}{S3Diff} &
    \includegraphics[width=0.20\textwidth]{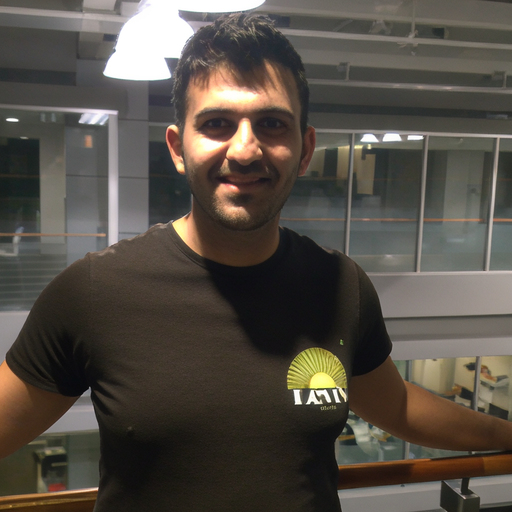} &
    \includegraphics[height=\cropheight]{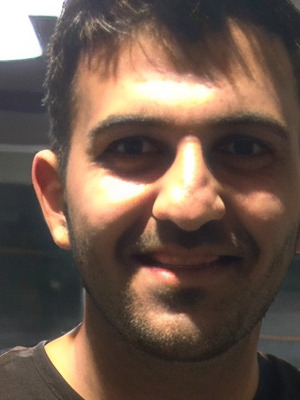} &
    \includegraphics[height=\cropheight]{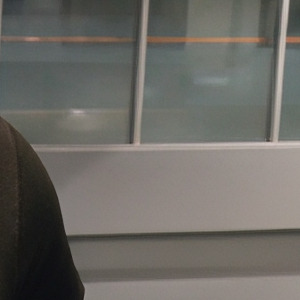} &
    \includegraphics[width=0.20\textwidth]{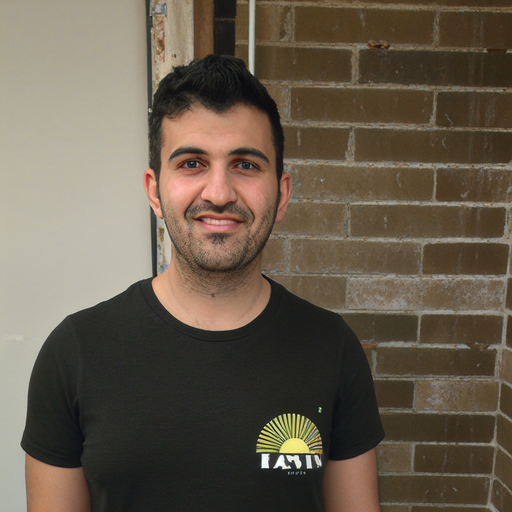} &
    \includegraphics[height=\cropheight]{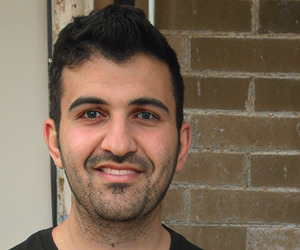} &
    \includegraphics[height=\cropheight]{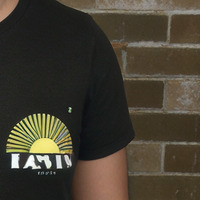} \\
\end{tabular}
    }
    \caption{\textbf{Visual comparison across restoration methods on real test samples.}}
    \label{fig:real_test_samples}
\end{figure*}

To evaluate robustness to spatially non-uniform degradations, we modify the \emph{first blur stage} of the Real-ESRGAN degradation pipeline to apply a spatially varying blur kernel to generate new LQ samples for InScene Synthetic Validation Dataset. Following Lin \textit{et al.}~\cite{lin2025learning}, we model blur as a position-dependent Point Spread Function (PSF) field, yielding milder blur near the image center and progressively stronger blur toward the image corners. Quantitative results are reported in~\autoref{tab:spatial_blur}. Although Face2Scene is trained under the assumption of globally consistent degradations, it remains robust in this setting and outperforms S3Diff on most perceptual and no-reference quality metrics. Representative qualitative examples are also shown in~\autoref{fig:real_test_samples}. We find that Face2Scene restores facial regions reliably even under this more challenging degradation, whereas most competing methods struggle to recover clear facial details. Some background regions still exhibit mild residual blur, highlighting a limitation of the current formulation when the degradation is not spatially uniform.

\begin{table}[t]
\centering
\footnotesize
\resizebox{1.0\columnwidth}{!}{%
\begin{tabular}{l|ccccc|ccccc}
\toprule
\textbf{Method}
& \textbf{DISTS$\downarrow$}
& \textbf{LPIPS$\downarrow$}
& \textbf{PSNR$\uparrow$}
& \textbf{SSIM$\uparrow$}
& \textbf{FID$\downarrow$}
& \textbf{MUSIQ$\uparrow$}
& \textbf{C-IQA$\uparrow$}
& \textbf{M-IQA$\uparrow$}
& \textbf{LIQE$\uparrow$}
& \textbf{TOPIQ$\uparrow$} \\
\midrule
\textbf{Face2Scene}
& \textbf{0.1137} & \textbf{0.2599} & 22.9552 & 0.5571 & \textbf{17.88} & \textbf{73.9026} & \textbf{0.7586} & \textbf{0.4269} & \textbf{4.6433} & \textbf{0.6405} \\
\textbf{S3Diff}
& 0.1275 & 0.2794 & \textbf{23.8162} & \textbf{0.5923} & 19.55 & 70.9485 & 0.6870 & 0.3800 & 4.2837 & 0.6143 \\
\bottomrule
\end{tabular}}
\caption{\textbf{Comparison under spatially varying blur.}
We replace the first blur stage of the Real-ESRGAN degradation pipeline with a position-dependent blur field, such that blur is weaker near the center and stronger toward the corners. Although trained assuming globally consistent degradations, Face2Scene remains robust and outperforms S3Diff on most perceptual and no-reference metrics, indicating better perceptual restoration under spatially varying blur.}
\label{tab:spatial_blur}
\end{table}

\begin{table}[t]
    \centering
    \footnotesize
    \resizebox{1.0\columnwidth}{!}{%
    \begin{tabular}{l|ccccc|ccccc}
    \toprule
    \textbf{Method}
    & \textbf{DISTS$\downarrow$}
    & \textbf{LPIPS$\downarrow$}
    & \textbf{PSNR$\uparrow$}
    & \textbf{SSIM$\uparrow$}
    & \textbf{FID$\downarrow$}
    & \textbf{MUSIQ$\uparrow$}
    & \textbf{C-IQA$\uparrow$}
    & \textbf{M-IQA$\uparrow$}
    & \textbf{LIQE$\uparrow$}
    & \textbf{TOPIQ$\uparrow$} \\
    \midrule
    \textbf{Face2Scene}
    & \textbf{0.1007} & \textbf{0.2421} & 22.9040 & 0.5574 & \textbf{15.26}
    & \textbf{74.7630} & \textbf{0.7640} & \textbf{0.4347} & \textbf{4.7157} & \textbf{0.6515} \\
    
    \textbf{RedegNet}~\cite{li2022face}
    & 0.1675 & 0.3868 & \textbf{24.7300} & \textbf{0.6290} & 36.06
    & 66.9653 & 0.4893 & 0.3056 & 3.1354 & 0.5181 \\
    \bottomrule
    \end{tabular}}
    \caption{\textbf{Comparison between Face2Scene and RedegNet~\cite{li2022face} on InScene Synthetic Validation Dataset.}
    Face2Scene achieves better perceptual quality and no-reference image quality metrics, while RedegNet attains higher PSNR and SSIM. ReDegNet learns realistic degradation representations from face images and transfers them to natural images to synthesize degraded training data for blind image super-resolution.}
    \label{tab:face2scene_redegnet}
\end{table}

\section{Failure Modes}

\begin{figure*}[t]
    \centering
    \includegraphics[width=1\linewidth]{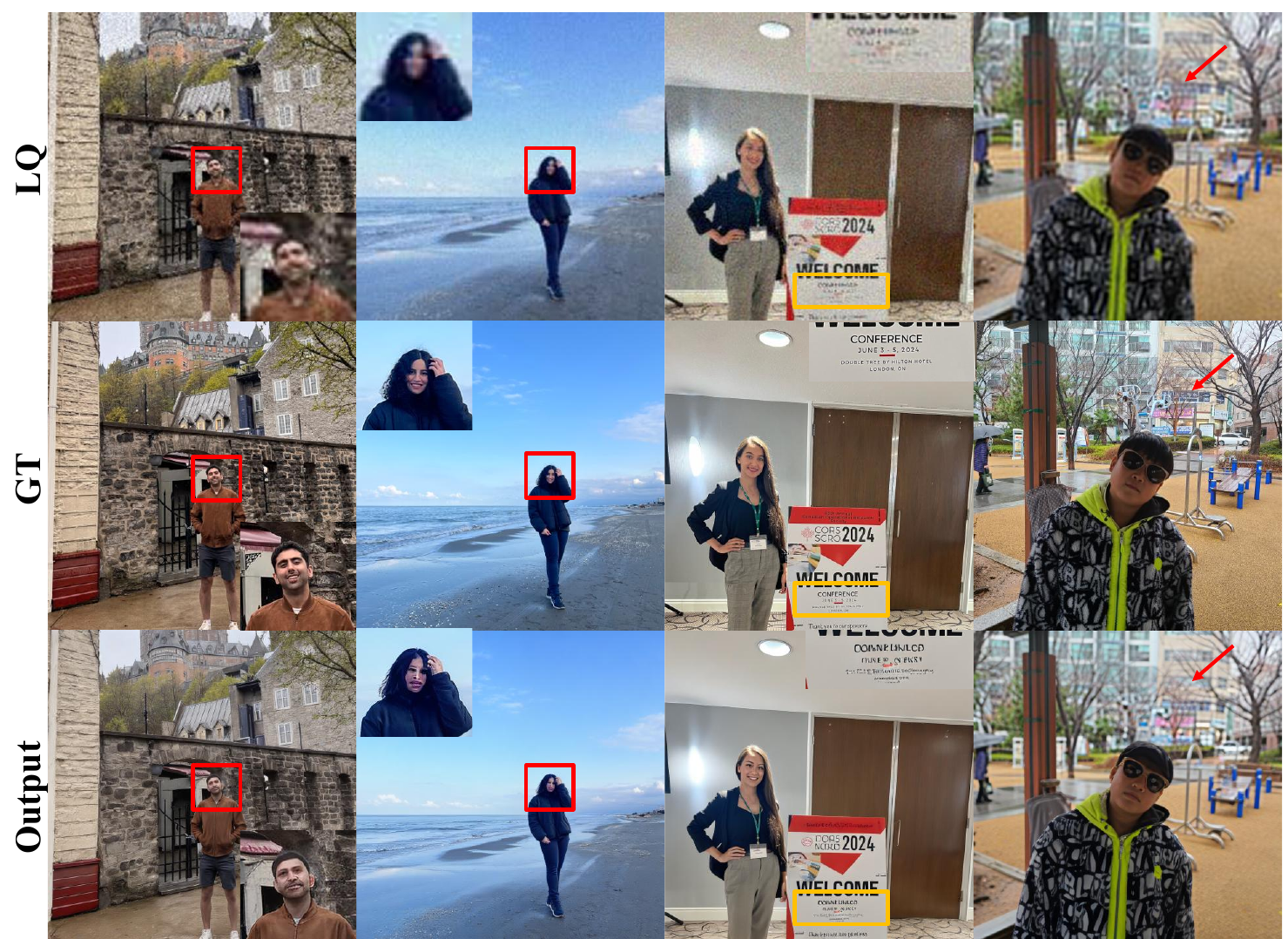}
    \caption{\textbf{Failure Cases.} 
    Our method, like other SD~2.1-based restorers, struggles to recover fine text and very small faces, reflecting the backbone’s limited text rendering ability and difficulty preserving details in tiny spatial regions. More advanced backbones such as FLUX or SD3.5, which provide higher native resolution and stronger text/face modeling, could alleviate these issues.}
    \label{fig:failure}
\end{figure*}

We show some failure cases of our model in~\autoref{fig:failure}. In particular, we observe that our model has difficulty restoring the text regions and small faces. Though such content is inherently difficult, these issues are also due in part to using stable diffusion (SD) 2.1~\cite{sauer2024adversarial} as our diffusion backbone, which we observe has limitations in handling text and reconstructions with small contexts. Using FLUX~\cite{batifol2025flux} or SD3.5~\cite{esser2024scaling}, which generate bigger images with more details and can also better handle text, would help overcome these limitations.
Separately, another limitation is restoring different degrees of blur due to depth of field. As our method's degradations are based on the face region, it does not account for different levels of degradations (i.e., spatial variation). This problem can be alleviated, for example, by adding a depth-dependent blur estimation module to compensate for such cases.

{
  \small
  \bibliographystyle{ieeenat_fullname}
  \bibliography{main}

\begin{thebibliography}{68}
\providecommand{\natexlab}[1]{#1}
\providecommand{\url}[1]{\texttt{#1}}
\expandafter\ifx\csname urlstyle\endcsname\relax
  \providecommand{\doi}[1]{doi: #1}\else
  \providecommand{\doi}{doi: \begingroup \urlstyle{rm}\Url}\fi

\bibitem[Batifol et~al.(2025)Batifol, Blattmann, Boesel, Consul, Diagne, Dockhorn, English, English, Esser, Kulal, et~al.]{batifol2025flux}
Stephen Batifol, Andreas Blattmann, Frederic Boesel, Saksham Consul, Cyril Diagne, Tim Dockhorn, Jack English, Zion English, Patrick Esser, Sumith Kulal, et~al.
\newblock Flux. 1 kontext: Flow matching for in-context image generation and editing in latent space.
\newblock \emph{arXiv e-prints}, pages arXiv--2506, 2025.

\bibitem[Bi et~al.(2024)Bi, Luo, Shen, Zhang, Yue, and Yang]{bi2024deedsr}
Chunyang Bi, Xin Luo, Sheng Shen, Mengxi Zhang, Huanjing Yue, and Jingyu Yang.
\newblock Deedsr: Towards real-world image super-resolution via degradation-aware stable diffusion.
\newblock \emph{arXiv preprint arXiv:2404.00661}, 2024.

\bibitem[Blau and Michaeli(2018)]{blau2018perception}
Yochai Blau and Tomer Michaeli.
\newblock The perception-distortion tradeoff.
\newblock In \emph{Proceedings of the IEEE conference on computer vision and pattern recognition}, pages 6228--6237, 2018.

\bibitem[Caron et~al.(2021)Caron, Touvron, Misra, J{\'e}gou, Mairal, Bojanowski, and Joulin]{caron2021emerging}
Mathilde Caron, Hugo Touvron, Ishan Misra, Herv{\'e} J{\'e}gou, Julien Mairal, Piotr Bojanowski, and Armand Joulin.
\newblock Emerging properties in self-supervised vision transformers.
\newblock In \emph{Proceedings of the IEEE/CVF international conference on computer vision}, pages 9650--9660, 2021.

\bibitem[Changpinyo et~al.(2021)Changpinyo, Sharma, Ding, and Soricut]{changpinyo2021conceptual}
Soravit Changpinyo, Piyush Sharma, Nan Ding, and Radu Soricut.
\newblock Conceptual 12m: Pushing web-scale image-text pre-training to recognize long-tail visual concepts.
\newblock In \emph{Proceedings of the IEEE/CVF conference on computer vision and pattern recognition}, pages 3558--3568, 2021.

\bibitem[Chari et~al.(2025)Chari, Ma, Ostashev, Kadambi, Krishnan, Wang, and Aberman]{chari2025personalized}
Pradyumna Chari, Sizhuo Ma, Daniil Ostashev, Achuta Kadambi, Gurunandan Krishnan, Jian Wang, and Kfir Aberman.
\newblock Personalized restoration via dual-pivot tuning.
\newblock \emph{IEEE Transactions on Image Processing}, 2025.

\bibitem[Chen et~al.(2024)Chen, Mo, Hou, Wu, Liao, Sun, Yan, and Lin]{chen2024topiq}
Chaofeng Chen, Jiadi Mo, Jingwen Hou, Haoning Wu, Liang Liao, Wenxiu Sun, Qiong Yan, and Weisi Lin.
\newblock Topiq: A top-down approach from semantics to distortions for image quality assessment.
\newblock \emph{IEEE Transactions on Image Processing}, 33:\penalty0 2404--2418, 2024.

\bibitem[Chen et~al.(2020)Chen, Kornblith, Norouzi, and Hinton]{chen2020simple}
Ting Chen, Simon Kornblith, Mohammad Norouzi, and Geoffrey Hinton.
\newblock A simple framework for contrastive learning of visual representations.
\newblock In \emph{International conference on machine learning}, pages 1597--1607. PmLR, 2020.

\bibitem[Chen et~al.(2023)Chen, Zhang, Gu, Yuan, Kong, Chen, and Yang]{chen2023image}
Zheng Chen, Yulun Zhang, Jinjin Gu, Xin Yuan, Linghe Kong, Guihai Chen, and Xiaokang Yang.
\newblock Image super-resolution with text prompt diffusion.
\newblock \emph{arXiv preprint arXiv:2311.14282}, 2023.

\bibitem[Chong et~al.(2025)Chong, Xu, Zhang, Wang, Forsyth, Krishnan, Wu, and Wang]{chong2025copy}
Min~Jin Chong, Dejia Xu, Yi Zhang, Zhangyang Wang, David Forsyth, Gurunandan Krishnan, Yicheng Wu, and Jian Wang.
\newblock Copy or not? reference-based face image restoration with fine details.
\newblock In \emph{2025 IEEE/CVF Winter Conference on Applications of Computer Vision (WACV)}, pages 9660--9669. IEEE, 2025.

\bibitem[Deng et~al.(2019)Deng, Guo, Xue, and Zafeiriou]{deng2019arcface}
Jiankang Deng, Jia Guo, Niannan Xue, and Stefanos Zafeiriou.
\newblock Arcface: Additive angular margin loss for deep face recognition.
\newblock In \emph{Proceedings of the IEEE/CVF conference on computer vision and pattern recognition}, pages 4690--4699, 2019.

\bibitem[Ding et~al.(2020)Ding, Ma, Wang, and Simoncelli]{ding2020image}
Keyan Ding, Kede Ma, Shiqi Wang, and Eero~P Simoncelli.
\newblock Image quality assessment: Unifying structure and texture similarity.
\newblock \emph{IEEE transactions on pattern analysis and machine intelligence}, 44\penalty0 (5):\penalty0 2567--2581, 2020.

\bibitem[Ding et~al.(2024)Ding, Zhang, Tu, and Xia]{ding2024restoration}
Zheng Ding, Xuaner Zhang, Zhuowen Tu, and Zhihao Xia.
\newblock Restoration by generation with constrained priors.
\newblock In \emph{Proceedings of the IEEE/CVF Conference on Computer Vision and Pattern Recognition}, pages 2567--2577, 2024.

\bibitem[Esser et~al.(2024)Esser, Kulal, Blattmann, Entezari, M{\"u}ller, Saini, Levi, Lorenz, Sauer, Boesel, et~al.]{esser2024scaling}
Patrick Esser, Sumith Kulal, Andreas Blattmann, Rahim Entezari, Jonas M{\"u}ller, Harry Saini, Yam Levi, Dominik Lorenz, Axel Sauer, Frederic Boesel, et~al.
\newblock Scaling rectified flow transformers for high-resolution image synthesis.
\newblock In \emph{Forty-first international conference on machine learning}, 2024.

\bibitem[Gong et~al.(2025{\natexlab{a}})Gong, Wang, Chen, Liu, Gu, Zhang, and Yang]{gong2025human}
Jue Gong, Jingkai Wang, Zheng Chen, Xing Liu, Hong Gu, Yulun Zhang, and Xiaokang Yang.
\newblock Human body restoration with one-step diffusion model and a new benchmark.
\newblock In \emph{International Conference on Machine Learning}, 2025{\natexlab{a}}.

\bibitem[Gong et~al.(2025{\natexlab{b}})Gong, Yang, Wang, Chen, Liu, Gu, Zhang, and Yang]{gong2025haodiff}
Jue Gong, Tingyu Yang, Jingkai Wang, Zheng Chen, Xing Liu, Hong Gu, Yulun Zhang, and Xiaokang Yang.
\newblock Haodiff: Human-aware one-step diffusion via dual-prompt guidance.
\newblock \emph{arXiv preprint arXiv:2505.19742}, 2025{\natexlab{b}}.

\bibitem[He et~al.(2020)He, Fan, Wu, Xie, and Girshick]{he2020momentum}
Kaiming He, Haoqi Fan, Yuxin Wu, Saining Xie, and Ross Girshick.
\newblock Momentum contrast for unsupervised visual representation learning.
\newblock In \emph{Conference on computer vision and pattern recognition}, 2020.

\bibitem[Heusel et~al.(2017)Heusel, Ramsauer, Unterthiner, Nessler, and Hochreiter]{heusel2017gans}
Martin Heusel, Hubert Ramsauer, Thomas Unterthiner, Bernhard Nessler, and Sepp Hochreiter.
\newblock Gans trained by a two time-scale update rule converge to a local nash equilibrium.
\newblock \emph{Advances in neural information processing systems}, 30, 2017.

\bibitem[Ho and Salimans(2022)]{ho2022classifier}
Jonathan Ho and Tim Salimans.
\newblock Classifier-free diffusion guidance.
\newblock \emph{arXiv preprint arXiv:2207.12598}, 2022.

\bibitem[Ho et~al.(2020)Ho, Jain, and Abbeel]{ho2020denoising}
Jonathan Ho, Ajay Jain, and Pieter Abbeel.
\newblock Denoising diffusion probabilistic models.
\newblock \emph{Advances in neural information processing systems}, 33:\penalty0 6840--6851, 2020.

\bibitem[Hsiao et~al.(2024)Hsiao, Liu, Yang, Kuo, Jou, and Chen]{hsiao2024ref}
Chi-Wei Hsiao, Yu-Lun Liu, Cheng-Kun Yang, Sheng-Po Kuo, Kevin Jou, and Chia-Ping Chen.
\newblock Ref-ldm: A latent diffusion model for reference-based face image restoration.
\newblock \emph{Advances in Neural Information Processing Systems}, 37:\penalty0 74840--74867, 2024.

\bibitem[Ioffe and Szegedy(2015)]{ioffe2015batch}
Sergey Ioffe and Christian Szegedy.
\newblock Batch normalization: Accelerating deep network training by reducing internal covariate shift.
\newblock In \emph{International conference on machine learning}, 2015.

\bibitem[Jiang et~al.(2025)Jiang, Yan, Jia, Liu, Kang, and Lu]{jiang2025infiniteyou}
Liming Jiang, Qing Yan, Yumin Jia, Zichuan Liu, Hao Kang, and Xin Lu.
\newblock {InfiniteYou}: Flexible photo recrafting while preserving your identity.
\newblock \emph{International conference on computer vision (ICCV)}, 2025.

\bibitem[Ke et~al.(2021)Ke, Wang, Wang, Milanfar, and Yang]{ke2021musiq}
Junjie Ke, Qifei Wang, Yilin Wang, Peyman Milanfar, and Feng Yang.
\newblock Musiq: Multi-scale image quality transformer.
\newblock In \emph{Proceedings of the IEEE/CVF international conference on computer vision}, pages 5148--5157, 2021.

\bibitem[Kong et~al.(2025)Kong, Li, Wang, Xu, Pei, Li, and Ren]{kong2025dual}
Dehong Kong, Fan Li, Zhixin Wang, Jiaqi Xu, Renjing Pei, Wenbo Li, and WenQi Ren.
\newblock Dual prompting image restoration with diffusion transformers.
\newblock In \emph{Proceedings of the Computer Vision and Pattern Recognition Conference}, pages 12809--12819, 2025.

\bibitem[Kumari et~al.(2022)Kumari, Zhang, Shechtman, and Zhu]{kumari2022ensembling}
Nupur Kumari, Richard Zhang, Eli Shechtman, and Jun-Yan Zhu.
\newblock Ensembling off-the-shelf models for gan training.
\newblock In \emph{Proceedings of the IEEE/CVF conference on computer vision and pattern recognition}, pages 10651--10662, 2022.

\bibitem[Li et~al.(2022{\natexlab{a}})Li, Chen, Lin, Zuo, and Zhang]{li2022face}
Xiaoming Li, Chaofeng Chen, Xianhui Lin, Wangmeng Zuo, and Lei Zhang.
\newblock From face to natural image: Learning real degradation for blind image super-resolution.
\newblock In \emph{European Conference on Computer Vision}, pages 376--392. Springer, 2022{\natexlab{a}}.

\bibitem[Li et~al.(2022{\natexlab{b}})Li, Zhang, Zhou, Zhang, and Zuo]{li2022learning}
Xiaoming Li, Shiguang Zhang, Shangchen Zhou, Lei Zhang, and Wangmeng Zuo.
\newblock Learning dual memory dictionaries for blind face restoration.
\newblock \emph{IEEE Transactions on Pattern Analysis and Machine Intelligence}, 45\penalty0 (5):\penalty0 5904--5917, 2022{\natexlab{b}}.

\bibitem[Lin et~al.(2025)Lin, Wang, Lin, Miau, Kainz, Chen, Zhang, Lindell, and Kutulakos]{lin2025learning}
Esther~YH Lin, Zhecheng Wang, Rebecca Lin, Daniel Miau, Florian Kainz, Jiawen Chen, Xuaner Zhang, David~B Lindell, and Kiriakos~N Kutulakos.
\newblock Learning lens blur fields.
\newblock \emph{IEEE Transactions on Pattern Analysis and Machine Intelligence}, 2025.

\bibitem[Lin et~al.(2024)Lin, He, Chen, Lyu, Dai, Yu, Qiao, Ouyang, and Dong]{lin2024diffbir}
Xinqi Lin, Jingwen He, Ziyan Chen, Zhaoyang Lyu, Bo Dai, Fanghua Yu, Yu Qiao, Wanli Ouyang, and Chao Dong.
\newblock Diffbir: Toward blind image restoration with generative diffusion prior.
\newblock In \emph{European conference on computer vision}, pages 430--448. Springer, 2024.

\bibitem[Liu et~al.(2025)Liu, Duan, OuYang, Fu, Park, Liu, Guo, and Li]{liu2025faceme}
Siyu Liu, Zheng-Peng Duan, Jia OuYang, Jiayi Fu, Hyunhee Park, Zikun Liu, Chun-Le Guo, and Chongyi Li.
\newblock Faceme: Robust blind face restoration with personal identification.
\newblock In \emph{Proceedings of the AAAI Conference on Artificial Intelligence}, pages 5567--5575, 2025.

\bibitem[Liu et~al.(2021)Liu, Zhang, Xu, Yang, and Tai]{liu2021accurate}
Yunan Liu, Shanshan Zhang, Jie Xu, Jian Yang, and Yu-Wing Tai.
\newblock An accurate and lightweight method for human body image super-resolution.
\newblock \emph{IEEE Transactions on Image Processing}, 30:\penalty0 2888--2897, 2021.

\bibitem[Liu et~al.(2024)Liu, He, Liu, Lin, Yu, Hu, Qiao, and Dong]{liu2024adaptbir}
Yingqi Liu, Jingwen He, Yihao Liu, Xinqi Lin, Fanghua Yu, Jinfan Hu, Yu Qiao, and Chao Dong.
\newblock Adaptbir: Adaptive blind image restoration with latent diffusion prior for higher fidelity.
\newblock \emph{Pattern Recognition}, 155:\penalty0 110659, 2024.

\bibitem[Mou et~al.(2022)Mou, Wu, Wang, Dong, Zhang, and Shan]{mou2022metric}
Chong Mou, Yanze Wu, Xintao Wang, Chao Dong, Jian Zhang, and Ying Shan.
\newblock Metric learning based interactive modulation for real-world super-resolution.
\newblock In \emph{European conference on computer vision}, pages 723--740. Springer, 2022.

\bibitem[Qi et~al.(2023)Qi, Tu, Ye, Delbracio, Milanfar, Chen, and Talebi]{qi2023tip}
Chenyang Qi, Zhengzhong Tu, Keren Ye, Mauricio Delbracio, Peyman Milanfar, Qifeng Chen, and Hossein Talebi.
\newblock Tip: Text-driven image processing with semantic and restoration instructions.
\newblock \emph{CoRR}, 2023.

\bibitem[Redmon et~al.(2016)Redmon, Divvala, Girshick, and Farhadi]{redmon2016you}
Joseph Redmon, Santosh Divvala, Ross Girshick, and Ali Farhadi.
\newblock You only look once: Unified, real-time object detection.
\newblock In \emph{Proceedings of the IEEE conference on computer vision and pattern recognition}, pages 779--788, 2016.

\bibitem[Rombach et~al.(2022)Rombach, Blattmann, Lorenz, Esser, and Ommer]{rombach2022high}
Robin Rombach, Andreas Blattmann, Dominik Lorenz, Patrick Esser, and Bj{\"o}rn Ommer.
\newblock High-resolution image synthesis with latent diffusion models.
\newblock In \emph{Proceedings of the IEEE/CVF conference on computer vision and pattern recognition}, pages 10684--10695, 2022.

\bibitem[Sauer et~al.(2024)Sauer, Lorenz, Blattmann, and Rombach]{sauer2024adversarial}
Axel Sauer, Dominik Lorenz, Andreas Blattmann, and Robin Rombach.
\newblock Adversarial diffusion distillation.
\newblock In \emph{European Conference on Computer Vision}, pages 87--103. Springer, 2024.

\bibitem[Tao et~al.(2024)Tao, Gu, Zhang, Wang, and Cheng]{tao2024overcoming}
Keda Tao, Jinjin Gu, Yulun Zhang, Xiucheng Wang, and Nan Cheng.
\newblock Overcoming false illusions in real-world face restoration with multi-modal guided diffusion model.
\newblock \emph{arXiv preprint arXiv:2410.04161}, 2024.

\bibitem[TAO et~al.(2025)TAO, Gu, Zhang, Wang, and Cheng]{tao2025overcoming}
Keda TAO, Jinjin Gu, Yulun Zhang, Xiucheng Wang, and Nan Cheng.
\newblock Overcoming false illusions in real-world face restoration with multi-modal guided diffusion model.
\newblock In \emph{The Thirteenth International Conference on Learning Representations}, 2025.

\bibitem[Varanka et~al.(2024)Varanka, Toivonen, Tripathy, Zhao, and Acar]{varanka2024pfstorer}
Tuomas Varanka, Tapani Toivonen, Soumya Tripathy, Guoying Zhao, and Erman Acar.
\newblock Pfstorer: Personalized face restoration and super-resolution.
\newblock In \emph{Proceedings of the IEEE/CVF Conference on Computer Vision and Pattern Recognition}, pages 2372--2381, 2024.

\bibitem[Wang et~al.(2023{\natexlab{a}})Wang, Chan, and Loy]{wang2023exploring}
Jianyi Wang, Kelvin~CK Chan, and Chen~Change Loy.
\newblock Exploring clip for assessing the look and feel of images.
\newblock In \emph{Proceedings of the AAAI conference on artificial intelligence}, pages 2555--2563, 2023{\natexlab{a}}.

\bibitem[Wang et~al.(2025{\natexlab{a}})Wang, Gong, Zhang, Chen, Liu, Gu, Liu, Zhang, and Yang]{wang2025osdface}
Jingkai Wang, Jue Gong, Lin Zhang, Zheng Chen, Xing Liu, Hong Gu, Yutong Liu, Yulun Zhang, and Xiaokang Yang.
\newblock Osdface: One-step diffusion model for face restoration.
\newblock In \emph{Proceedings of the Computer Vision and Pattern Recognition Conference}, pages 12626--12636, 2025{\natexlab{a}}.

\bibitem[Wang et~al.(2025{\natexlab{b}})Wang, Miao, Gong, Chen, Liu, Gu, Liu, and Zhang]{wang2025honestface}
Jingkai Wang, Wu Miao, Jue Gong, Zheng Chen, Xing Liu, Hong Gu, Yutong Liu, and Yulun Zhang.
\newblock Honestface: Towards honest face restoration with one-step diffusion model.
\newblock \emph{arXiv preprint arXiv:2505.18469}, 2025{\natexlab{b}}.

\bibitem[Wang et~al.(2021{\natexlab{a}})Wang, Wang, Dong, Xu, Yang, An, and Guo]{wang2021unsupervised}
Longguang Wang, Yingqian Wang, Xiaoyu Dong, Qingyu Xu, Jungang Yang, Wei An, and Yulan Guo.
\newblock Unsupervised degradation representation learning for blind super-resolution.
\newblock In \emph{Proceedings of the IEEE/CVF conference on computer vision and pattern recognition}, pages 10581--10590, 2021{\natexlab{a}}.

\bibitem[Wang et~al.(2024{\natexlab{a}})Wang, Sang, Liu, Wang, Lu, and Sun]{wang2024prior}
Simiao Wang, Yu Sang, Yunan Liu, Chunpeng Wang, Mingyu Lu, and Jinguang Sun.
\newblock Prior based pyramid residual clique network for human body image super-resolution.
\newblock \emph{Pattern Recognition}, 150:\penalty0 110352, 2024{\natexlab{a}}.

\bibitem[Wang et~al.(2018)Wang, Yu, Wu, Gu, Liu, Dong, Qiao, and Change~Loy]{wang2018esrgan}
Xintao Wang, Ke Yu, Shixiang Wu, Jinjin Gu, Yihao Liu, Chao Dong, Yu Qiao, and Chen Change~Loy.
\newblock Esrgan: Enhanced super-resolution generative adversarial networks.
\newblock In \emph{Proceedings of the European conference on computer vision (ECCV) workshops}, pages 0--0, 2018.

\bibitem[Wang et~al.(2021{\natexlab{b}})Wang, Xie, Dong, and Shan]{wang2021real}
Xintao Wang, Liangbin Xie, Chao Dong, and Ying Shan.
\newblock Real-esrgan: Training real-world blind super-resolution with pure synthetic data.
\newblock In \emph{Proceedings of the IEEE/CVF international conference on computer vision}, pages 1905--1914, 2021{\natexlab{b}}.

\bibitem[Wang et~al.(2024{\natexlab{b}})Wang, Yang, Chen, Wang, Guo, Chau, Liu, Qiao, Kot, and Wen]{wang2024sinsr}
Yufei Wang, Wenhan Yang, Xinyuan Chen, Yaohui Wang, Lanqing Guo, Lap-Pui Chau, Ziwei Liu, Yu Qiao, Alex~C Kot, and Bihan Wen.
\newblock Sinsr: diffusion-based image super-resolution in a single step.
\newblock In \emph{Proceedings of the IEEE/CVF conference on computer vision and pattern recognition}, pages 25796--25805, 2024{\natexlab{b}}.

\bibitem[Wang et~al.(2004)Wang, Bovik, Sheikh, and Simoncelli]{wang2004image}
Zhou Wang, Alan~C Bovik, Hamid~R Sheikh, and Eero~P Simoncelli.
\newblock Image quality assessment: from error visibility to structural similarity.
\newblock \emph{IEEE transactions on image processing}, 13\penalty0 (4):\penalty0 600--612, 2004.

\bibitem[Wang et~al.(2023{\natexlab{b}})Wang, Zhang, Chen, Wang, and Luo]{wang2023restoreformer++}
Zhouxia Wang, Jiawei Zhang, Tianshui Chen, Wenping Wang, and Ping Luo.
\newblock Restoreformer++: Towards real-world blind face restoration from undegraded key-value pairs.
\newblock \emph{IEEE Transactions on Pattern Analysis and Machine Intelligence}, 45\penalty0 (12):\penalty0 15462--15476, 2023{\natexlab{b}}.

\bibitem[Wu et~al.(2024)Wu, Sun, Ma, and Zhang]{wu2024one}
Rongyuan Wu, Lingchen Sun, Zhiyuan Ma, and Lei Zhang.
\newblock One-step effective diffusion network for real-world image super-resolution.
\newblock \emph{Advances in Neural Information Processing Systems}, 37:\penalty0 92529--92553, 2024.

\bibitem[Yang et~al.(2022)Yang, Wu, Shi, Lao, Gong, Cao, Wang, and Yang]{yang2022maniqa}
Sidi Yang, Tianhe Wu, Shuwei Shi, Shanshan Lao, Yuan Gong, Mingdeng Cao, Jiahao Wang, and Yujiu Yang.
\newblock Maniqa: Multi-dimension attention network for no-reference image quality assessment.
\newblock In \emph{Proceedings of the IEEE/CVF conference on computer vision and pattern recognition}, pages 1191--1200, 2022.

\bibitem[Yang et~al.(2024)Yang, Wu, Ren, Xie, and Zhang]{yang2024pixel}
Tao Yang, Rongyuan Wu, Peiran Ren, Xuansong Xie, and Lei Zhang.
\newblock Pixel-aware stable diffusion for realistic image super-resolution and personalized stylization.
\newblock In \emph{European conference on computer vision}, pages 74--91. Springer, 2024.

\bibitem[Yin et~al.(2025)Yin, Chen, Liu, Wang, Li, Pei, Li, Lau, and Zuo]{yin2025refstar}
Zhicun Yin, Junjie Chen, Ming Liu, Zhixin Wang, Fan Li, Renjing Pei, Xiaoming Li, Rynson~WH Lau, and Wangmeng Zuo.
\newblock Refstar: Blind facial image restoration with reference selection, transfer, and reconstruction.
\newblock \emph{arXiv preprint arXiv:2507.10470}, 2025.

\bibitem[Ying et~al.(2024)Ying, Liu, Wu, Zhang, Yu, Fu, Cao, Wu, Yu, and Shen]{ying2024restorerid}
Jiacheng Ying, Mushui Liu, Zhe Wu, Runming Zhang, Zhu Yu, Siming Fu, Si-Yuan Cao, Chao Wu, Yunlong Yu, and Hui-Liang Shen.
\newblock Restorerid: Towards tuning-free face restoration with id preservation.
\newblock \emph{arXiv preprint arXiv:2411.14125}, 2024.

\bibitem[Yu et~al.(2024)Yu, Gu, Li, Hu, Kong, Wang, He, Qiao, and Dong]{yu2024scaling}
Fanghua Yu, Jinjin Gu, Zheyuan Li, Jinfan Hu, Xiangtao Kong, Xintao Wang, Jingwen He, Yu Qiao, and Chao Dong.
\newblock Scaling up to excellence: Practicing model scaling for photo-realistic image restoration in the wild.
\newblock In \emph{Proceedings of the IEEE/CVF conference on computer vision and pattern recognition}, pages 25669--25680, 2024.

\bibitem[Yue et~al.(2024)Yue, Wang, and Loy]{yue2024efficient}
Zongsheng Yue, Jianyi Wang, and Chen~Change Loy.
\newblock Efficient diffusion model for image restoration by residual shifting.
\newblock \emph{IEEE Transactions on Pattern Analysis and Machine Intelligence}, 2024.

\bibitem[Yue et~al.(2025)Yue, Liao, and Loy]{yue2025arbitrary}
Zongsheng Yue, Kang Liao, and Chen~Change Loy.
\newblock Arbitrary-steps image super-resolution via diffusion inversion.
\newblock In \emph{Proceedings of the Computer Vision and Pattern Recognition Conference}, pages 23153--23163, 2025.

\bibitem[Zhang et~al.(2024{\natexlab{a}})Zhang, Yue, Pei, Ren, and Cao]{zhang2024degradation}
Aiping Zhang, Zongsheng Yue, Renjing Pei, Wenqi Ren, and Xiaochun Cao.
\newblock Degradation-guided one-step image super-resolution with diffusion priors.
\newblock \emph{arXiv preprint arXiv:2409.17058}, 2024{\natexlab{a}}.

\bibitem[Zhang et~al.(2025)Zhang, Alaluf, Ma, Kadambi, Wang, and Aberman]{zhang2025instantrestore}
Howard Zhang, Yuval Alaluf, Sizhuo Ma, Achuta Kadambi, Jian Wang, and Kfir Aberman.
\newblock Instantrestore: Single-step personalized face restoration with shared-image attention.
\newblock In \emph{Proceedings of the Special Interest Group on Computer Graphics and Interactive Techniques Conference Conference Papers}, pages 1--10, 2025.

\bibitem[Zhang et~al.(2023{\natexlab{a}})Zhang, Rao, and Agrawala]{zhang2023adding}
Lvmin Zhang, Anyi Rao, and Maneesh Agrawala.
\newblock Adding conditional control to text-to-image diffusion models.
\newblock In \emph{Proceedings of the IEEE/CVF international conference on computer vision}, pages 3836--3847, 2023{\natexlab{a}}.

\bibitem[Zhang et~al.(2018)Zhang, Isola, Efros, Shechtman, and Wang]{zhang2018unreasonable}
Richard Zhang, Phillip Isola, Alexei~A Efros, Eli Shechtman, and Oliver Wang.
\newblock The unreasonable effectiveness of deep features as a perceptual metric.
\newblock In \emph{Proceedings of the IEEE conference on computer vision and pattern recognition}, pages 586--595, 2018.

\bibitem[Zhang et~al.(2023{\natexlab{b}})Zhang, Zhai, Wei, Yang, and Ma]{zhang2023blind}
Weixia Zhang, Guangtao Zhai, Ying Wei, Xiaokang Yang, and Kede Ma.
\newblock Blind image quality assessment via vision-language correspondence: A multitask learning perspective.
\newblock In \emph{Proceedings of the IEEE/CVF conference on computer vision and pattern recognition}, pages 14071--14081, 2023{\natexlab{b}}.

\bibitem[Zhang et~al.(2024{\natexlab{b}})Zhang, Huang, Ma, Li, Luo, Xie, Qin, Luo, Li, Liu, et~al.]{zhang2024recognize}
Youcai Zhang, Xinyu Huang, Jinyu Ma, Zhaoyang Li, Zhaochuan Luo, Yanchun Xie, Yuzhuo Qin, Tong Luo, Yaqian Li, Shilong Liu, et~al.
\newblock Recognize anything: A strong image tagging model.
\newblock In \emph{Proceedings of the IEEE/CVF Conference on Computer Vision and Pattern Recognition}, pages 1724--1732, 2024{\natexlab{b}}.

\bibitem[Zhang et~al.(2024{\natexlab{c}})Zhang, Wang, Li, Yuan, Zhang, Sun, Zhong, and Wang]{zhang2024diffbody}
Yiming Zhang, Zhe Wang, Xinjie Li, Yunchen Yuan, Chengsong Zhang, Xiao Sun, Zhihang Zhong, and Jian Wang.
\newblock Diffbody: Human body restoration by imagining with generative diffusion prior.
\newblock \emph{arXiv preprint arXiv:2404.03642}, 2024{\natexlab{c}}.

\bibitem[Zhou et~al.(2025)Zhou, Ye, Shah, Mei, Delbracio, Milanfar, Patel, and Talebi]{zhou2025reference}
Mo Zhou, Keren Ye, Viraj Shah, Kangfu Mei, Mauricio Delbracio, Peyman Milanfar, Vishal~M Patel, and Hossein Talebi.
\newblock Reference-guided identity preserving face restoration.
\newblock \emph{arXiv preprint arXiv:2505.21905}, 2025.

\bibitem[Zhou et~al.(2022)Zhou, Chan, Li, and Loy]{zhou2022towards}
Shangchen Zhou, Kelvin Chan, Chongyi Li, and Chen~Change Loy.
\newblock Towards robust blind face restoration with codebook lookup transformer.
\newblock \emph{Advances in Neural Information Processing Systems}, 35:\penalty0 30599--30611, 2022.

\end{thebibliography}
}


\end{document}